\documentclass{article}

\usepackage[final, nonatbib]{neurips_2021}

\usepackage{graphicx}
\usepackage[utf8]{inputenc} %
\usepackage[T1]{fontenc}    %
\usepackage{url}            %
\usepackage{booktabs}       %
\usepackage{amsfonts}       %
\usepackage{nicefrac}       %
\usepackage{microtype}      %
\usepackage{soul}
\usepackage[table]{xcolor}
\usepackage{floatrow}
\usepackage{xspace}
\usepackage[colorlinks=true,citecolor=blue]{hyperref}
\usepackage{amsmath}
\usepackage{multirow}
\usepackage{wrapfig}
\usepackage{setspace}
\usepackage{listings}
\usepackage{enumitem}

\definecolor{custom_color}{HTML}{C0C0C0}
\sethlcolor{custom_color}

\usepackage{subfiles} %

\newcommand{\modelname}{Primer\xspace}
\newcommand{\modelnamelite}{Primer-EZ\xspace}
\newcommand{\objectivename}{implicit efficiency objective\xspace}

\newfloatcommand{capbtabbox}{table}[][\FBwidth]

\newcommand{\footlabel}[2]{%
    \addtocounter{footnote}{1}%
    \footnotetext[\thefootnote]{%
        \addtocounter{footnote}{-1}%
        \refstepcounter{footnote}\label{#1}%
        #2%
    }%
    $^{\ref{#1}}$%
}

\newcommand{\footref}[1]{%
    $^{\ref{#1}}$%
}

\title{Primer: Searching for Efficient Transformers\\ for Language Modeling}

\author{%
  David R. So, Wojciech Mańke, Hanxiao Liu, Zihang Dai, Noam Shazeer, Quoc V. Le\\
  Google Research, Brain Team\\
  \texttt{\{davidso, 
wojciechm, 
hanxiaol, 
zihangd, noam, qvl\}@google.com}\\
}

\begin{document}

\maketitle

\begin{abstract}

Large Transformer models have been central to recent advances in natural language processing. The training and inference costs of these models, however, have grown rapidly and become prohibitively expensive. Here we aim to reduce the costs of Transformers by searching for a more efficient variant. Compared to previous approaches, our search is performed at a lower level, over the primitives that define a Transformer TensorFlow program. We identify an architecture, named \modelname, that has a smaller training cost than the original Transformer and other variants for auto-regressive language modeling. \modelname's improvements can be mostly attributed to two simple modifications: squaring ReLU activations and adding a depthwise convolution layer after each Q, K, and V projection in self-attention.

Experiments show \modelname's gains over Transformer increase as compute scale grows and follow a power law with respect to quality at optimal model sizes.
We also verify empirically that Primer can be dropped into different codebases to significantly speed up training without additional tuning. For example, at a 500M parameter size, \modelname improves the original T5 architecture on
C4 auto-regressive language modeling, reducing the training cost by 4X. 
Furthermore, the reduced training cost means Primer needs much less compute to reach a target one-shot performance. For instance, in a 1.9B parameter configuration similar to GPT-3 XL, Primer uses 1/3 of the training compute to achieve the same one-shot performance as Transformer. We open source our models and several comparisons in T5 to help with reproducibility.\footlabel{open_source}{\url{https://github.com/google-research/google-research/tree/master/primer}}\looseness=-1

\end{abstract}

\section{Introduction}

Transformers~\cite{Vaswani2017AttentionIA} have been used extensively in many NLP advances over the past few years (e.g.,~\cite{devlin2018bert, yang2019xlnet,liu2019roberta, 2020t5, Adiwardana2020TowardsAH,brown2020language}). With scaling, Transformers have produced increasingly better performance~\cite{yang2019xlnet, brown2020language, Fedus2021SwitchTS,Kaplan2020ScalingLF}, but the costs of training larger models have become prohibitively expensive.

In this paper, we aim to reduce the training costs of Transformer  language models. To this end, we propose searching for more efficient alternatives to Transformer by modifying its TensorFlow computation graph~\cite{Abadi2016TensorFlowAS}. Given a search space of TensorFlow programs, we use evolution~\cite{RealMSSSLK17,liu2017hierarchical,so2019evolved, liu2020evolving, Yao1999EvolvingAN, schmidhuber:1987:srl, Stanley2019DesigningNN} to search for models that achieve as low of a validation loss as possible given a fixed amount of training compute.  An advantage of using TensorFlow programs as the search space is that it is easier to find simple low-level improvements to optimize Transformers. We focus on decoder-only auto-regressive language modeling (LM), because of its  generality and success~\cite{Radford2019LanguageMA, brown2020language, Schick2021ItsNJ, Wang2021EntailmentAF, Gao2020MakingPL}.\footnote{We provide details of our primitives search in TensorFlow, but the same approach can also be applied to other deep learning libraries.}

The discovered model, named \modelname (PRIMitives searched transformER), exhibits strong performance improvements over common Transformer variants on auto-regressive language modeling. Our experiments show that \modelname has the  benefits of (1) achieving a target quality using a smaller training cost, (2) achieving higher quality given a fixed training cost, and (3) achieving a target quality using a smaller inference cost. 
These benefits are robust and hold across  model sizes (20M to 1.9B parameters), across compute scales (10 to 10\textsuperscript{5} accelerator hours), across datasets (LM1B, C4, PG19~\cite{Rae2020CompressiveTF}), across hardware platforms (TPUv2, TPUv3, TPUv4 and V100), across multiple Transformer codebases using default configurations (Tensor2Tensor, Lingvo, and T5) and across multiple model families (dense Transformers~\cite{Vaswani2017AttentionIA}, sparse mixture-of-experts Switch Transformers~\cite{Fedus2021SwitchTS}, and Synthesizers~\cite{tay2020synthesizer}). We open source these comparisons to help with the reproducibility of our results.\footref{open_source}\looseness=-1

Our main finding is that the compute savings of \modelname over Transformers increase as training cost grows, when controlling for model size and quality. These savings follow a power law with respect to quality when using optimally sized models. To demonstrate \modelname's savings in an established training setup, we compare 500M parameter \modelname{} to the original T5 architecture, using the exact configuration used by Raffel et al.~\cite{2020t5} applied to auto-regressive language modeling. In this setting, \modelname{} achieves an improvement of 0.9 perplexity given the same training cost, and reaches quality parity with the T5 baseline model using 4.2X less compute. We further demonstrate that \modelname's savings transfer to one-shot evaluations by comparing \modelname to Transformer at 1.9B parameters in a setup similar to GPT-3 XL~\cite{brown2020language}. There, using 3X less training compute, \modelname achieves similar performance to Transformer on both pretraining perplexity and downstream one-shot tasks.

Our analysis  shows that the improvements of \modelname over Transformer can be mostly attributed to two main modifications: squaring ReLU activations and adding a depthwise convolution layer after each Q, K, and V projection in self-attention. These two modifications are simple and can be dropped into existing Transformer codebases to obtain significant gains for auto-regressive language modeling. \looseness=-1

\section{Search Space and Search Method}
\label{sect:methods}

\paragraph{Searching Over TensorFlow Programs:}
To construct a search space for Transformer alternatives, we use operations from TensorFlow (TF). In this search space, each program defines the stackable decoder block of an \emph{auto-regressive language model}. Given  input tensors $X \in \mathbb{R}^{n \times d}$ that represent sequences of length $n$ with embedding length $d$, our programs return tensors of the same shape. When stacked, their outputs represent next-token prediction embeddings at each sequence position. Our programs only specify model architectures and nothing else. In other words, the input and output embedding matrices themselves, as well as input preprocessing and weight optimization are not within the scope of our programs.\looseness=-1

\begin{figure}[h!]
\centering
\centering
\includegraphics[width=0.65\linewidth]{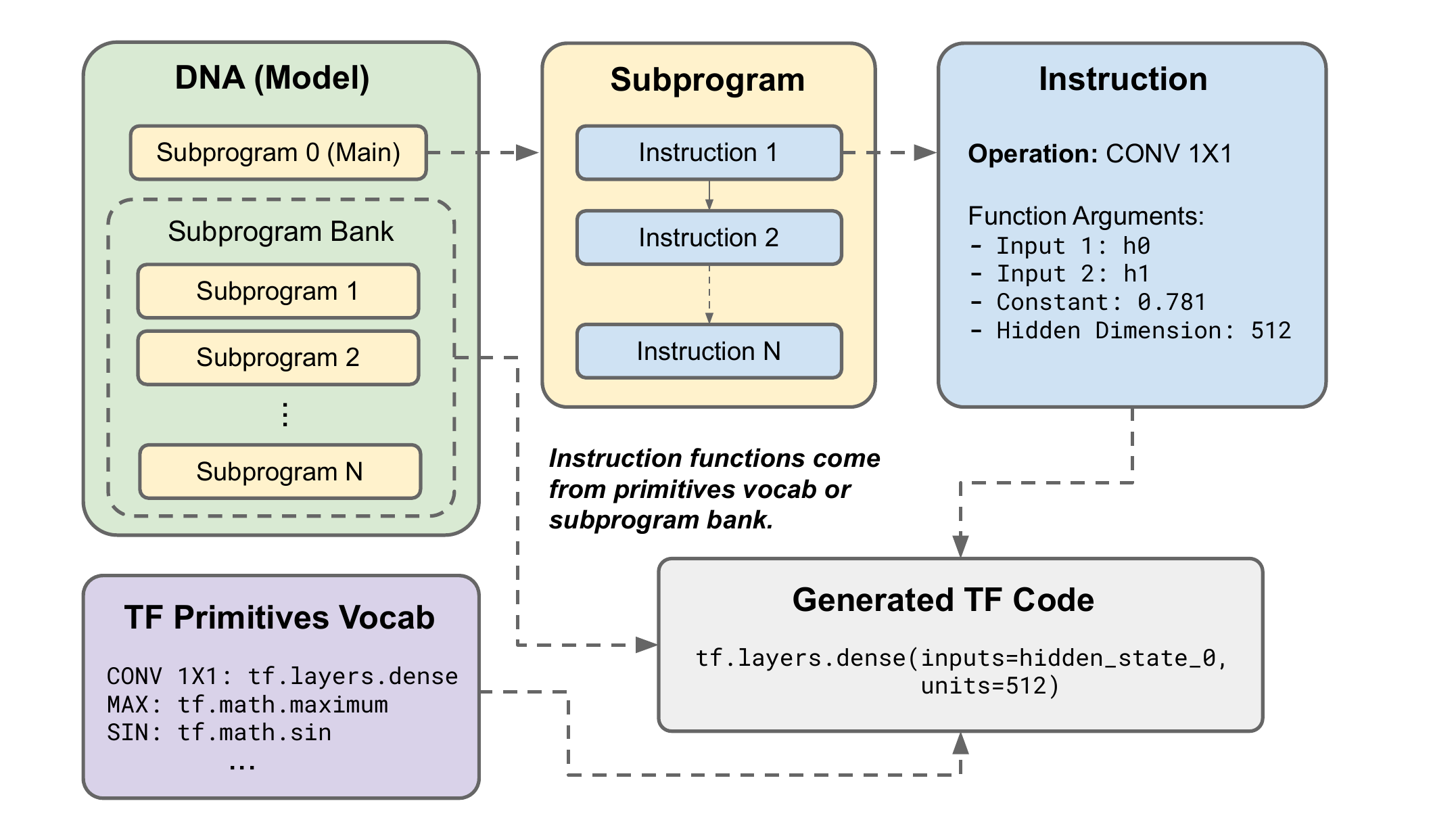}
\caption{Overview of DNAs that define a decoder model program (i.e., an auto-regressive language model). Each DNA has a collection of subprograms, where \textsc{subprogram 0} is the \textsc{main()} function entry point. Each subprogram is comprised of instructions, which are converted to lines of TensorFlow code. Instruction operations map to either basic TensorFlow library functions from the primitives vocabulary or one of the parent DNA's subprograms. The operation's arguments are filled using the parent instruction's argument set, which contains values for all potential operation arguments; arguments that are not used by a particular operation are simply ignored.}
\label{fig:dna_depiction}
\end{figure}

Figure~\ref{fig:dna_depiction} shows how programs are constructed in our search space. Each program is built from an evolutionary search \textit{DNA}, which is an indexed collection of \textit{subprograms}. \textsc{subprogram 0} is the \textsc{main()} function that is the execution entry point, and the other subprograms are part of the DNA's \textit{subprogram bank}. Each subprogram is an indexed array of \textit{instructions} with no length constraints. An instruction is an \textit{operation} with a set of input \textit{arguments}. The operation denotes the function that the instruction executes. Each operation maps to either a TF function from the \textit{primitives vocabulary} or another subprogram in the DNA subprogram bank. The primitives vocabulary is comprised of simple primitive TF functions, such as \textsc{add}, \textsc{log}, and \textsc{matmul} (see Appendix~\ref{sect:app_tf_vocab} for details). It is worth emphasizing that high-level building blocks such as self-attention are not operations in the search space, but can be constructed from our low-level operations. The DNA's subprogram bank is comprised of additional programs that can be executed as functions by instructions. Each subprogram can only call subprograms with a higher index in the subprogram bank, which removes the possibility of cycles.\looseness=-1

Each instruction's argument set contains a list of potential argument values for each instruction operation. The set of argument fields represents the union of fields that all the operation primitives use:\looseness=-1

\begin{itemize}%
  \item \textit{Input 1}: The index of the hidden state that will be used as the first tensor input. The index of each hidden state is the index of the instruction that produced it, with the subprogram's input states at indexes 0 and 1. An example of an operation that uses this is \textsc{sin}.\looseness=-1
  \item \textit{Input 2}: The index of the second tensor input. This is only used by operations that are binary with respect to tensor inputs. An example of an operation that uses this is \textsc{add}.
  \item \textit{Constant}: A real valued constant. An example of an operation that uses this is \textsc{max}; \textit{tf.math.maximum(x, C)} for $C=0$ is how we express the Transformer's ReLU activation.
  \item \textit{Dimension Size}: An integer representing the output dimension size for transformations that utilize weight matrices. An example of an operation that uses this is \textsc{conv 1x1}, the dense projection used by the Transformer's attention projections and feed forward portions. See Appendix~\ref{sect:app_tf_graphs} for how we employ relative dimensions \cite{so2019evolved} to resize our models.
\end{itemize}

Our search subprograms are converted to TF programs by converting each subprogram instruction to a corresponding line of TF code, one at a time in indexing order. To create the TF line, the instruction operation is mapped to the corresponding TF primitive function or DNA subprogram, and any relevant arguments are plugged in (see Appendix~\ref{sect:app_tf_vocab} for the full TF primitives vocabulary, including argument mappings); the other arguments are ignored. The TF tensor that is generated by the final instruction is taken as the subprogram output. We do not use TF Eager and so a useful property of the constructed programs is that irrelevant nodes that do not contribute to the programs' outputs are ignored as per TF's original deferred execution design \cite{Abadi2016TensorFlowAS}. See Figure~\ref{fig:program_to_graph} for an illustration of how subprograms are converted to TF graphs and see Appendix~\ref{sect:app_tf_graphs} for more details on how TF graphs are constructed, including how we handle causal masking.

\begin{figure}[h!]
\ffigbox{
  \centering
  \includegraphics[width=0.92\linewidth]{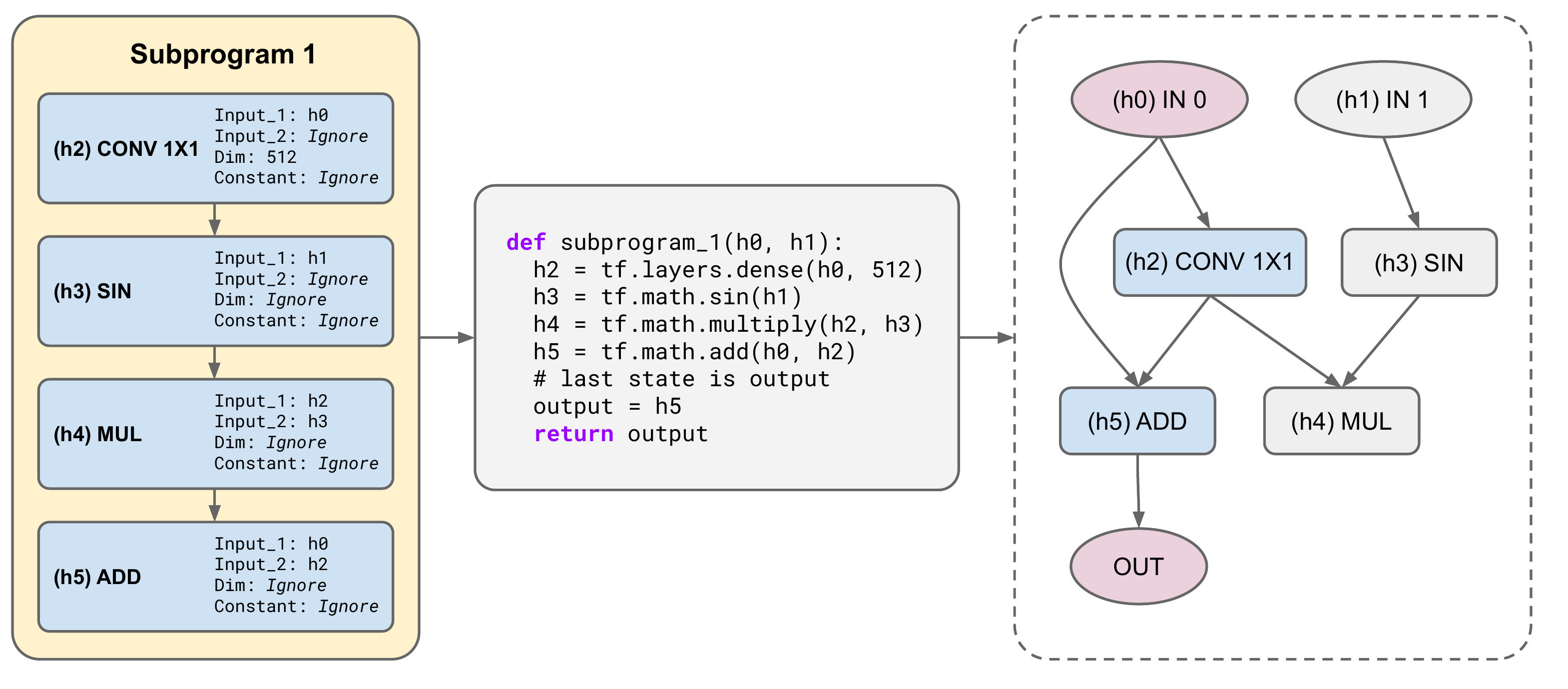}
}{
  \caption{Example of a program converted into its corresponding TensorFlow graph. Nodes that do not contribute to the program output are not executed thanks to TensorFlow's deferred execution design.\looseness=-1}
  \label{fig:program_to_graph}
}
\end{figure}

\paragraph{Evolutionary Search:} The goal of our evolutionary search is to find the most training efficient architecture in the search space. To do this, we give each model a fixed training budget (24 TPUv2 hours) and define its fitness as its perplexity on the One Billion Words Benchmark (LM1B) \cite{chelba2014billion} in Tensor2Tensor~\cite{tensor2tensor}. This approach, which we call an~\textit{\objectivename} by fixed training budget, contrasts previous architecture search works that explicitly aim to reduce training or inference step time when optimizing for efficiency~\cite{tan2019efficientnet, Tan2019MnasNetPN, Cai2019ProxylessNASDN, Elsken2019EfficientMN}. Our objective is different in that the trade-off between step time and sample efficiency is implicit. For instance, a modification that doubles step time, but triples sample efficiency is a good modification in our search, as it ultimately makes the architecture more compute efficient. Indeed, the modifications we find to be most beneficial, squaring ReLUs and adding depthwise convolutions to attention, increase training step time. However, they improve the sample efficiency of the model so much that they decrease the total compute needed to reach a target quality, by drastically reducing the number of training steps needed to get there.\looseness=-1

The search algorithm we use is Regularized Evolution \cite{Real2019RegularizedEF} with hurdles \cite{so2019evolved}. We configure our hurdles using a $50^{th}$ percentile passing bar and space them such that equal compute is invested in each hurdle band; this reduces the search cost by a factor of 6.25X compared to the same experiment with full model evaluations (see Appendix~\ref{sect:app_halving_hurdles} for more details). Additionally, we use 7 training hours as a proxy for a full day's training because a vanilla Transformer comes within 90\% of its 24 hour training perplexity with just 7 hours of training. This reduces the search cost further by a  factor of 3.43X, for a total compute reduction factor of 21.43X. So, although our target is to improve 24 hour performance, it only takes about 1.1 hours to evaluate an individual on average (see Appendix~\ref{sect:app_evolution_details} for more search specifics, including mutation details and hyperparameters).
We run our search for $\sim$25K individuals and retrain the top 100 individuals on the search task to select the best one.

\begin{wrapfigure}{r}{0.325\linewidth}
\centering
\includegraphics[width=1.0\linewidth]{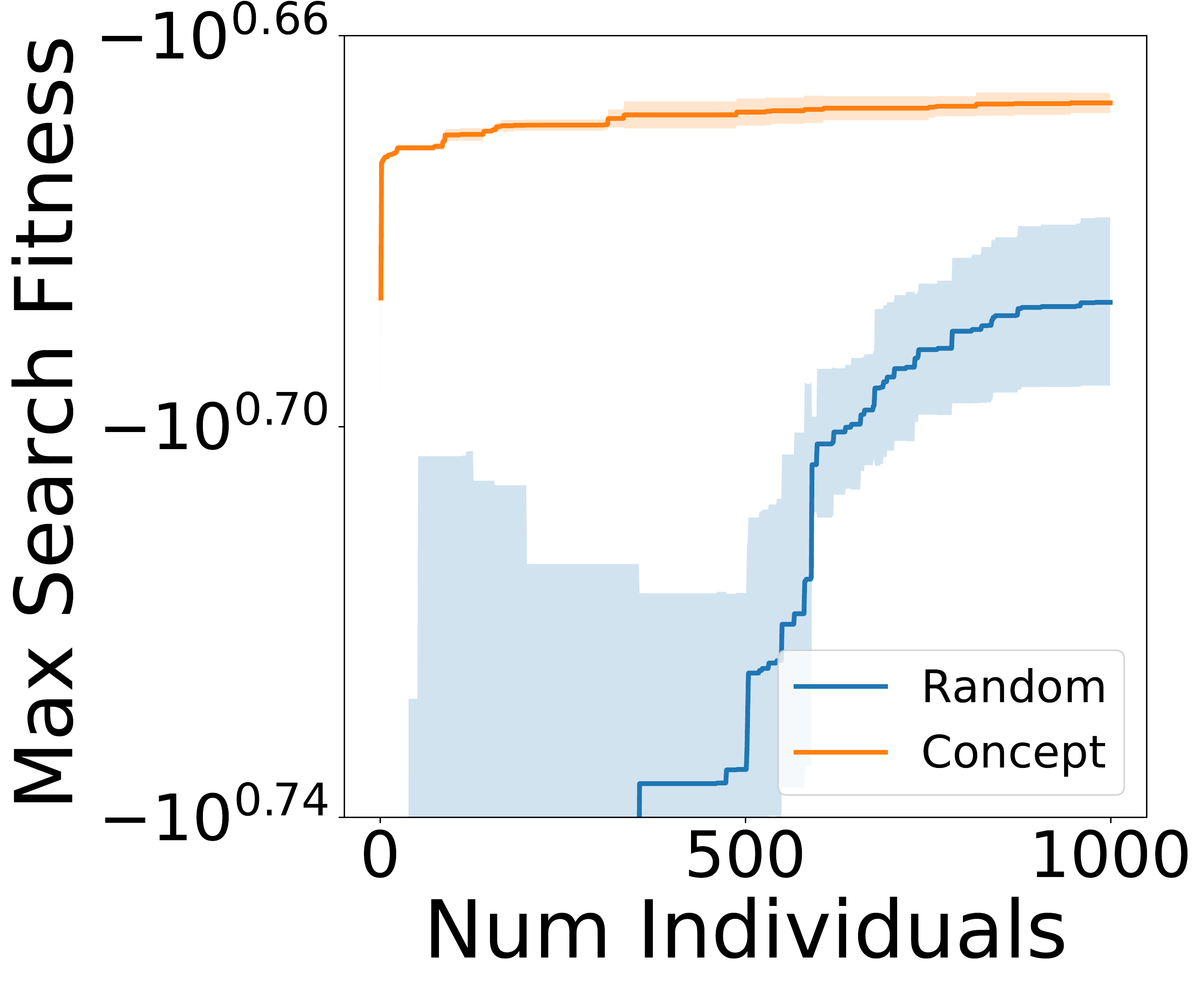}
\caption{Small scale searches (10 hours on 10 TPUv2 chips) comparing conceptual initialization to random initialization. Our search space is open-ended enough that it is infeasible to search without strong initialization.}
\label{fig:rand_vs_concept}
\end{wrapfigure}

Our search space is different from previous  search spaces (see architecture search survey by \cite{Elsken2019NeuralAS}), which are often heavily biased such that random search performs well (see analysis by \cite{Li2019RandomSA, Sciuto2020EvaluatingTS, Bender2020CanWS}). As our search space does not have this bias, 78\% of random programs in our space with length equal to a Transformer program cannot train more than five minutes, due to numerical instability. Because of this open-endedness and abundance of degenerate programs, it is necessary to initialize the search population with copies of the Transformer~\cite{so2019evolved} (input embedding size $d_{model}=512$, feed forward upwards projection size $d_{ff}=2048$, and number of layers $L=6$) (Figure~\ref{fig:rand_vs_concept}). To apply this initialization to our search space, we must determine how to divide the Transformer program into subprograms. To do this, we divide along the lines of the machine learning concepts that constitute it. For instance, we create one subprogram each for self-attention, ReLU and layer norm, using commonly used implementations (see Appendix~\ref{sect:app_search_model_subprograms} for the complete list). We call this method \textit{conceptual initialization} because it introduces a bias to the search through initialization, while leaving the search space for evolution and the action space for mutations open-ended. This contrasts the large amount of previous works that introduce bias through the search space. Although some works have also explored searching spaces that are open-ended like ours on miniature tasks \cite{Real2020AutoMLZeroEM}, we demonstrate that our techniques can scale to full sized deep learning regimes (see Section~\ref{sec:results}).

\section{\modelname}
\label{sect:primer_model}

\paragraph{\modelname:}
We name the discovered model \textit{\modelname}, which stands for PRIMitives searched transformER (See Appendix Figure~\ref{fig:primer_and_transformer_full} for the full program). Primer shows significant improvement when retrained on the search task, requiring less than half the compute of Transformer to reach the same quality (Figure~\ref{fig:search_task_bars}). In   Section~\ref{sec:results}, we additionally show that \modelname makes equally large gains when transferred to other codebases, training regimes, datasets, and downstream one-shot tasks.

\begin{figure}[h!]
  \centering
\begin{minipage}{0.69\linewidth}
    \centering
  \includegraphics[width=0.98\linewidth]{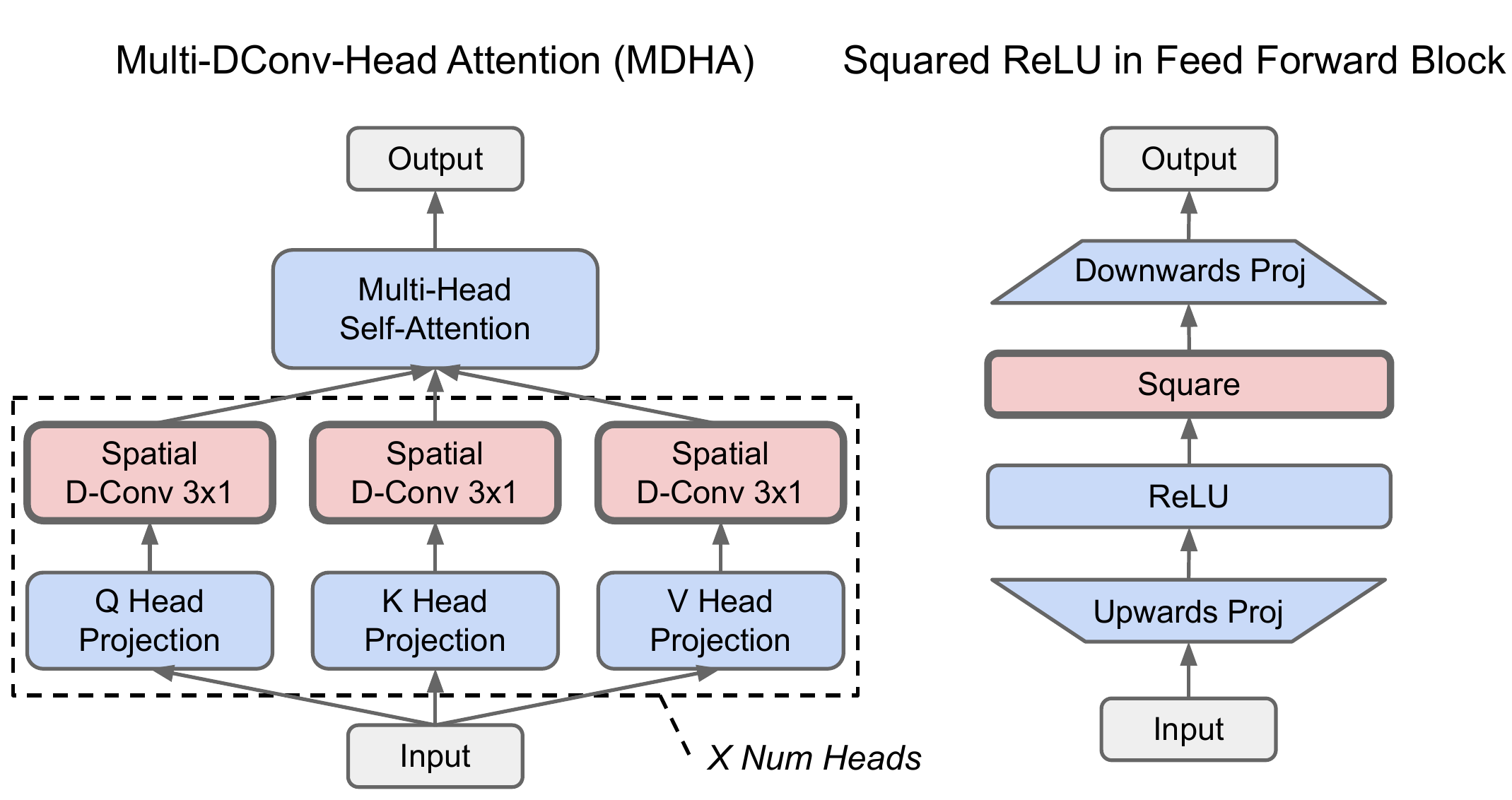}
  \end{minipage}
  \begin{minipage}{0.30\linewidth}
  \centering
  \small{MDHA Projection Pseudo-code}
    \begin{lstlisting}[
  language=python]
# Use to create each K, Q, and V head of size `hs'.
def mdha_projection(x, hs):
  # Create head.
  x = proj(x,
            head_size=hs,
            axis="channel")

  # Apply D-Conv to head.    
  x = d_conv(x,
              width=3,
              head_size=hs,
              axis="spatial",
              mask="causal")
  return x
\end{lstlisting}
    \end{minipage}
  \caption{The two main modifications that give \modelname most of its gains: depthwise convolution added to attention multi-head projections and squared ReLU activations. These modifications are easy to implement and transfer well across codebases. We call the model with just these two modifications \modelnamelite. Blue indicates portions of the original Transformer and red signifies one of our proposed modifications.\looseness=-1}
  \label{fig:efficientseq_lite_model}
\end{figure}

\paragraph{\modelnamelite:}
A core motivation of this work is to develop simple techniques that can be easily adopted by language modeling practitioners. To accomplish this, we perform ablation tests across two codebases (T5~\cite{2020t5} and Tensor2Tensor~\cite{tensor2tensor}) and determine which \modelname modifications are generally useful (Appendix Figure~\ref{fig:ablation_study}). The two that produce the most robust improvements are squaring feed forward ReLUs and adding depthwise convolution to attention multi-head projections (Figure~\ref{fig:efficientseq_lite_model}). We refer to a Transformer with just these two easy modifications as \textit{\modelnamelite}; this is our recommended starting point for language modeling practitioners interested in using \modelname. We now explain these modifications and then measure their empirical effectiveness. 

\begin{figure}[h!]
  \centering
  \begin{minipage}{0.32\linewidth}
    \centering
    \includegraphics[width=0.78\linewidth]{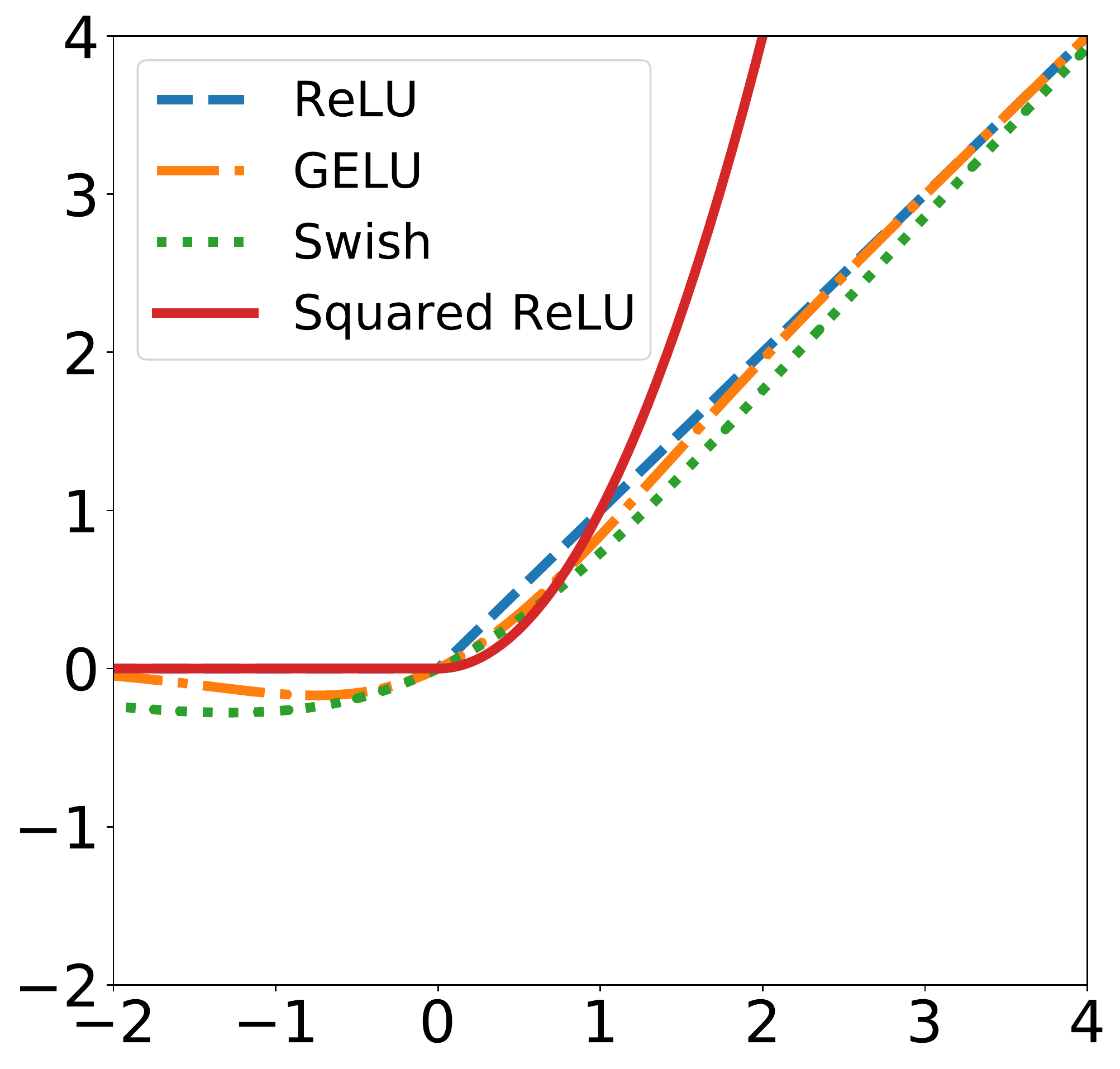}
  \end{minipage}
    \begin{minipage}{0.32\linewidth}
    \centering
    \includegraphics[width=0.78\linewidth]{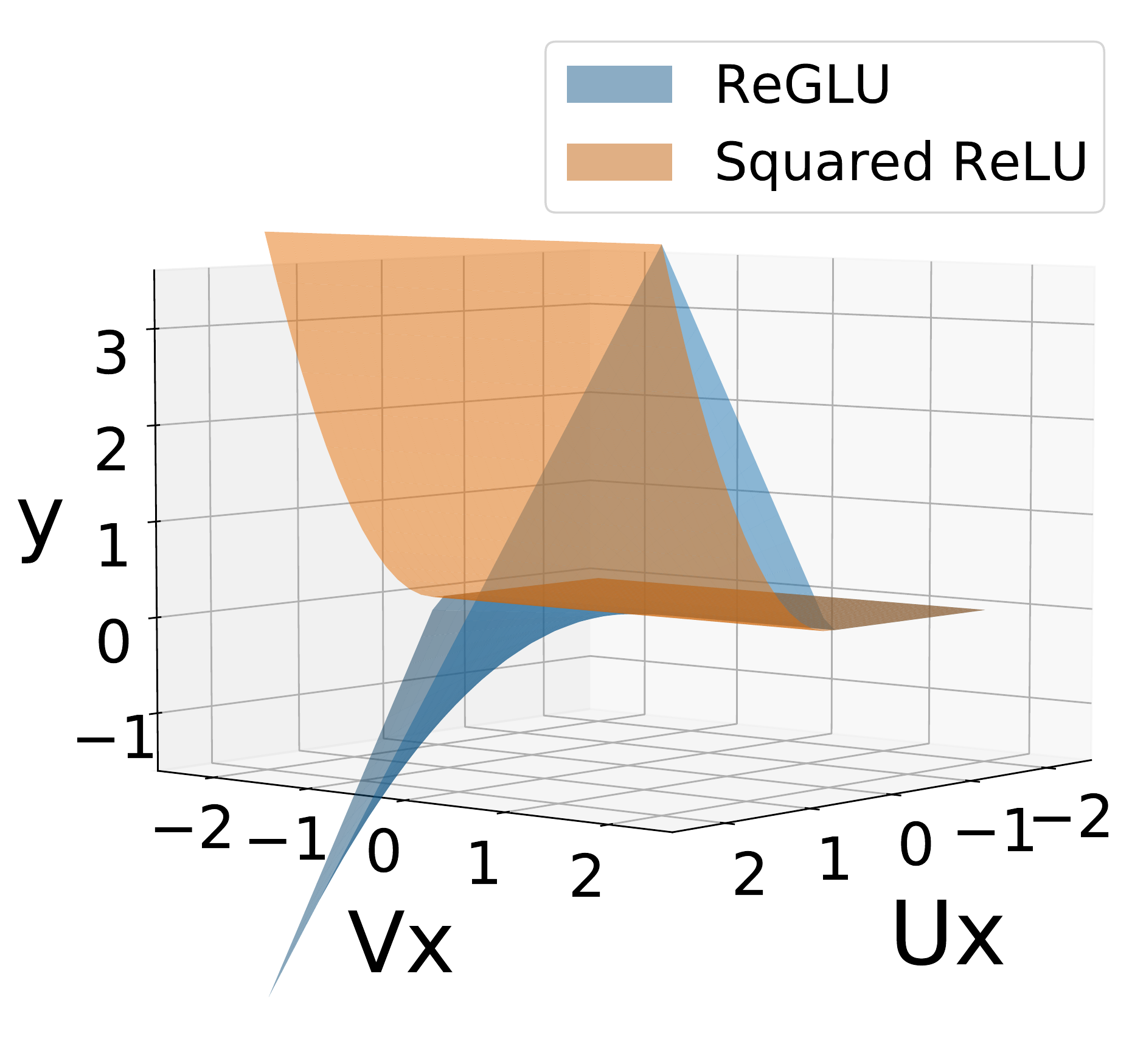}
  \end{minipage}
    \begin{minipage}{0.32\linewidth}
    \centering
    \includegraphics[width=0.78\linewidth]{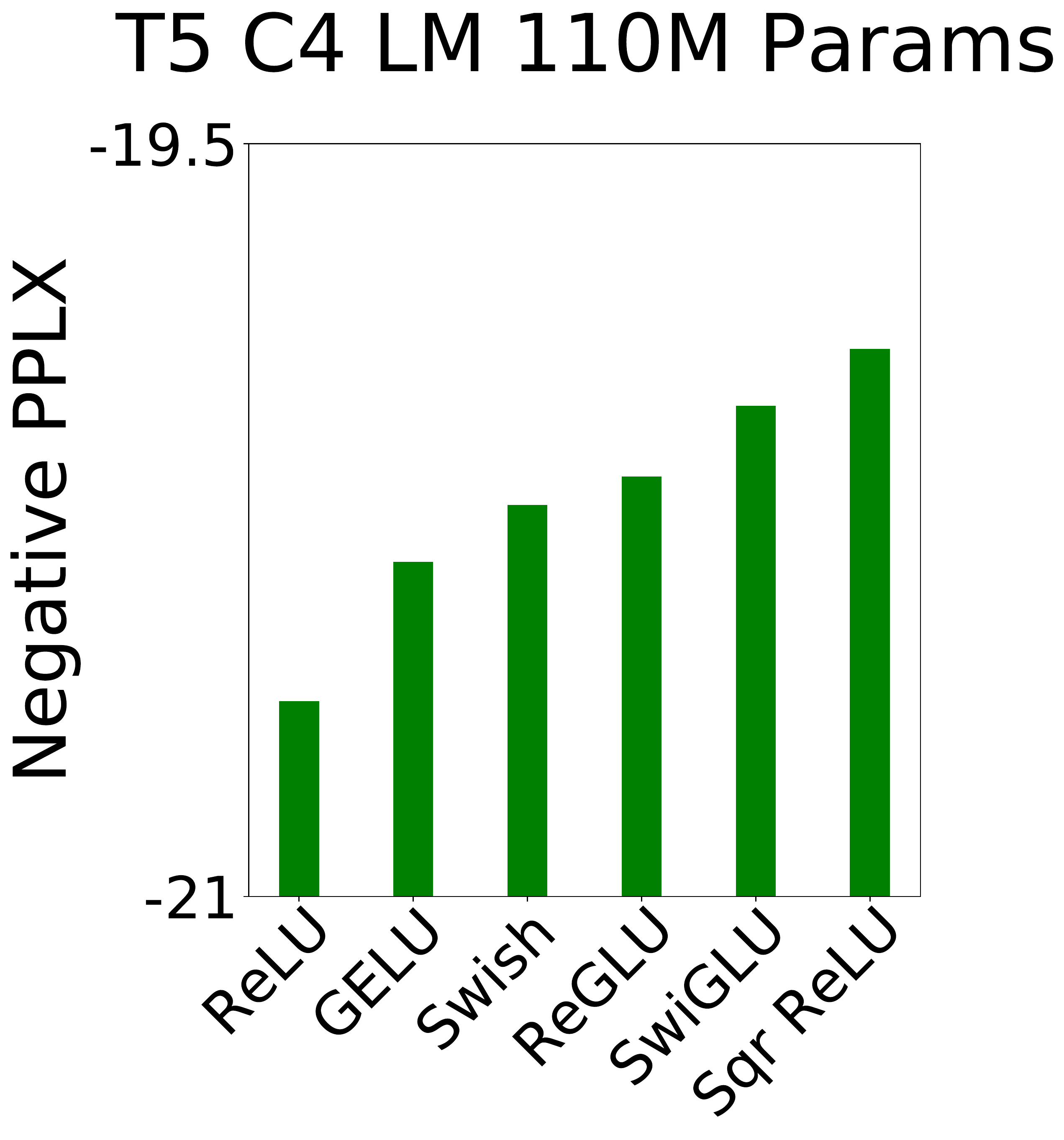}
  \end{minipage}
  \caption{Left: Squared ReLU has starkly different asymptotics compared to other common activation functions. Center: Squared ReLU has significant overlap with GLU variants~\cite{shazeer2020glu} that use activations with ReLU-like asymptotics, such as ReGLU and SwiGLU. Our experiments indicate that squared ReLU is better than these GLU variants in Transformer language models. Right: Comparison of different nonlinearities in Transformers trained on C4 auto-regressive LM for 525K steps.}
  \label{fig:activation_function_graph}
\end{figure}

\paragraph{Squared ReLU:} The most effective modification is the improvement from a ReLU activation to a squared ReLU activation in the Transformer's feed forward block. Rectified polynomials of varying degrees have been studied in the context of neural network activation functions~\cite{Krotov2016DenseAM}, but are not commonly used; to the best of our knowledge, this is the first time such rectified polynomial activations are demonstrated to be useful in Transformers. Interestingly, the effectiveness of higher order polynomials~\cite{Jayakumar2020MultiplicativeI} can also be observed in other effective Transformer nonlinearities, such as GLU~\cite{dauphin2017language} variants like ReGLU~\cite{shazeer2020glu} ($y = Ux \odot \max(Vx, 0)$ where $\odot$ is an  element-wise product) and point-wise activations like approximate GELU~\cite{Hendrycks2016BridgingNA} ($y = 0.5x(1 + \tanh(\sqrt{2 / \pi}(x + 0.044715x^3)))$). However, squared ReLU has drastically different asymptotics as $x \xrightarrow{} \infty$ compared to the most commonly used activation functions: ReLU, GELU and Swish (Figure~\ref{fig:activation_function_graph} left side). Squared ReLU does have significant overlap with ReGLU and in fact is equivalent when ReGLU's $U$ and $V$ weight matrices are the same and squared ReLU is immediately preceded by a linear transformation with weight matrix $U$. This leads us to believe that squared ReLUs capture the benefits of these GLU variants, while being simpler, without additional parameters, and delivering better quality (Figure~\ref{fig:activation_function_graph} right side).\looseness=-1

\paragraph{Multi-DConv-Head Attention (MDHA):} Another effective modification is adding 3x1 depthwise convolutions after each of the multi-head projections for query $Q$, key $K$ and value $V$ in self-attention. These depthwise convolutions are performed over the spatial dimension of each dense projection's output. Interestingly, this ordering of pointwise followed by depthwise convolution is the reverse of typical separable convolution, which we find to be less effective in Appendix~\ref{sect:app_lm1b_t2t}. We also find that wider depthwise convolution and standard convolution not only do not improve performance, but in several cases hurt it. Although depthwise convolutions have been used for Transformers before~\cite{wei2018qanet,gulati2020conformer}, using them after each dense head projection has not been done to the best of our knowledge. MDHA is similar to Convolutional Attention~\cite{Wu2021CvTIC}, which uses separable convolution instead of depthwise convolution and does not apply convolution operations per attention head as we do.\looseness=-1

\paragraph{Other Modifications:} The other \modelname modifications are less effective. Graphs for each modification can be found in Appendix \ref{sect:app_search_model_subprograms} and an ablation study can be found in Appendix~\ref{sec:ablation_and_insertion}. We briefly describe the modifications and their usefulnesses here:

\begin{itemize}%
  \item \textit{Shared Q and K Depthwise Representation}: \modelname shares some weight matrices for $Q$ and $K$. $K$ is created using the previously described MDHA projection and $Q=KW$ for learnable weight matrix $W \in \mathbb{R}^{d \times d}$. We find that this generally hurts performance.\looseness=-1
  \item \textit{Pre and Post Normalization}: The standard practice for Transformers has become putting normalization before both the self-attention and feed forward transformations~\cite{Baevski2019AdaptiveIR, Xiong2020OnLN}. \modelname uses normalization before self-attention but applies the second normalization \textit{after} the feed forward transformation. We find this is helpful in some but not all cases.\looseness=-1
  \item \textit{Custom Normalization}: \modelname uses a modified version of layer normalization~\cite{Ba2016LayerN} that uses $x(x - \mu)$ instead of $(x-\mu)^2$, but we find this is not always effective.
  \item \textit{12X Bottleneck Projection}: The discovered model uses a smaller $d_{model}$ size of 384 (compared to the baseline's 512) and a larger $d_{ff}$ size of 4608 (compared to the baseline's 2048). We find this larger projection improves results dramatically at smaller sizes ($\sim$35M parameters), but is less effective for larger models, as has been previously noted~\cite{Kaplan2020ScalingLF}. For this reason we do not include this modification when referencing \modelname or \modelnamelite.
  \item \textit{Post-Softmax Spatial Gating}: The  discovered model has a set of per-channel learnable scalars after the attention softmax, which improves perplexity for fixed length sequences. However, these scalars cannot be applied to variable sequence lengths and so we do not include this modification in \modelname for our experiments.
  \item \textit{Extraneous Modifications}: There are a handful of additional modifications that produce no meaningful difference in the discovered architecture. For example, hidden states being multiplied by -1.12. Verifying that these modifications neither help nor hurt quality, we exclude them from discussion in the main text and do not include them when experimenting with \modelname. These extraneous modifications can still be found in Appendix~\ref{sect:app_search_model_subprograms}. \looseness=-1
  
\end{itemize}

\section{Results}
\label{sec:results}

In our experiments, we compare \modelname against three Transformer variants:

\begin{itemize}[leftmargin=*]
  \item \textit{Vanilla Transformer}: The original Transformer~\cite{Vaswani2017AttentionIA} with ReLU activations and layer normalization~\cite{Ba2016LayerN} outside of the residual path.
  \item \textit{Transformer+GELU}: A commonly used variant of the vanilla Transformer that uses a GELU~\cite{Hendrycks2016BridgingNA} approximation activation function~\cite{devlin2018bert, brown2020language}.
  \item \textit{Transformer++}: A Transformer with the following enhancements: RMS normalization~\cite{Zhang2019RootMS}, Swish activation~\cite{Ramachandran2018SearchingFA} and a GLU multiplicative branch~\cite{dauphin2017language} in the feed forward inverted bottleneck (SwiGLU)~\cite{shazeer2020glu}.
  These modifications were benchmarked and shown to be effective in T5~\cite{narang2021}.
\end{itemize}

We conduct our comparisons across three different codebases: Tensor2Tensor (T2T)~\cite{tensor2tensor},  T5~\cite{2020t5}, and Lingvo~\cite{Shen2019LingvoAM}. Tensor2Tensor is the codebase we use for searching and so a majority of our side-by-sides are done in T5 and Lingvo to prove transferability. In all cases, we use the default Transformer hyperparameters for each codebase, with regularization disabled. See Appendix~\ref{sect:app_full_training_details} for more hyperparameter details.\looseness=-1 

In the following sections, we will present our results in four main experiments on auto-regressive language modeling. First, we will show that \modelname outperforms the baseline models on the search task. Next, we will show that the relationship between \modelname's compute savings over Transformers and model quality follow a power law at optimal model sizes. These savings also transfer across datasets and codebases. Then, we will study \modelname's gains in an established training regime and show that it enables 4.2X compute savings at a 500M parameter size using full compute T5 training. Finally, we will demonstrate that these gains transfer to the pretraining and one-shot downstream task setup established by GPT-3~\cite{brown2020language}.\looseness=-1 

\subsection{Search Task Comparison}
\label{sect:search_task_comparison}

\begin{figure}[h!]
  \centering
  \begin{minipage}{0.1\linewidth}
    \centering
  \end{minipage}
  \begin{minipage}{0.25\linewidth}
    \centering
    \includegraphics[width=0.91\linewidth]{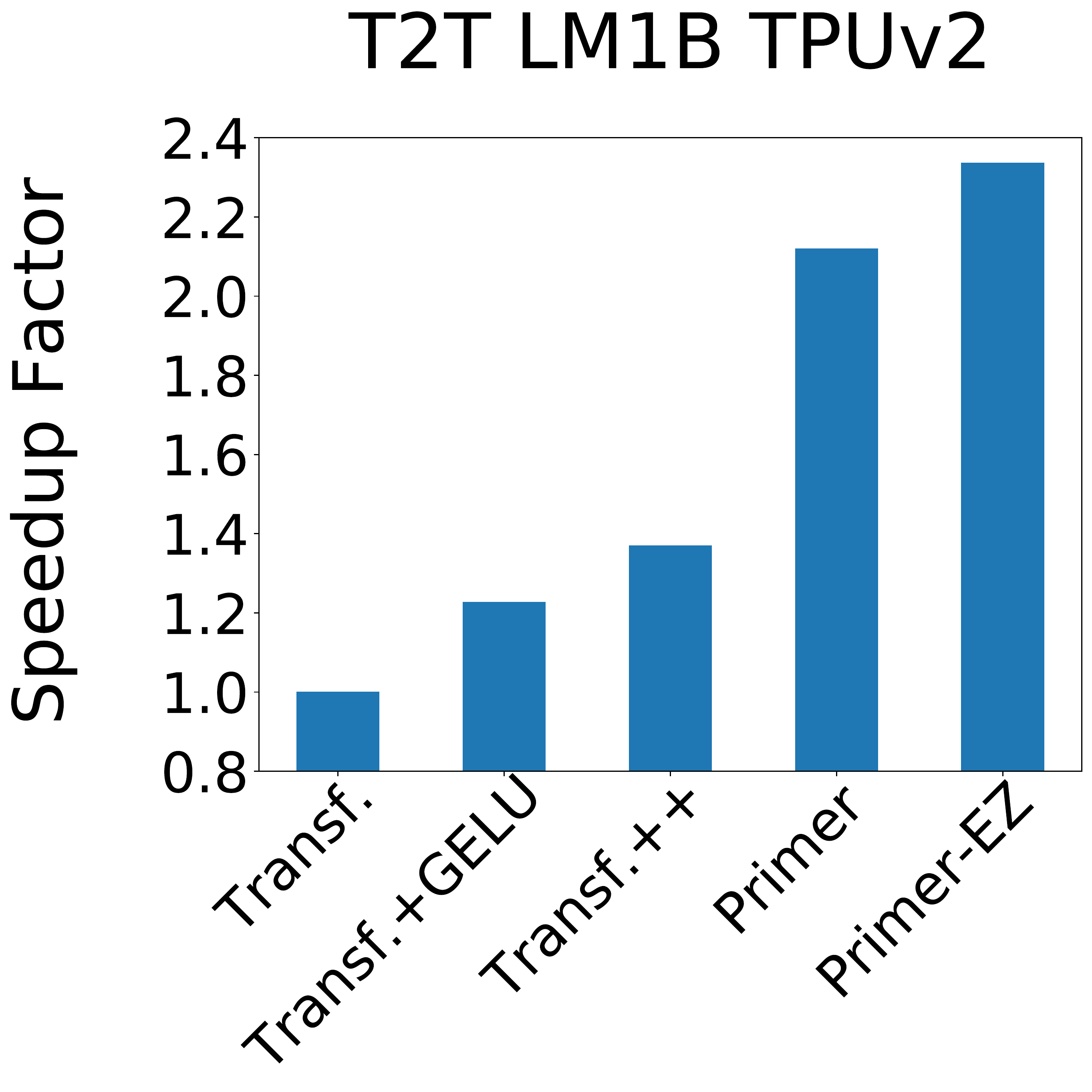}
  \end{minipage}
\begin{minipage}{0.25\linewidth}
    \centering
    \includegraphics[width=0.75\linewidth]{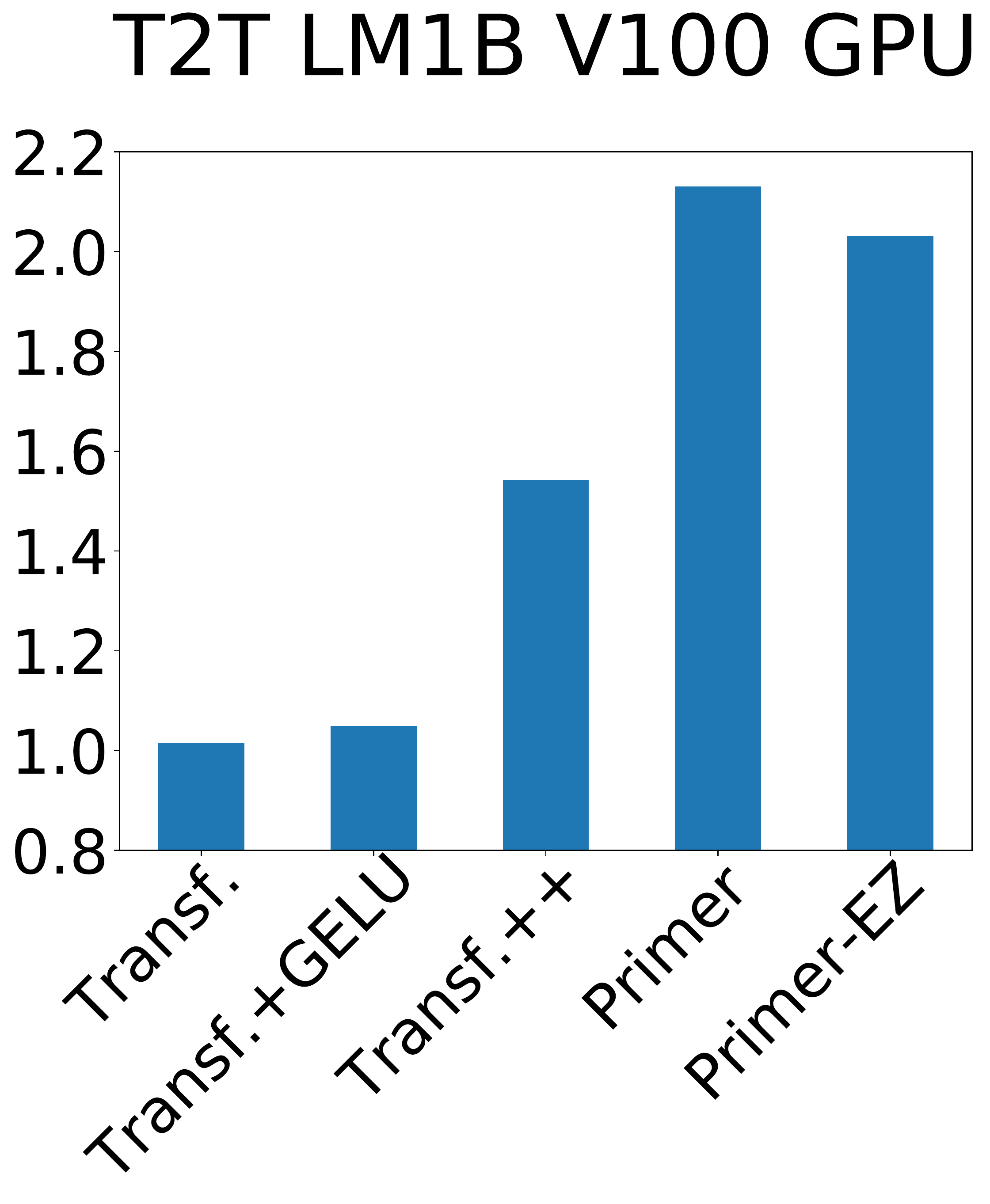}
  \end{minipage}
  \begin{minipage}{0.21\linewidth}
    \centering
    \includegraphics[width=0.91\linewidth]{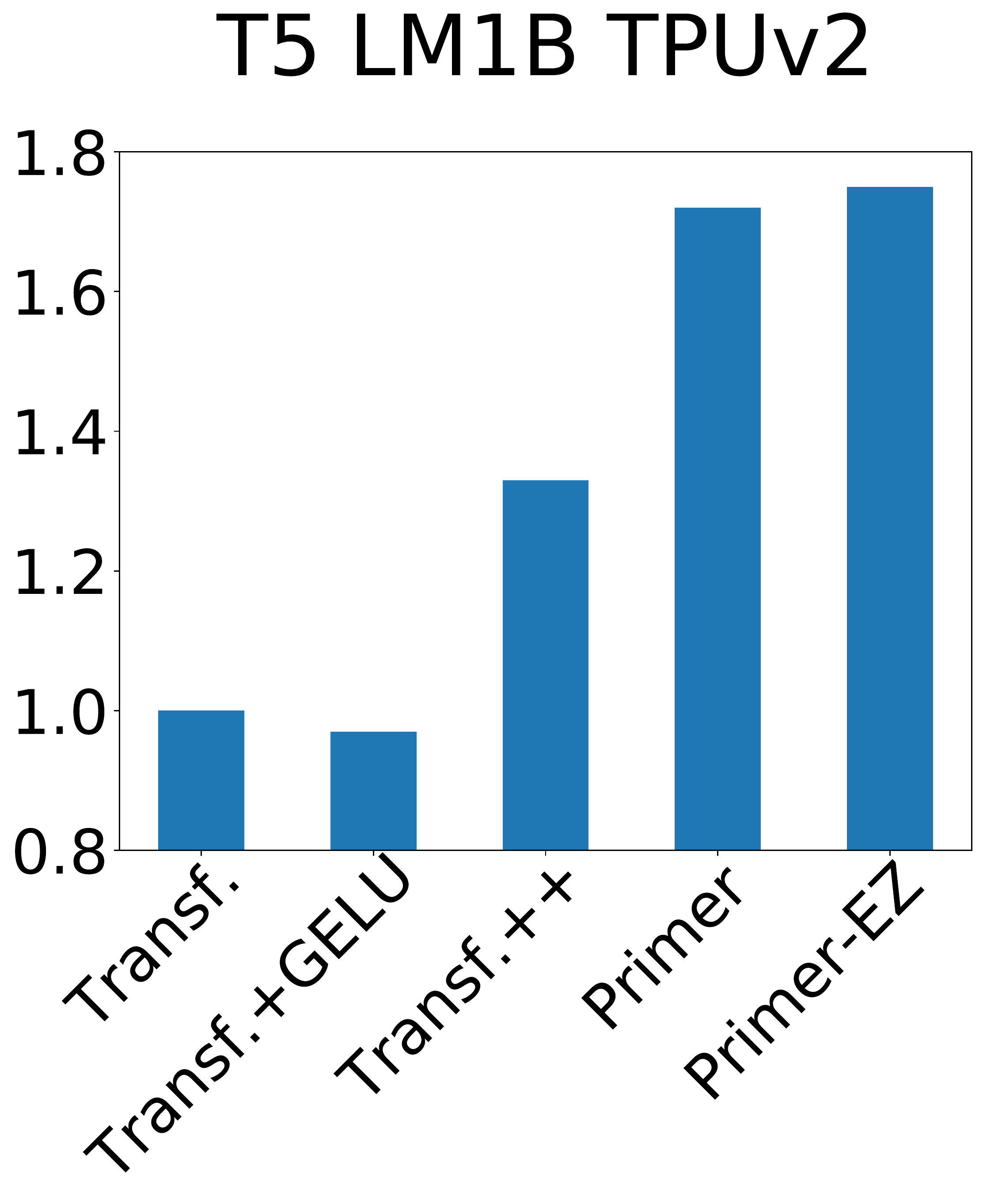}
  \end{minipage}
  \begin{minipage}{0.14\linewidth}
    \centering
  \end{minipage}
  \caption{Comparison on the LM1B search task using 35M parameter models. ``Speedup Factor'' refers to the fraction of compute each model needs to reach quality parity with the vanilla Transformer trained for 24 hours. \modelname and \modelnamelite both achieve over 1.7X speedup in all cases. Note that results transfer across codebases and different hardware.}
  \label{fig:search_task_bars}
\end{figure}

We first analyze \modelname's performance on the search task: LM1B language modeling with sequence length 64, $\sim$35M model parameters, batches of 4096 tokens and 24 hours of training. We compare against the baseline models in both Tensor2Tensor (T2T)~\cite{tensor2tensor} and T5~\cite{2020t5} and on TPUv2s and V100 GPUs. We grade each model's performance according to how much faster it reaches the vanilla Transformer's final quality, which we will refer to as its \textit{speedup factor}.  Figure~\ref{fig:search_task_bars} shows that \modelname provides a speedup factor of 1.7X or more over Transformer in all cases. Figure~\ref{fig:search_task_bars} also shows that both \modelname and \modelnamelite generalize to other hardware platforms and codebases.\looseness=-1

\begin{figure}[h!]
  \centering
  \begin{minipage}{0.45\linewidth}
    \centering
    \includegraphics[width=0.6\linewidth]{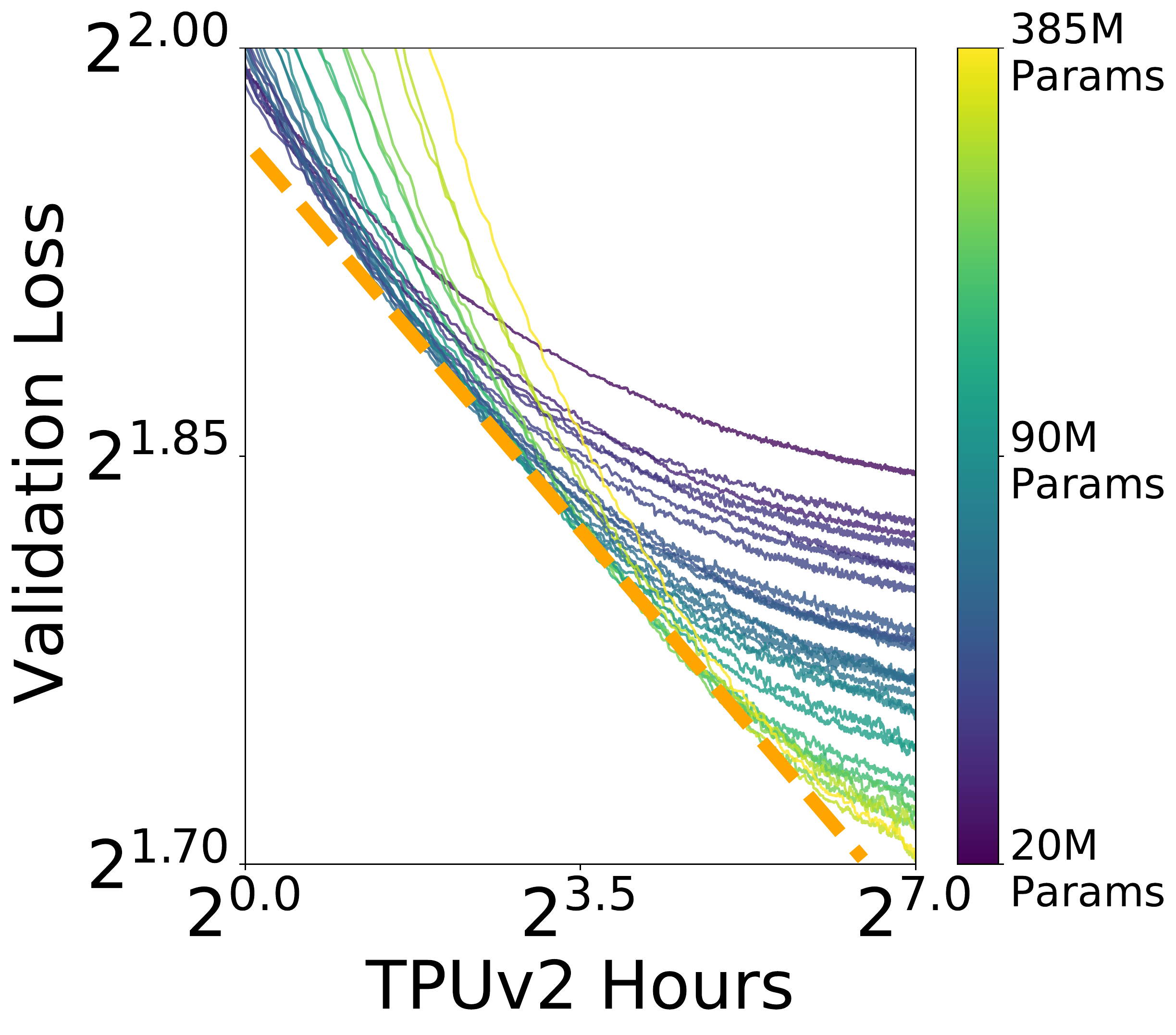}
  \end{minipage}
    \begin{minipage}{0.45\linewidth}
    \centering
    \includegraphics[width=0.6\linewidth]{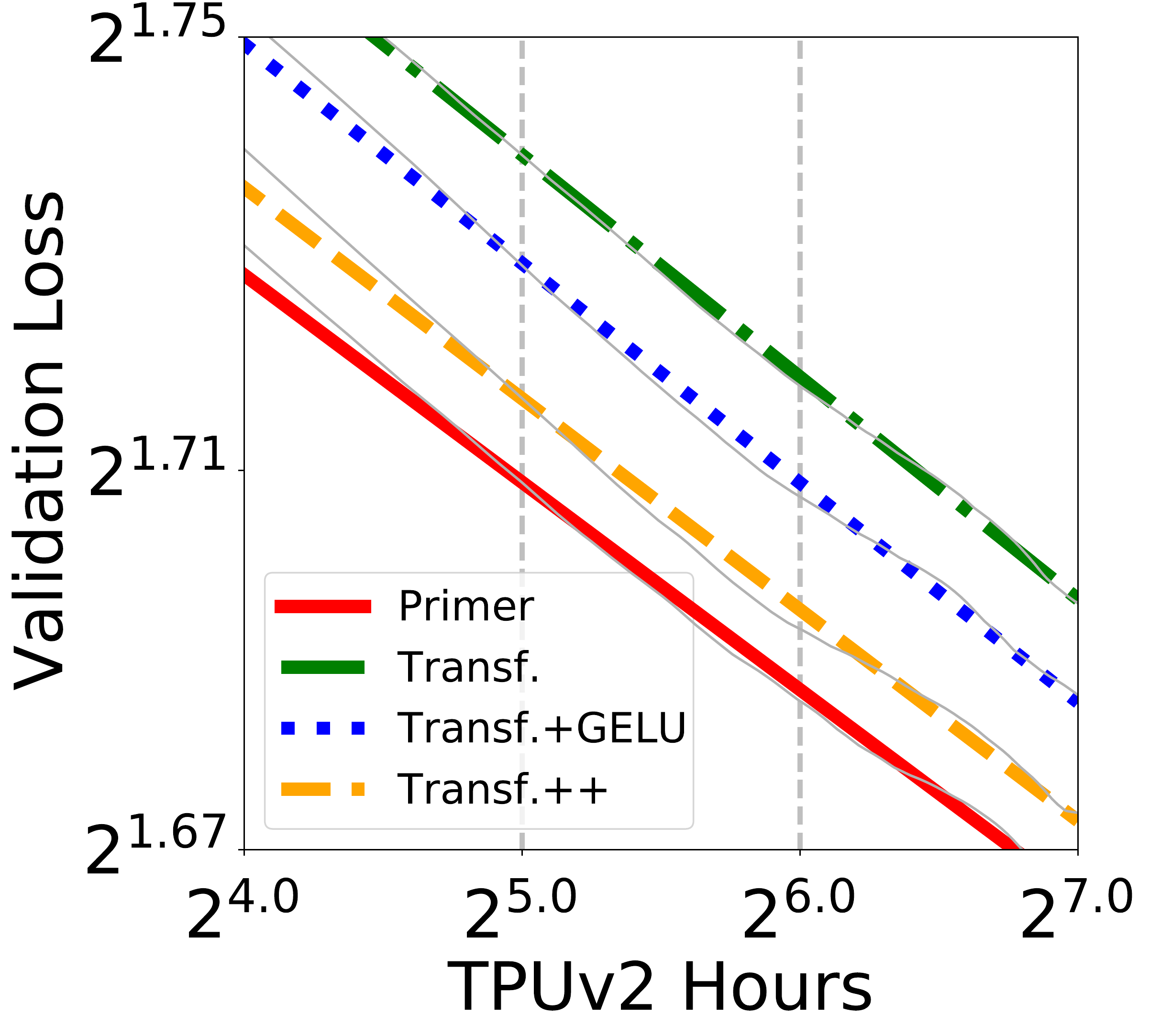}
  \end{minipage}
  \caption{Left: When sweeping over optimal model sizes, the relationship between Transformer language model quality and training compute roughly obeys a power law~\cite{Kaplan2020ScalingLF}. Right: Comparison of these power law lines for varying Transformer modifications, fit with smoothed MSE loss. That the model lines are parallel to one another implies that compute savings by using superior modeling also scales as a power law with quality. Spacings between vertical dotted lines represent 2X differences in compute. Note that the x and y-axes in both plots are in $\log$.
  }
  \label{fig:power_law_comparison}
\end{figure}

\begin{wrapfigure}{R}{0.28\linewidth}
\centering
\includegraphics[width=0.97\linewidth]{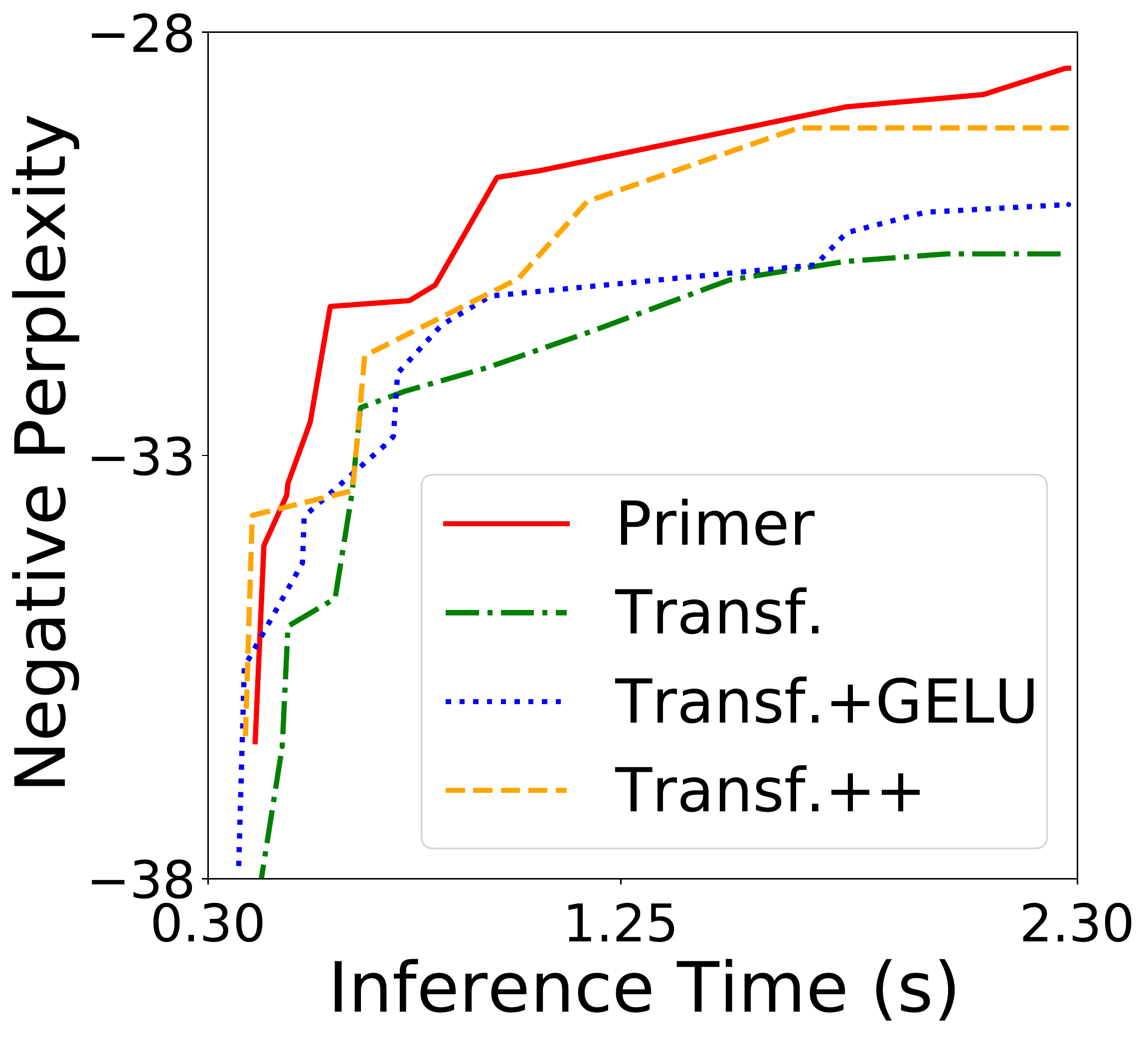}
\caption{Pareto optimal inference comparison on LM1B. \modelname~demonstrates improved inference at a majority of target qualities. We observe these models have a 0.97 correlation between their train step and inference times.}
\label{fig:pareto_inference}
\end{wrapfigure}

Next we study the scaling laws of \modelname. Here we compare \modelname to our baselines over many sizes by training each model using every permutation of $L \in \{6, 9, 12\}$ layers, $d_{model}\in \{384, 512, 1024\}$ initial embedding size, and $p \in \{4, 8, 12\}$ feed forward upwards projection ratio, creating a parameter range from 23M to 385M. The results, shown in Figure~\ref{fig:power_law_comparison}, corroborate previous claims that, at optimal parameters sizes, the relationship between compute and language model quality roughly follows a power law~\cite{Kaplan2020ScalingLF}. That is, the relationship between validation loss, $l$, and training compute, $c$, follows the relationship $l=ac^{-k}$, for empirical constants $a$ and $k$. This is represented as a line in double log space (Figure~\ref{fig:power_law_comparison}): $\log l = -k\log c + \log a$. However, these lines are not the same for each architecture. The lines are roughly parallel but shifted up and down. In Appendix~\ref{sect:app_power_law_derivations} we show that, given a vertical spacing of $\log b^k$, parallel lines such as these indicate compute savings, $s$, for superior modeling also follow a power law of the form $l = a_1(1 - 1/b)^ks^{-k}$. The intuition behind this is that ${b}$ is a constant compute reduction factor for all $l$ and thus a power law investment of training compute with relation to $l$ results in a power law savings with relation to $l$ as well (see Appendix~\ref{sect:app_power_law_derivations}).

\modelname also has the capacity to improve inference, despite our search focusing on training compute. Figure~\ref{fig:pareto_inference} shows a Pareto front comparison of quality vs. inference, when using feed forward pass timing as a proxy for inference. We use forward pass timing as a proxy for inference because there are multiple ways to decode a language model, each with varying compute costs. A more in depth study could be conducted analyzing \modelname's inference performance across different decoding methods, serving platforms, datasets, etc., but that is beyond the scope of this work.

\subsection{\modelname Transferability to Other Codebases, Datasets, and Model Types}
\label{sect:t5_medium_experiments}

\begin{figure}[h!]
  \centering

  \begin{minipage}{0.78\linewidth}
    \centering
    \includegraphics[width=0.85\linewidth]{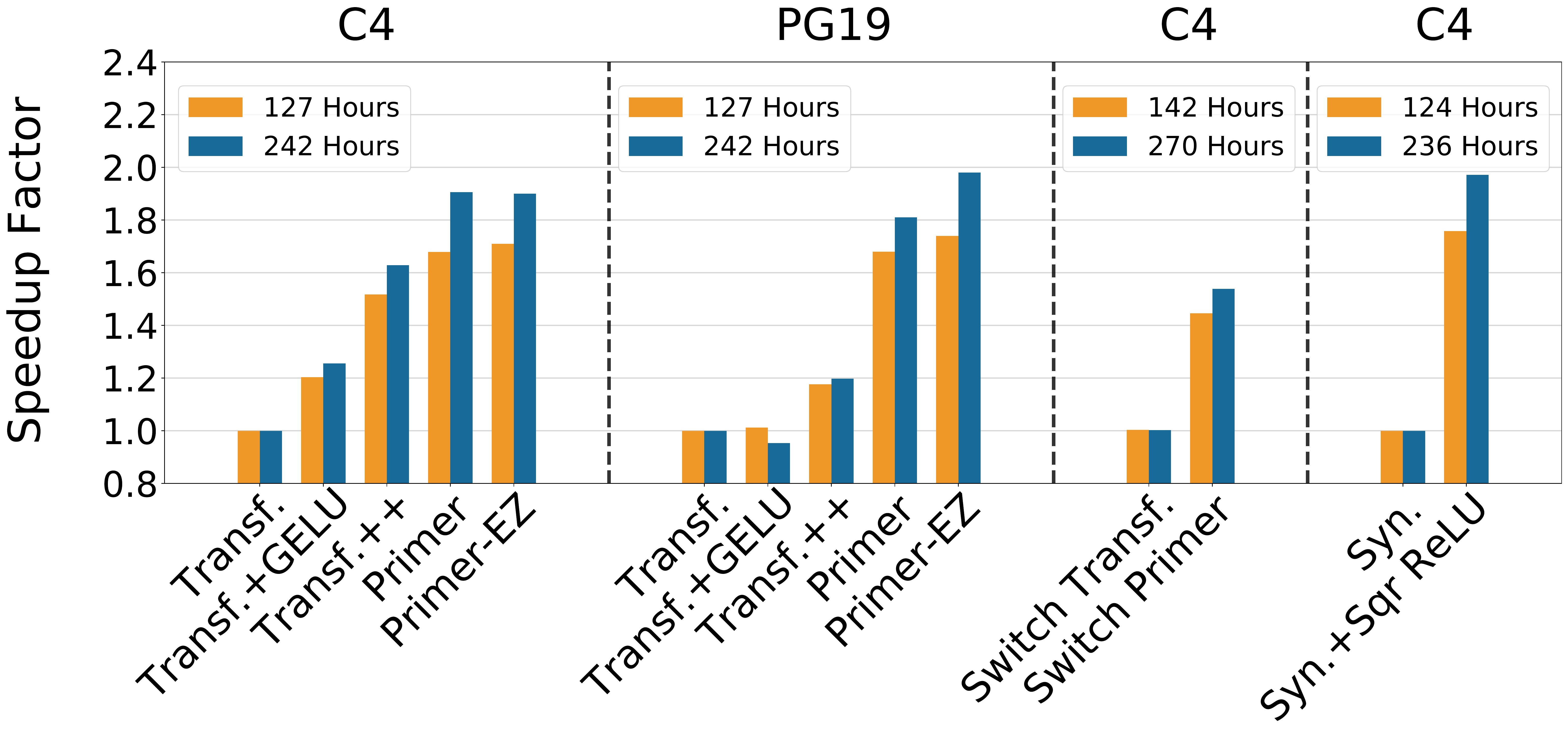}
  \end{minipage}
  \caption{Comparison transferring \modelname to larger datasets (C4 and PG19) and different model families (Switch Transformer and Synthesizer) in a different codebase (T5) with an order of magnitude more compute than the search task. Compared to the vanilla baseline, \modelname and \modelnamelite are at least 1.8X more efficient at the end of training on both PG19 and C4. In all cases, the fraction of \modelname compute savings increases as more compute is invested. \modelnamelite modifications also improve Switch Transformer (550M params) and Synthesizer (145M params), showing that it is compatible with other efficient methods. Compute budgets are selected according to how long it takes each baseline to train for 525K and 1M steps. See Appendix~\ref{sect:app_t5_medium_numbers} for exact numbers.}
  \label{fig:robustness}
\end{figure}

We now study \modelname's ability to transfer to larger datasets, PG19 and C4, in another codebase, T5. We additionally scale up to a higher compute regime that has been used as a proxy for large scale training by previous studies~\cite{narang2021, 2020t5}; the batches are increased to 65K tokens, the sequence lengths are a longer 512, each decoder is 110M parameters ($d_{model}=768$, $d_{ff}=3072$, $L=12$) and each model is trained to $\sim$525K steps on 4 TPUv3 chips. We also continue training each model to 1M steps to study the effect of larger compute budgets on \modelname savings. The results, shown in Figure~\ref{fig:robustness}, indicate that the \modelname models are as strong in larger data, higher compute regimes, as they are in the smaller LM1B regime. Compared to the vanilla baseline, \modelname and \modelnamelite are at least 1.8X more efficient at the end of training on both PG19 and C4.\looseness=-1

Figure~\ref{fig:robustness} also shows that the \modelname modifications are compatible with other efficient model families, such as large sparse mixture-of-experts like Switch Transformer~\cite{Fedus2021SwitchTS} and efficient Transformer approximations like Synthesizer~\cite{tay2020synthesizer}. For these experiments, we use the T5 implementations provided by Narang et al.~\cite{narang2021}. The \modelnamelite techniques of added depthwise convolutions and squared ReLUs reduce Switch Transformer's compute cost by a factor of 1.5X; this translates to a 0.6 perplexity improvement when controlling for compute (see Appendix~\ref{sect:app_t5_medium_numbers}). Adding squared ReLUs to Synthesizer reduces training costs by a factor of 2.0X and improves perplexity by 0.7 when fully trained.\looseness=-1

\subsection{Large Scale T5 Auto-Regressive Language Model Training}
\label{sect:t5_large_experiments}

\begin{wraptable}{R}{0.45\linewidth}
\begin{tabular}{@{}l|ccc@{}}
\toprule
Model        & Steps      & TPUv3 Hours     & PPLX      \\ \midrule
Original T5 & 1M &	15.7K & 13.25 \\
T5++ & 251K & 4.6K & 13.25 \\
\modelname & 207K & \textbf{3.8K} & 13.25 \\ \midrule
T5++  & 1M         & 16.5K           & 12.69     \\
\modelname & 480K      & \textbf{8.3K}            & 12.69     \\ \midrule
\modelname & 1M	& 17.3K & 12.35 \\ \midrule
\end{tabular}
  \caption{Comparison in compute usage to reach target qualities on C4 LM at 537M parameters using the full T5 compute scale. Target qualities are selected according to the final performances of the baseline models. \modelname achieves the same quality as the original T5 architecture using 4.2X less compute.
  }
  \label{tab:large_t5}
\end{wraptable}

In large scale compute configurations, the \modelname compute savings ratios are even \textit{higher}. To demonstrate \modelname's savings in an established high compute training setup, we scale up to the full T5 compute regime, copying Raffel et al. exactly~\cite{2020t5}. This is the same as the C4 configuration in the previous section, but uses batches of $\sim$1M tokens, 64 TPUv3 chips and 537M parameters ($d_{model}=1024$, $d_{ff}=8192$, $L=24$). \modelname is 4.2X more compute efficient than the original T5 model and 2X more efficient than our strengthened Transformer++ baseline (Table~\ref{tab:large_t5}).\looseness=-1 

\begin{figure}[h!]
  \centering
  \includegraphics[width=0.7\linewidth]{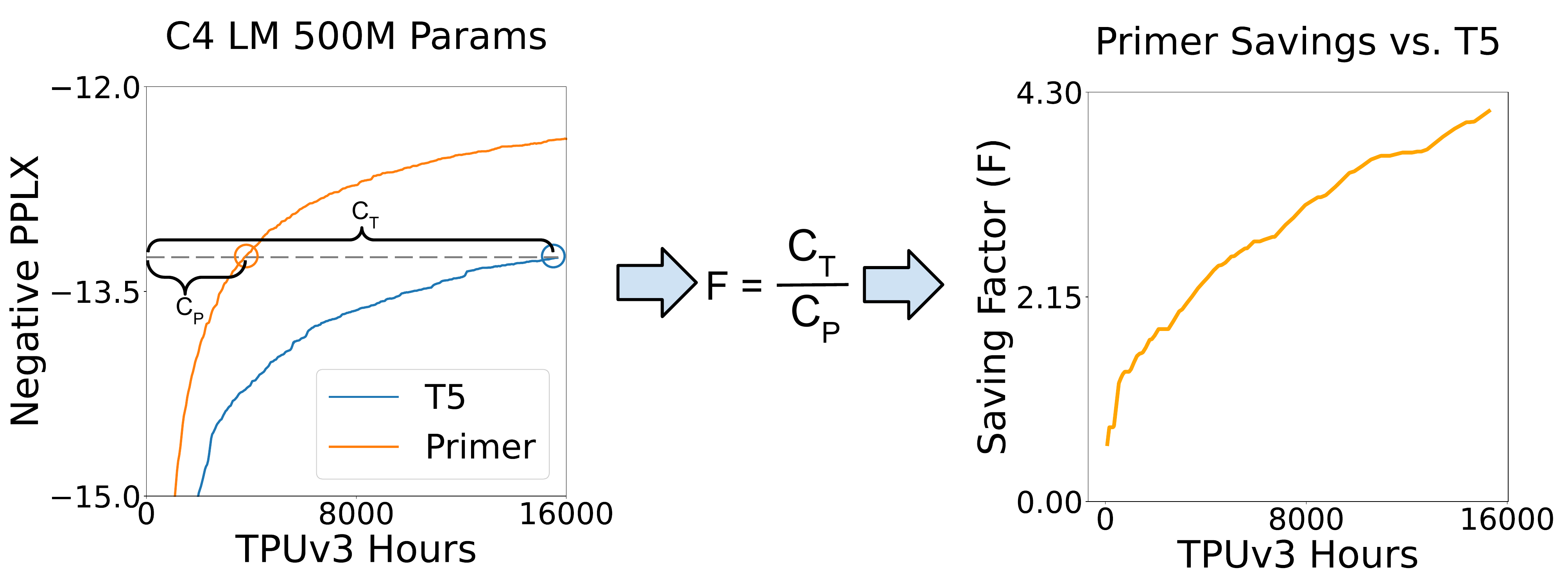}
  \caption{Compute savings of \modelname~ vs. the original T5 architecture on C4 LM over time. The more compute  invested in training, the higher the savings due to the asymptotic nature of both perplexity curves. \modelname achieves the same performance as the original T5 architecture with 4.2X less compute.}
  \label{fig:compute_savings_over_time}
\end{figure}

The reason why savings are even better here is because, at fixed sizes, more compute invested yields higher \modelname compute savings. Figure~\ref{fig:compute_savings_over_time} shows how the fraction of compute \modelname needs to achieve parity with the original T5 architecture shrinks as the models are trained for longer; this is due to the asymptotic nature of both the control and variable perplexity training curves.
This differs from the power law savings described in Section~\ref{sect:app_lm1b_t2t}. There, we use the optimal number of parameters for each compute budget, and so the compute saving factor, $b$, remains constant. For fixed model sizes, the compute saving factor grows as more compute is invested, meaning that compute savings can exceed the power law estimation. Note, this means that comparisons such as the ones given here can be ``gamed'' by investing more compute than is necessary for baseline models. It is for this reason that we use an exact replica of Raffel et al.'s~\cite{2020t5} training regime: to demonstrate Primer's savings in an already published training configuration.

\subsection{\modelname Transferability to Downstream One-Shot Tasks}
\label{sect:one_shot}

In our final comparison, we demonstrate Primer's improvements also hold in the \textit{pretraining} and \textit{one-shot downstream} task transfer regime. Recent trends in language modeling have moved towards training large models on large datasets, which is referred to as ``pretraining.'' These models are then transferred to unseen datasets and tasks, and, without much or any additional training, demonstrate the capacity to perform well on those ``downstream'' tasks~\cite{devlin2018bert, Dai2015SemisupervisedSL}. In the decoder-only auto-regressive language modeling configuration we study here, the most impressive results have been achieved by GPT-3~\cite{brown2020language}, which showed that large language models can exhibit strong performance on unseen tasks given only one example -- referred to as ``one-shot'' learning. In this section, we demonstrate that \modelname's training compute savings stretch beyond reaching a target pretraining perplexity and indeed transfer to downstream one-shot task performance.

To do this, we replicate the GPT-3 pretraining and one-shot evaluation setup. This replication is not exactly the same as the one used for GPT-3 because GPT-3 was not open sourced. Thus, these experiments are not meant to compare directly to GPT-3, as there are configuration differences. Instead, these experiments are used as a controlled comparison of the Transformer and \modelname architectures. We conduct these experiments in the Lingvo codebase using a proprietary pretraining dataset. The downstream tasks are configured in the same one-shot way described by Brown et al.~\cite{brown2020language}, with single prefix examples fed into each model with each task's inputs. We compare (1) a baseline 1.9B parameter Transformer ($d_{model}=2048$, $d_{ff}=12288$, $L=24$) with GELU activations, meant to approximate the GPT-3 XL architecture, and (2) a full \modelname without shared QK representations, which only hurt performance according to Appendix~\ref{sec:ablation_and_insertion}. Each model is trained using batches of $\sim$2M tokens using 512 TPUv4 chips for $\sim$140 hours ($\sim$71.8K total accelerator hours or $\sim$1M train steps). We once again use the T5 training hyperparemeters without any additional tuning.

\begin{figure}[h!]
  \centering
  \begin{minipage}{0.24\linewidth}
    \includegraphics[width=0.98\linewidth]{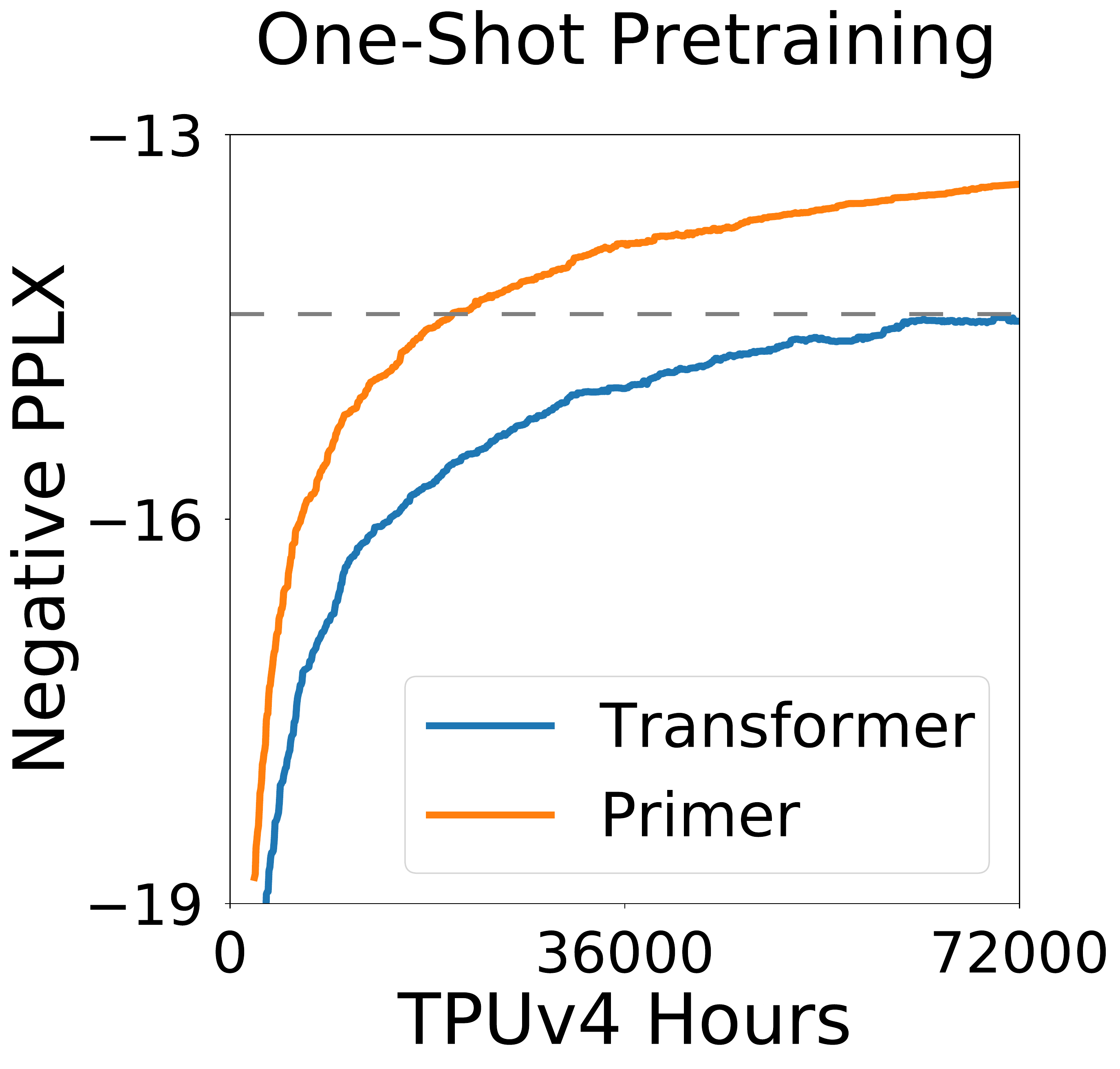}
  \end{minipage}
  \begin{minipage}{0.02\linewidth}
  \includegraphics[width=0.98\linewidth]{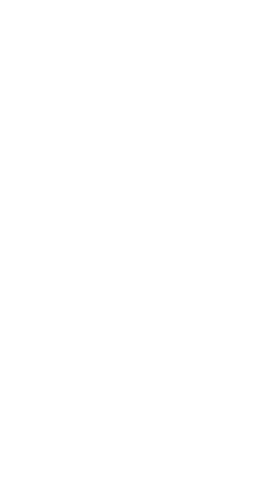}
  \end{minipage}
  \begin{minipage}{0.72\linewidth}
    \includegraphics[width=1.0\linewidth]{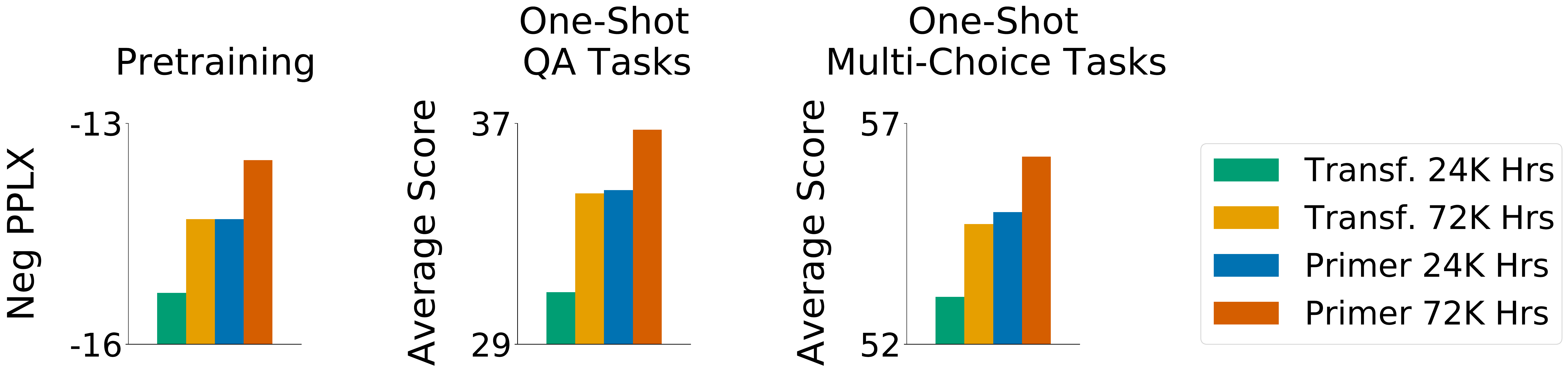}
  \end{minipage}
  \caption{Comparison between Transformer+GELU and \modelname at 1.9B parameters and varying training compute budgets on downstream one-shot tasks, similar to GPT-3. \modelname achieves slightly better performance than Transformer when given 3X less pretraining compute and substantially better performance when given the same pretraining compute. Here we stop at 72K TPUv4 hours to roughly match the quality of GPT-3 XL, but the compute savings of \modelname would be larger if we let the two models run longer (see Figure~\ref{fig:compute_savings_over_time}). Note, this is a crude comparison that uses averaged scores from the 27 one-shot tasks we evaluate. See Appendix~\ref{sect:individual_one_shot} (Table~\ref{table:one_shot_scores} and Figure~\ref{fig:one_shot_bars}) for exact task scores.}
  \label{fig:one_shot_aggregates}
\end{figure}

Figure~\ref{fig:one_shot_aggregates} shows that \modelname achieves the same pretraining perplexity and one-shot downstream performance as Transformer+GELU while using 3X less compute. Table~\ref{table:one_shot_scores} in the Appendix gives the exact performance numbers for each of the 27 evaluated downstream tasks. \modelname, despite using 3X less compute, outperforms Transfomer+GELU on 5 tasks, does worse on 1 task, and performs equivalently on the remaining 21 tasks. The same table shows that when given equivalent compute, \modelname outperforms Transformer+GELU on 15 tasks, does worse on 2 tasks, and performs equivalently on the remaining 10 tasks. This result shows that not only can \modelname improve language modeling perplexity, but the improvements also transfer to downstream NLP tasks.

\section{Conclusion}

\paragraph{Limitations:} There are limitations to this study. First, although our comparisons are at substantial sizes, they are still orders of magnitude smaller than state-of-the-art models such as the full-scale GPT-3~\cite{brown2020language}. Second, we focus primarily on decoder-only models, while encoder-based sequence models are still widely used~\cite{Vaswani2017AttentionIA,devlin2018bert,yang2019xlnet,liu2019roberta,Adiwardana2020TowardsAH,sutskever2014sequence}. In Appendix~\ref{sect:app_masked_language_modeling}, we perform encoder-decoder masked language modeling comparisons in T5, but do not study the results in significant depth. The main finding there is that, although \modelname modifications improve upon vanilla Transformer, they  perform only as well as Transformer++. This result suggests that architectural modifications that work well for decoder-only auto-regressive language models may not necessarily be as effective for encoder-based masked language models. Developing better encoders is a topic of our future research.

\paragraph{Recommendations and Future Directions:} We recommend the adoption of \modelname and \modelnamelite for \emph{auto-regressive} language modeling because of their strong performance, simplicity, and robustness to hyperparameter and codebase changes. To prove their potential, we simply dropped them into established codebases and, without any changes, showed that they can give significant performance boosts. Furthermore, in practice, additional tuning could further improve their performance.

We also hope our work  encourages more research into the development of efficient Transformers. For example, an important finding of this study is that small changes to activation functions can yield more efficient training. In the effort to reduce the cost of Transformers, more investment in the development of such simple changes could be a promising area for future exploration.

\section*{Acknowledgements}

We thank Zhen Xu for his help with infrastructure. We also thank Gabriel Bender, Hallie Cramer, Andrew Dai, Nan Du, Yanping Huang, Daphne Ippolito, Norm Jouppi, Lluis-Miquel Munguia, Sharan Narang, Ruoming Pang, David Patterson, Yanqi Zhou, and the Google Brain Team for their help and feedback.\looseness=-1

\bibliographystyle{unsrt}
\bibliography{main}

\begin{thebibliography}{10}

\bibitem{Vaswani2017AttentionIA}
Ashish Vaswani, Noam~M. Shazeer, Niki Parmar, Jakob Uszkoreit, Llion Jones,
  Aidan~N. Gomez, Lukasz Kaiser, and Illia Polosukhin.
\newblock Attention is all you need.
\newblock In {\em Advances in Neural Information Processing Systems}, 2017.

\bibitem{devlin2018bert}
Jacob Devlin, Ming-Wei Chang, Kenton Lee, and Kristina Toutanova.
\newblock Bert: Pre-training of deep bidirectional transformers for language
  understanding.
\newblock In {\em NAACL}, 2018.

\bibitem{yang2019xlnet}
Zhilin Yang, Zihang Dai, Yiming Yang, Jaime Carbonell, Ruslan Salakhutdinov,
  and Quoc~V Le.
\newblock Xlnet: Generalized autoregressive pretraining for language
  understanding.
\newblock In {\em Advances in Neural Information Processing Systems}, 2019.

\bibitem{liu2019roberta}
Yinhan Liu, Myle Ott, Naman Goyal, Jingfei Du, Mandar Joshi, Danqi Chen, Omer
  Levy, Mike Lewis, Luke Zettlemoyer, and Veselin Stoyanov.
\newblock Roberta: A robustly optimized bert pretraining approach.
\newblock {\em arXiv preprint arXiv:1907.11692}, 2019.

\bibitem{2020t5}
Colin Raffel, Noam Shazeer, Adam Roberts, Katherine Lee, Sharan Narang, Michael
  Matena, Yanqi Zhou, Wei Li, and Peter~J. Liu.
\newblock Exploring the limits of transfer learning with a unified text-to-text
  transformer.
\newblock {\em Journal of Machine Learning Research}, 21(140):1--67, 2020.

\bibitem{Adiwardana2020TowardsAH}
Daniel Adiwardana, Minh-Thang Luong, David~R. So, J.~Hall, Noah Fiedel,
  R.~Thoppilan, Z.~Yang, Apoorv Kulshreshtha, Gaurav Nemade, Yifeng Lu, and
  Quoc~V. Le.
\newblock Towards a human-like open-domain chatbot.
\newblock {\em arXiv preprint arXiv:2001.09977}, 2020.

\bibitem{brown2020language}
Tom~B Brown, Benjamin Mann, Nick Ryder, Melanie Subbiah, Jared Kaplan, Prafulla
  Dhariwal, Arvind Neelakantan, Pranav Shyam, Girish Sastry, Amanda Askell,
  et~al.
\newblock Language models are few-shot learners.
\newblock In {\em Advances in Neural Information Processing Systems}, 2020.

\bibitem{Fedus2021SwitchTS}
William Fedus, Barret Zoph, and Noam~M. Shazeer.
\newblock Switch transformers: Scaling to trillion parameter models with simple
  and efficient sparsity.
\newblock {\em arXiv preprint arXiv:2101.03961}, 2021.

\bibitem{Kaplan2020ScalingLF}
Jared Kaplan, Sam McCandlish, Tom Henighan, Tom~B. Brown, Benjamin Chess, Rewon
  Child, Scott Gray, Alec Radford, Jeffrey Wu, and Dario Amodei.
\newblock Scaling laws for neural language models.
\newblock {\em arXiv preprint arXiv:2001.08361}, 2020.

\bibitem{Abadi2016TensorFlowAS}
Mart{\'i}n Abadi, P.~Barham, J.~Chen, Z.~Chen, Andy Davis, J.~Dean, M.~Devin,
  Sanjay Ghemawat, Geoffrey Irving, M.~Isard, M.~Kudlur, Josh Levenberg, Rajat
  Monga, Sherry Moore, D.~Murray, Benoit Steiner, P.~Tucker, Vijay Vasudevan,
  Pete Warden, Martin Wicke, Y.~Yu, and Xiaoqiang Zhang.
\newblock Tensorflow: A system for large-scale machine learning.
\newblock In {\em OSDI}, 2016.

\bibitem{RealMSSSLK17}
Esteban Real, Sherry Moore, Andrew Selle, Saurabh Saxena, Yutaka~Leon Suematsu,
  Quoc~V. Le, and Alex Kurakin.
\newblock Large-scale evolution of image classifiers.
\newblock In {\em ICML}, 2017.

\bibitem{liu2017hierarchical}
Hanxiao Liu, Karen Simonyan, Oriol Vinyals, Chrisantha Fernando, and Koray
  Kavukcuoglu.
\newblock Hierarchical representations for efficient architecture search.
\newblock In {\em ICLR}, 2018.

\bibitem{so2019evolved}
David~R. So, Chen Liang, and Quoc~V. Le.
\newblock The evolved transformer.
\newblock In {\em ICML}, 2019.

\bibitem{liu2020evolving}
Hanxiao Liu, Andrew Brock, Karen Simonyan, and Quoc~V Le.
\newblock Evolving normalization-activation layers.
\newblock In {\em NeurIPS}, 2020.

\bibitem{Yao1999EvolvingAN}
Xin Yao.
\newblock Evolving artificial neural networks.
\newblock {\em Proceedings of the IEEE}, 87(9):1423--1447, 1999.

\bibitem{schmidhuber:1987:srl}
Jurgen Schmidhuber.
\newblock Evolutionary principles in self-referential learning. (on learning
  how to learn: The meta-meta-... hook.).
\newblock Diploma thesis, Technische Universitat Munchen, Germany, 1987.

\bibitem{Stanley2019DesigningNN}
Kenneth Stanley, Jeff Clune, Joel Lehman, and Risto Miikkulainen.
\newblock Designing neural networks through neuroevolution.
\newblock {\em Nature Machine Intelligence}, 1:24--35, 2019.

\bibitem{Radford2019LanguageMA}
Alec Radford, Jeffrey Wu, R.~Child, David Luan, Dario Amodei, and Ilya
  Sutskever.
\newblock Language models are unsupervised multitask learners.
\newblock In {\em Technical report, OpenAI}, 2019.

\bibitem{Schick2021ItsNJ}
Timo Schick and Hinrich Schütze.
\newblock It's not just size that matters: Small language models are also
  few-shot learners.
\newblock {\em arXiv preprint arXiv:2009.07118}, 2021.

\bibitem{Wang2021EntailmentAF}
Sinong Wang, Han Fang, Madian Khabsa, Hanzi Mao, and Hao Ma.
\newblock Entailment as few-shot learner.
\newblock {\em arXiv preprint arXiv:2104.14690}, 2021.

\bibitem{Gao2020MakingPL}
Tianyu Gao, Adam Fisch, and Danqi Chen.
\newblock Making pre-trained language models better few-shot learners.
\newblock {\em arXiv preprint arXiv:2012.15723}, 2020.

\bibitem{Rae2020CompressiveTF}
Jack~W. Rae, Anna Potapenko, Siddhant~M. Jayakumar, and T.~Lillicrap.
\newblock Compressive transformers for long-range sequence modelling.
\newblock {\em ArXiv}, abs/1911.05507, 2020.

\bibitem{tay2020synthesizer}
Yi~Tay, Dara Bahri, Donald Metzler, Da-Cheng Juan, Zhe Zhao, and Che Zheng.
\newblock Synthesizer: Rethinking self-attention in transformer models.
\newblock {\em arXiv preprint arXiv:2005.00743}, 2020.

\bibitem{chelba2014billion}
Ciprian Chelba, Tomas Mikolov, Mike Schuster, Qi~Ge, Thorsten Brants, Phillipp
  Koehn, and Tony Robinson.
\newblock One billion word benchmark for measuring progress in statistical
  language modeling.
\newblock In {\em Interspeech}, 2014.

\bibitem{tensor2tensor}
Ashish Vaswani, Samy Bengio, Eugene Brevdo, Francois Chollet, Aidan~N. Gomez,
  Stephan Gouws, Llion Jones, \L{}ukasz Kaiser, Nal Kalchbrenner, Niki Parmar,
  Ryan Sepassi, Noam Shazeer, and Jakob Uszkoreit.
\newblock Tensor2tensor for neural machine translation.
\newblock {\em arXiv preprint arXiv:1803.07416}, 2018.

\bibitem{tan2019efficientnet}
Mingxing Tan and Quoc Le.
\newblock Efficientnet: Rethinking model scaling for convolutional neural
  networks.
\newblock In {\em ICML}, 2019.

\bibitem{Tan2019MnasNetPN}
Mingxing Tan, Bo~Chen, Ruoming Pang, Vijay Vasudevan, and Quoc~V. Le.
\newblock Mnasnet: Platform-aware neural architecture search for mobile.
\newblock {\em 2019 IEEE/CVF Conference on Computer Vision and Pattern
  Recognition (CVPR)}, pages 2815--2823, 2019.

\bibitem{Cai2019ProxylessNASDN}
Han Cai, Ligeng Zhu, and Song Han.
\newblock Proxylessnas: Direct neural architecture search on target task and
  hardware.
\newblock {\em arXiv preprint:1812.00332}, 2019.

\bibitem{Elsken2019EfficientMN}
Thomas Elsken, Jan~Hendrik Metzen, and Frank Hutter.
\newblock Efficient multi-objective neural architecture search via lamarckian
  evolution.
\newblock In {\em ICLR}, 2019.

\bibitem{Real2019RegularizedEF}
Esteban Real, Alok Aggarwal, Yanping Huang, and Quoc~V. Le.
\newblock Regularized evolution for image classifier architecture search.
\newblock In {\em AAAI}, 2019.

\bibitem{Elsken2019NeuralAS}
Thomas Elsken, Jan~Hendrik Metzen, and Frank Hutter.
\newblock Neural architecture search: A survey.
\newblock {\em Journal of Machine Learning Research}, 20(55):1--21, 2019.

\bibitem{Li2019RandomSA}
Liam Li and Ameet~S. Talwalkar.
\newblock Random search and reproducibility for neural architecture search.
\newblock In {\em UAI}, 2019.

\bibitem{Sciuto2020EvaluatingTS}
Kaicheng Yu, Christian Sciuto, Martin Jaggi, Claudiu Musat, and Mathieu
  Salzmann.
\newblock Evaluating the search phase of neural architecture search.
\newblock In {\em ICLR}, 2020.

\bibitem{Bender2020CanWS}
Gabriel Bender, Hanxiao Liu, Bo~Chen, Grace Chu, Shuyang Cheng, Pieter-Jan
  Kindermans, and Quoc~V. Le.
\newblock Can weight sharing outperform random architecture search? {A}n
  investigation with tunas.
\newblock In {\em CVPR}, 2020.

\bibitem{Real2020AutoMLZeroEM}
Esteban Real, Chen Liang, David~R. So, and Quoc~V. Le.
\newblock Automl-zero: Evolving machine learning algorithms from scratch.
\newblock In {\em ICML}, 2020.

\bibitem{Krotov2016DenseAM}
Dmitry Krotov and John~J. Hopfield.
\newblock Dense associative memory for pattern recognition.
\newblock In {\em Advances in Neural Information Processing Systems}, 2016.

\bibitem{Jayakumar2020MultiplicativeI}
Siddhant~M. Jayakumar, Jacob Menick, Wojciech~M. Czarnecki, Jonathan Schwarz,
  Jack~W. Rae, Simon Osindero, Y.~Teh, Tim Harley, and Razvan Pascanu.
\newblock Multiplicative interactions and where to find them.
\newblock In {\em ICLR}, 2020.

\bibitem{dauphin2017language}
Yann~N Dauphin, Angela Fan, Michael Auli, and David Grangier.
\newblock Language modeling with gated convolutional networks.
\newblock In {\em ICML}, 2017.

\bibitem{shazeer2020glu}
Noam Shazeer.
\newblock Glu variants improve transformer.
\newblock {\em arXiv preprint arXiv:2002.05202}, 2020.

\bibitem{Hendrycks2016BridgingNA}
Dan Hendrycks and Kevin Gimpel.
\newblock Bridging nonlinearities and stochastic regularizers with gaussian
  error linear units.
\newblock {\em arXiv preprint arXiv:1606.08415}, 2016.

\bibitem{wei2018qanet}
Adams~Wei Yu, David Dohan, Minh{-}Thang Luong, Rui Zhao, Kai Chen, Mohammad
  Norouzi, and Quoc~V. Le.
\newblock {QANet}: Combining local convolution with global self-attention for
  reading comprehension.
\newblock In {\em ICLR}, 2018.

\bibitem{gulati2020conformer}
Anmol Gulati, James Qin, Chung-Cheng Chiu, Niki Parmar, Yu~Zhang, Jiahui Yu,
  Wei Han, Shibo Wang, Zhengdong Zhang, Yonghui Wu, and Ruoming Pang.
\newblock Conformer: Convolution-augmented transformer for speech recognition.
\newblock In {\em Interspeech}, 2020.

\bibitem{Wu2021CvTIC}
Haiping Wu, Bin Xiao, Noel Codella, Mengchen Liu, Xiyang Dai, Lu~Yuan, and Lei
  Zhang.
\newblock {CvT}: Introducing convolutions to vision transformers.
\newblock {\em arXiv preprint arXiv:2103.15808}, 2021.

\bibitem{Baevski2019AdaptiveIR}
Alexei Baevski and Michael Auli.
\newblock Adaptive input representations for neural language modeling.
\newblock In {\em arXiv preprint arXiv:1809.10853}, 2019.

\bibitem{Xiong2020OnLN}
Ruibin Xiong, Yunchang Yang, Di~He, Kai Zheng, S.~Zheng, Chen Xing, Huishuai
  Zhang, Yanyan Lan, L.~Wang, and T.~Liu.
\newblock On layer normalization in the transformer architecture.
\newblock {\em arXiv preprint arXiv:2002.04745}, 2020.

\bibitem{Ba2016LayerN}
Jimmy Ba, Jamie Kiros, and Geoffrey~E. Hinton.
\newblock Layer normalization.
\newblock {\em arXiv preprint arXiv:1607.06450}, 2016.

\bibitem{Zhang2019RootMS}
Biao Zhang and Rico Sennrich.
\newblock Root mean square layer normalization.
\newblock In {\em NeurIPS}, 2019.

\bibitem{Ramachandran2018SearchingFA}
Prajit Ramachandran, Barret Zoph, and Quoc~V. Le.
\newblock Searching for activation functions.
\newblock {\em arXiv preprint arXiv:1710.05941}, 2018.

\bibitem{narang2021}
Sharan Narang, Hyung~Won Chung, Yi~Tay, William Fedus, Thibault F{\'{e}}vry,
  Michael Matena, Karishma Malkan, Noah Fiedel, Noam Shazeer, Zhenzhong Lan,
  Yanqi Zhou, Wei Li, Nan Ding, Jake Marcus, Adam Roberts, and Colin Raffel.
\newblock Do transformer modifications transfer across implementations and
  applications?
\newblock {\em arXiv preprint arXiv:2102.11972}, 2021.

\bibitem{Shen2019LingvoAM}
Jonathan Shen, P.~Nguyen, Yonghui Wu, Z.~Chen, M.~Chen, Ye~Jia, Anjuli Kannan,
  T.~Sainath, Yuan Cao, C.~Chiu, Yanzhang He, J.~Chorowski, Smit Hinsu,
  S.~Laurenzo, James Qin, Orhan Firat, Wolfgang Macherey, Suyog Gupta, Ankur
  Bapna, Shuyuan Zhang, Ruoming Pang, Ron~J. Weiss, Rohit Prabhavalkar, Qiao
  Liang, Benoit Jacob, Bowen Liang, HyoukJoong Lee, Ciprian Chelba,
  S{\'e}bastien Jean, Bo~Li, M.~Johnson, Rohan Anil, Rajat Tibrewal, Xiaobing
  Liu, Akiko Eriguchi, Navdeep Jaitly, Naveen Ari, Colin Cherry, Parisa
  Haghani, Otavio Good, Youlong Cheng, R.~{\'A}lvarez, Isaac Caswell, Wei-Ning
  Hsu, Zongheng Yang, Kuan-Chieh Wang, Ekaterina Gonina, Katrin Tomanek, Ben
  Vanik, Zelin Wu, Llion Jones, M.~Schuster, Y.~Huang, Dehao Chen, Kazuki Irie,
  George~F. Foster, John Richardson, Uri Alon, and E.~al.
\newblock Lingvo: a modular and scalable framework for sequence-to-sequence
  modeling.
\newblock {\em ArXiv}, abs/1902.08295, 2019.

\bibitem{Dai2015SemisupervisedSL}
Andrew~M. Dai and Quoc~V. Le.
\newblock Semi-supervised sequence learning.
\newblock In {\em Advances in Neural Information Processing Systems}, 2015.

\bibitem{sutskever2014sequence}
Ilya Sutskever, Oriol Vinyals, and Quoc~V Le.
\newblock Sequence to sequence learning with neural networks.
\newblock In {\em Advances in Neural Information Processing Systems}, 2014.

\bibitem{pmlr-v28-karnin13}
Zohar Karnin, Tomer Koren, and Oren Somekh.
\newblock Almost optimal exploration in multi-armed bandits.
\newblock In {\em Proceedings of the 30th International Conference on Machine
  Learning}, 2013.

\bibitem{JMLR:v18:16-558}
Lisha Li, Kevin Jamieson, Giulia DeSalvo, Afshin Rostamizadeh, and Ameet
  Talwalkar.
\newblock Hyperband: A novel bandit-based approach to hyperparameter
  optimization.
\newblock {\em Journal of Machine Learning Research}, 18(185):1--52, 2018.

\bibitem{Helmuth2018ProgramSU}
Thomas Helmuth, N.~McPhee, and L.~Spector.
\newblock Program synthesis using uniform mutation by addition and deletion.
\newblock {\em Proceedings of the Genetic and Evolutionary Computation
  Conference}, 2018.

\bibitem{Shazeer2018AdafactorAL}
Noam~M. Shazeer and Mitchell Stern.
\newblock Adafactor: Adaptive learning rates with sublinear memory cost.
\newblock {\em arXiv preprint arXiv:1804.04235}, 2018.

\bibitem{Shaw2018SelfAttentionWR}
Peter Shaw, Jakob Uszkoreit, and Ashish Vaswani.
\newblock Self-attention with relative position representations.
\newblock In {\em NAACL-HLT}, 2018.

\bibitem{Kudo2018SentencePieceAS}
Taku Kudo and J.~Richardson.
\newblock Sentencepiece: A simple and language independent subword tokenizer
  and detokenizer for neural text processing.
\newblock In {\em EMNLP}, 2018.

\bibitem{Patterson2021CarbonEA}
David Patterson, Joseph Gonzalez, Quoc~V. Le, Chen Liang, Llu{\'i}s-Miquel
  Mungu{\'i}a, D.~Rothchild, David~R. So, Maud Texier, and J.~Dean.
\newblock Carbon emissions and large neural network training.
\newblock {\em arXiv preprint arXiv:2104.10350}, 2021.

\bibitem{googlecarbon}
24/7 carbon-free energy: Methodologies and metrics.
\newblock
  \url{https://www.gstatic.com/gumdrop/sustainability/24x7-carbon-free-energy-methodologies-metrics.pdf}.
\newblock Accessed: 2021-09-14.

\end{thebibliography}

\newpage
\appendix

\section{Appendix}

\subsection{TensorFlow Primitives Vocabulary}
\label{sect:app_tf_vocab}
\begin{table}[h!]
\begin{tabular}{@{}c|c|cccc@{}}
\multirow{2}{*}{Name} &
  \multirow{2}{*}{TF Function} &
  \multicolumn{4}{c}{\textit{Argument Mapping}}\\ \cmidrule(lr){3-6}
            &                                    & Input 1       & Input 2 & Constant & \begin{tabular}[c]{@{}c@{}}Dim\\ Size\end{tabular}                                                     \\ \midrule
ADD         & tf.math.add                        & x             & y       & -        & -\\
DIFFERENCE  & tf.math.subtract                   & x             & y       & -        & -\\
DIVIDE      & tf.math.divide                     & x             & y       & -        & -\\
MULTIPLY    & tf.math.multiply                   & x             & y       & -        & -\\
ABS ROOT   & tf.math.sqrt(tf.abs(x))            & x             & -       & -        & -\\
SQUARE      & tf.math.square                     & x             & -       & -        & -\\
EXP         & tf.exp                             & x             & -       & -        & -\\
LOG         & tf.log(tf.abs(x))                  & x             & -       & -        & -\\
C MUL      & tf.math.multiply                   & x             & -       & y        & -\\
ABS         & tf.abs                             & x             & -       & -        & -\\
RECIP       & tf.math.reciprocal\_no\_nan        & x             & -       & -        & -\\
SIGN        & tf.sign                            & x             & -       & -        & -\\
COS         & tf.cos                             & x             & -       & -        & -\\
SIN         & tf.sin                             & x             & -       & -        & -\\
TANH        & tf.tanh                            & x             & -       & -        & -\\
MAX         & tf.math.maximum                    & x             & -       & y        & -\\
MIN         & tf.math.minimum                    & x             & -       & y        & -\\
SCALE       & x+tf.Variable()                         & x             & -       & -        & -\\
SHIFT       & x*tf.Variable()                         & x             & -       & -        & -\\
SIGMOID     & tf.sigmoid                         & x             & -       & -        & -\\
MASK        & tf.linalg.band\_part               & input         & -       & -        & -\\
CUM PROD   & tf.math.cumprod                    & x             & -       &          & -\\
CUM SUM    & tf.math.cumsum                     & x             & -       & -        & -\\
RED MEAN   & tf.reduce\_mean                    & input\_tensor & -       & -        & -\\
RED SUM    & tf.reduce\_sum                     & input\_tensor & -       & -        & -\\
RED MIN    & tf.reduce\_min                     & input\_tensor & -       & -        & -\\
RED MAX    & tf.reduce\_max                     & input\_tensor & -       & -        & -\\
RED PROD   & tf.reduce\_prod                    & input\_tensor & -       & -        & -\\
MAT MUL    & tf.matmul                          & a             & b       & -        & -\\
T-MAT MUL & tf.matmul(transpose\_b=True) & a             & b       & -        & -\\
CONV 1X1       & tf.layers.dense                    & inputs        & -       & -        & units \\
CONV 3X1   & tf.nn.conv1d                       & input         & -       & -        & filters\\
CONV 7X1   & tf.nn.conv1d                       & input         & -       & -        & filters\\
CONV 15X1  & tf.nn.conv1d                       & input         & -       & -        & filters\\
CONV 31X1  & tf.nn.conv1d                       & input         & -       & -        & filters\\
DCONV 3X1  & tf.nn.depthwise\_conv2d            & input         & -       & -        & filters\\
DCONV 7X1  & tf.nn.depthwise\_conv2d            & input         & -       & -        & filters\\
DCONV 15X1 & tf.nn.depthwise\_conv2d            & input         & -       & -        & filters\\
DCONV 31X1 & tf.nn.depthwise\_conv2d            & input         & -       & -        & filters
\\\hline
\end{tabular}
 \caption{\label{tab:tf-primitives-vocab}TensorFlow (TF) Primitives Vocabulary. ``Name'' is the name of the operation in our search space. ``TF Function'' is the TensorFlow function that the name is mapped to when a DNA instruction is being converted to a line of TensorFlow code. ``Argument Mapping'' describes how the values in a DNA's argument set are mapped to the corresponding TensorFlow function arguments. This vocabulary is largely constructed from the lowest level TF operations needed to create Transformers (see Appendix \ref{sect:app_search_model_subprograms}). We additionally extend those operations to include adjacent operations; for example, we extend \textsc{max} to also include \textsc{min}, extend \textsc{red sum} to include \textsc{red product}, and extend \textsc{conv 1x1} to include \textsc{conv 3x1}. We also add commonly used math primitives such as \textsc{sin} and \textsc{abs}.}
\end{table}

\subsection{Constructing TensorFlow Graphs}
\label{sect:app_tf_graphs}

TensorFlow graphs are built from DNA programs as described in Section~\ref{sect:methods} of the main text. Here we provide additional implementation details.

\paragraph{Relative Dimensions:} We use relative dimensions~\cite{so2019evolved} instead of absolute dimensions for each instruction's ``dimension size'' argument. This allows us to resize the models to fit within our parameter limits (32M to 38M parameters). The vocabulary for these relative dimensions is [1, 2, 4, 8, 12, 16, 24, 32, 48, 64]. This vocabulary was not tuned.

\paragraph{Values Bank:} For ``constant'' and ``dimension size'' argument fields, we create a shared bank of values that each instruction references. The constants bank holds 2 values and the dimension sizes bank holds 6 values; these numbers were not tuned. Instead of each instruction possessing their own individual values for these arguments, they instead hold an index to these shared banks. This allows multiple instructions to share the same value and to change simultaneously when that value is changed. For example, each of the individual attention multi-head projections for $Q$, $K$ and $V$ start off sharing the same output dimension size so that they all change simultaneously if that value changes. See~\ref{sect:app_evolution_details} for an example of how these bank values are mutated.

\paragraph{Causal Masking:} An important part of teacher-forced language model training is that positions cannot ``see'' the token they are trying to predict. Each position should only get information from previous positions, otherwise the model will be degenerate when the targets are not provided. To enforce this causal constraint we add additional overhead to operations that move information spatially to mask out any information from future positions. For example, when applying convolutions we follow the standard practice of shifting the inputs spatially by (\textsc{kernel width} $-$ 1) so that each position only receives information from previous positions.

\paragraph{Branching: } To enable multi-head capabilities for the Transformer search seed, we add a meta argument to our instructions called ``branching.'' This argument can take any value in [1, 2, 4, 8, 16] and determines how many times that instruction is executed in parallel, with the resulting tensors being concatenated together along their embedding axes. Branching can be used with any of the TensorFlow primitives as well as with any of a DNA's subprograms. This allows us to initialize the search with multi-head self-attention by branching \textsc{subprogram 1} (self-attention) 8 times (see Appendix \ref{sect:app_search_model_subprograms} for subprogram implementations). \modelname does not utilize this branching capability in any meaningful way, beyond using the initialized multi-head attention.

\paragraph{Resolving Dimension Mismatches: } We do not constrain how tensor dimensions can be mutated and so programs may be invalid because they perform binary operations on tensors with incompatible sizes. For example, a program may describe adding together two tensors with differing embedding sizes. To resolve these dimension mismatch issues we deterministically pseudorandomly set one of the tensor dimensions to match the other.

\subsection{Halving Hurdles}
\label{sect:app_halving_hurdles}

We configure our hurdles~\cite{so2019evolved} such that the top 50\% of individuals passes each hurdle, according to fitness. We space the hurdles in such a way that the expected amount of compute devoted to training each hurdle band is roughly equal at the end of the search. That is, given that our maximum amount of training compute for an individual is 7 hours or 25,200 seconds (s), we construct hurdles at the 812.9s, 2438.7s, 5690.3s, and 12,193.5s marks. Thus, 1/5 of the compute budget is devoted to training every individual up to the first hurdle (812.9s), 1/5 of the compute budget is devoted to training the $\sim$50\% of individuals that are trained from the first to the second hurdle (2438.7s $-$ 812.9s = 1625.8s), 1/5 of the compute budget is devoted to training the $\sim$25\% of individuals that are trained from the second to the third hurdle (5690.3s $-$ 2438.7s = 3251.6s), etc. This configuration strategy, which we refer to as ``halving hurdles,'' requires setting only one hyperparameter, the number of hurdles, and removes the need to set hurdle threshold values and comparison steps, as has been previously done~\cite{so2019evolved, Real2020AutoMLZeroEM}. We choose four hurdles because five hurdles would require the first hurdle to be anchored at less than ten minutes of training, which we find empirically to be too noisy of a signal. Using hurdles in this way decreases the average train time per model to 4064s or about 1 hour and 8 minutes, reducing the compute cost by a factor of $\sim$6.2X.

This strategy is not unlike bandit algorithms such as Successive Halving\cite{pmlr-v28-karnin13} and Hyperband\cite{JMLR:v18:16-558}, however we do not use a static population of individuals created a priori, but integrate our halving with the changing evolutionary population.

\begin{figure}[h!]
  \centering
  \begin{minipage}{0.45\linewidth}
    \raggedleft
    \includegraphics[width=5cm]{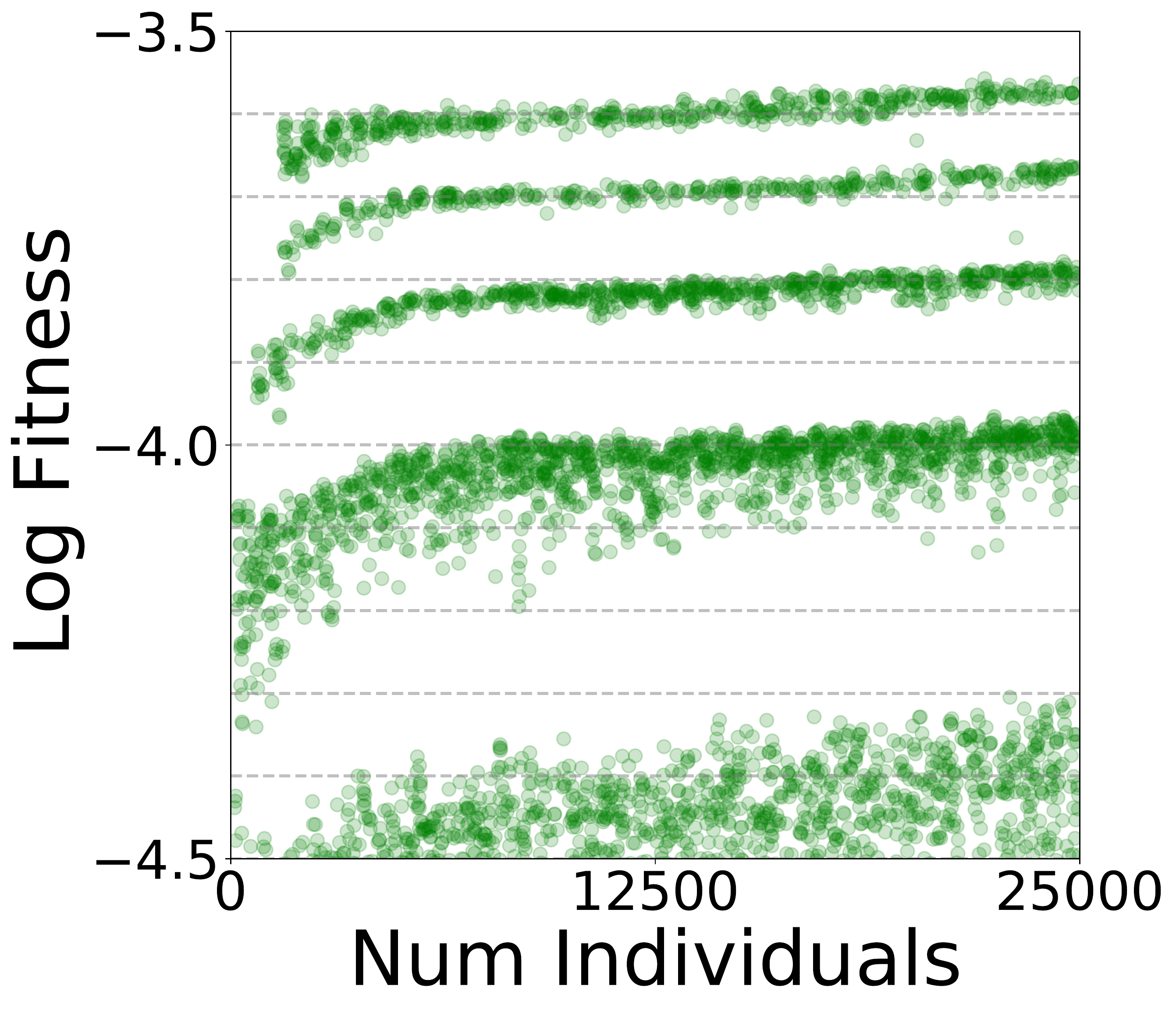}
  \end{minipage}
  \caption{Halving hurdles from our \modelname search. Each dot represents the final fitness of an individual generated by evolution. Different ``bands'' form because each hurdle has a different training allowance. All bands see improvement over time, meaning that the median fitness improves for all compute allowances. This correlation between a model's performances at different training budgets allows us to reduce our total search cost by roughly a factor of 6.2X.}
  \label{fig:hurdles}
\end{figure}

\subsection{Evolution Search Details}

We use Regularized Evolution~\cite{Real2019RegularizedEF} with a population size of 100 and a tournament selection size of 10. These values were not tuned. The mutations we use are as follows.

\label{sect:app_evolution_details}
\paragraph{Mutations:} To create new candidates in our search, we uniform randomly select a \textit{parent} from our search population and apply a single \textit{mutation} to it. We employ five different mutation types (selections and decisions are performed uniform randomly unless specified otherwise):
\begin{itemize}
  \item \textit{Delete}: Remove an instruction from a subprogram. 
  \item \textit{Insert}: Create an instruction and insert it into a subprogram.
  \item \textit{Delete and Insert}: Perform a delete mutation followed by an insert mutation~\cite{Helmuth2018ProgramSU}.
  \item \textit{Mutate Field}: Select a field from an instruction and change its value.
  \item \textit{Swap}: Swap the position of two instructions in a randomly selected subprogram. The input tensors for each instruction are also swapped so that the net effect is switching the positions of the instructions in the compute graph.
  \item \textit{Mutate Bank Value}: Change the value of a relative tensor dimension or constant in the corresponding bank. The values for relative tensor dimensions are selected from their vocabulary (see Appendix~\ref{sect:app_tf_graphs}). The values for constants are changed according to\newline $c_{new} := c_{prev} \cdot 10^{X} + Y$ for previous value $c_{prev}$, new value $c_{new}$ and random variables $X, Y\sim N(0, 1)$.
\end{itemize}

After a mutation is applied, we run a light check to see if the resulting candidate's compute graph is exactly equivalent to the parent's compute graph. If it is, we perform another mutation.

\subsection{Transformer and \modelname Program Comparisons}
\label{sect:app_search_model_subprograms}

\begin{figure}[h!]
\centering
\includegraphics[height=13cm]{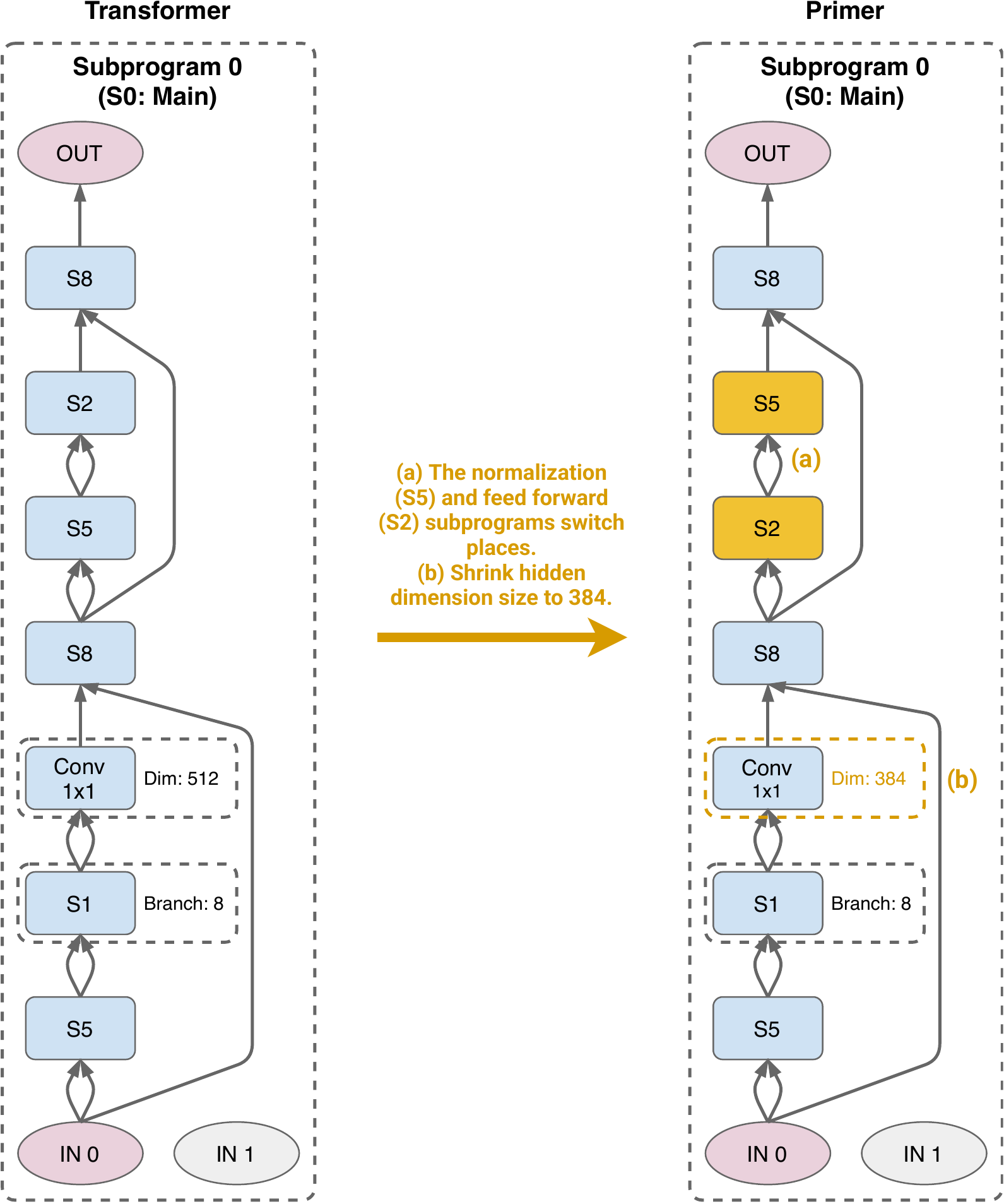}
\caption{Main subprograms. Changes are highlighted in orange.}
\label{fig:subprogram0_diff}
\end{figure}

\begin{figure}[h!]
\centering
\includegraphics[height=11.48cm]{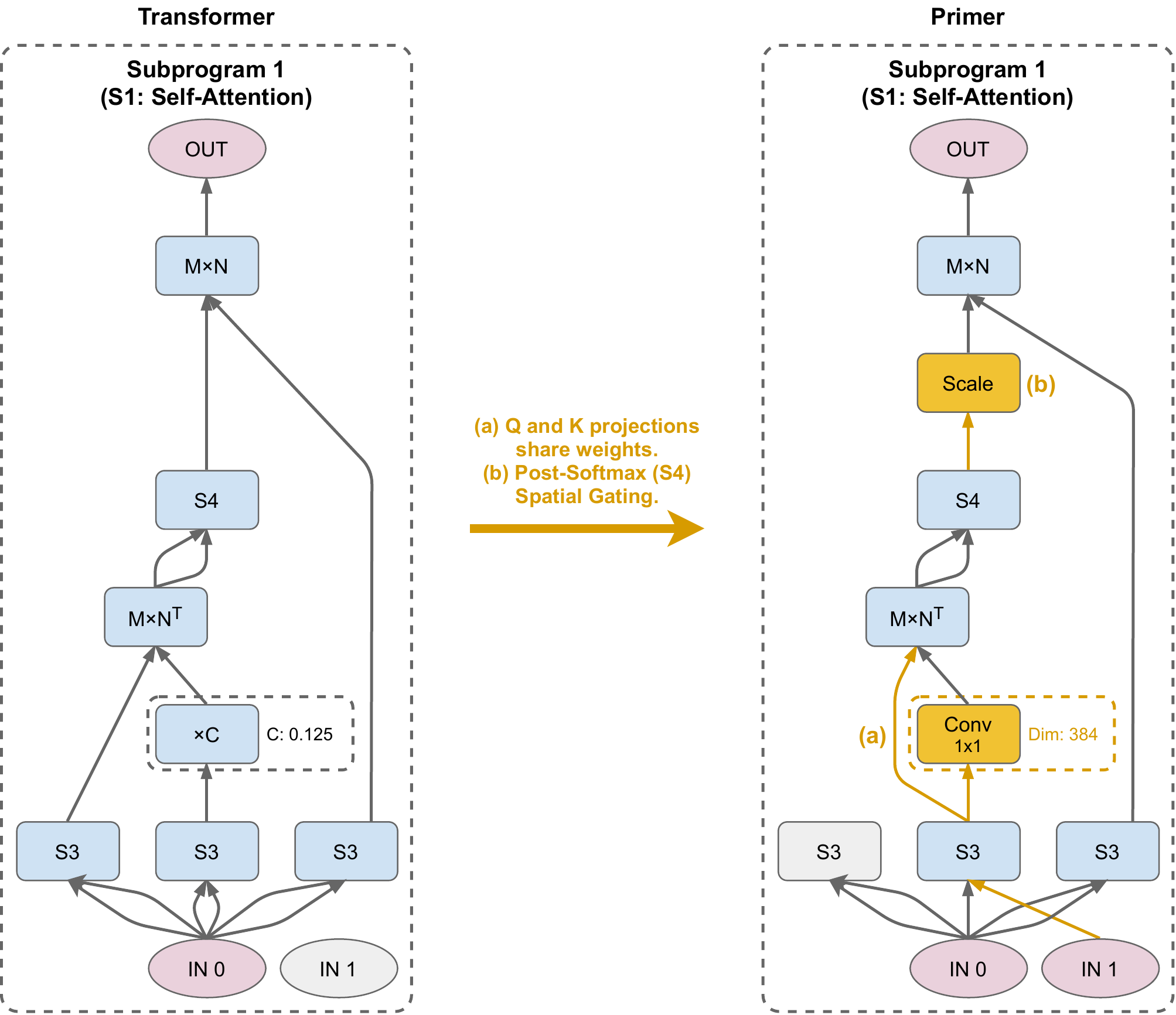}
\caption{Attention subprograms. Changes are highlighted in orange.}
\label{fig:subprogram1_diff}
\end{figure}

\begin{figure}[h!]
\centering
\includegraphics[height=10.2cm]{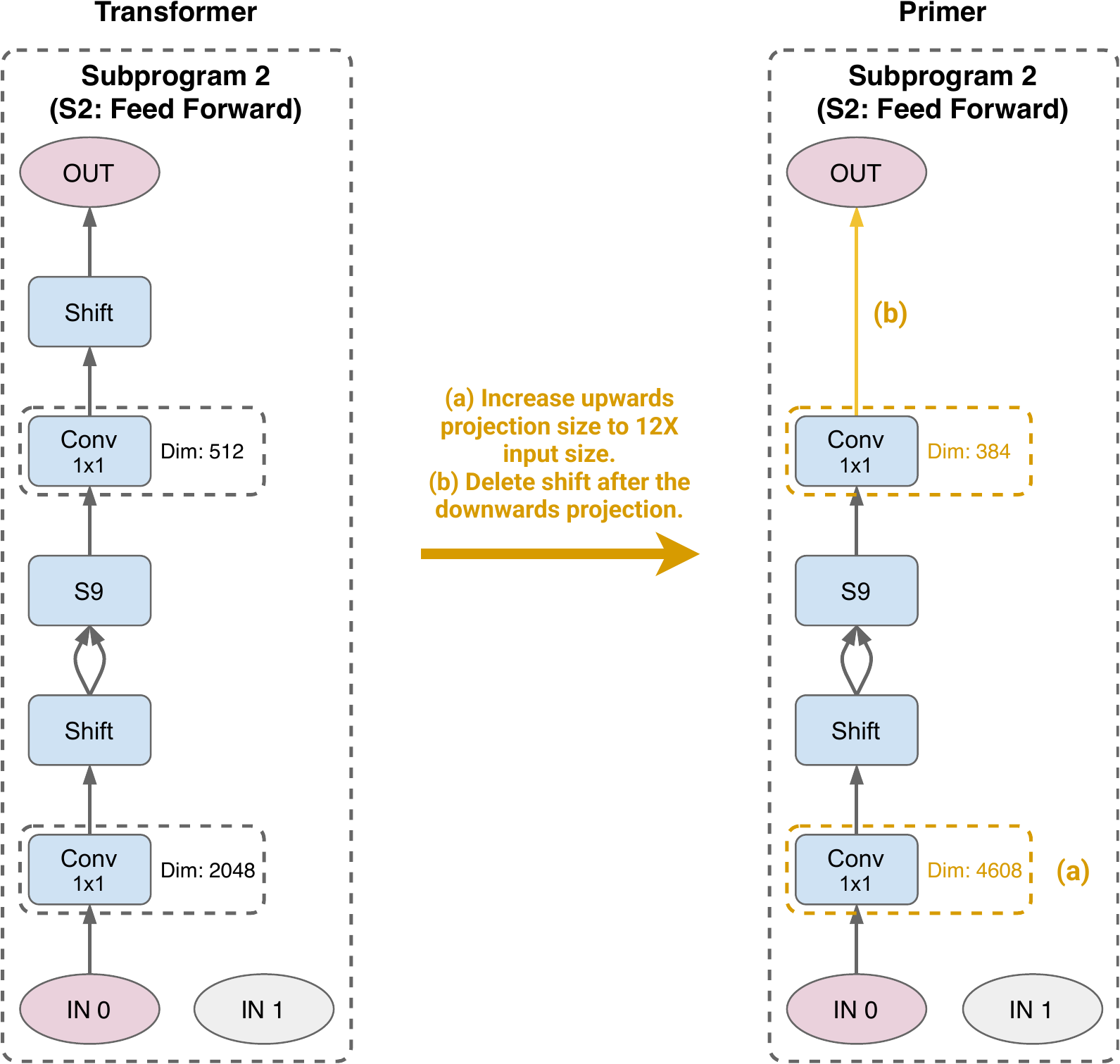}
\caption{Feed forward subprograms. Changes are highlighted in orange.}
\label{fig:subprogram2_diff}
\end{figure}

\begin{figure}[h!]
\centering
\includegraphics[height=7.4cm]{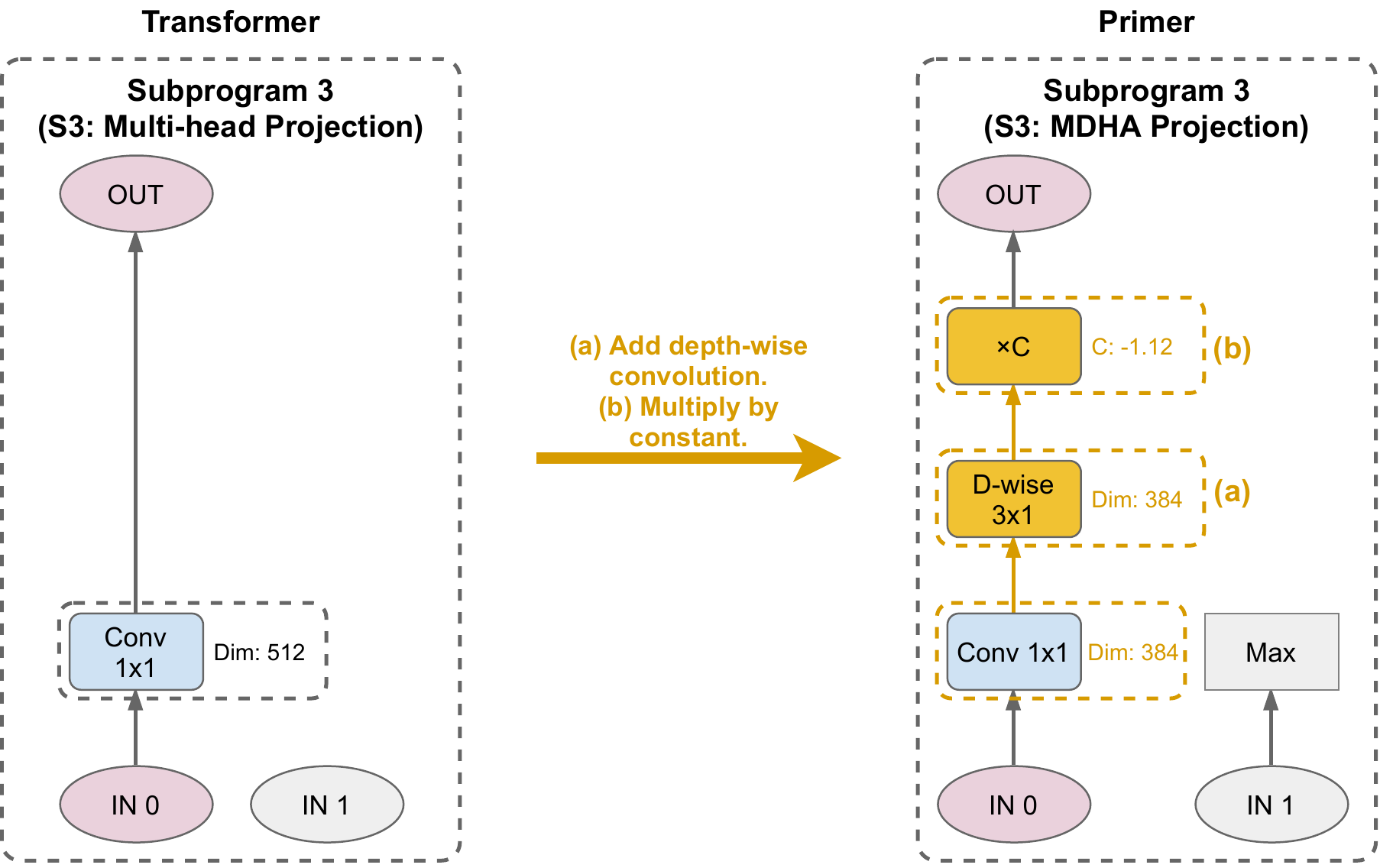}
\caption{Multi-head projection subprograms. Changes are highlighted in orange.}
\label{fig:subprogram3_diff}
\end{figure}

\begin{figure}[h!]
\centering
\includegraphics[height=10.39cm]{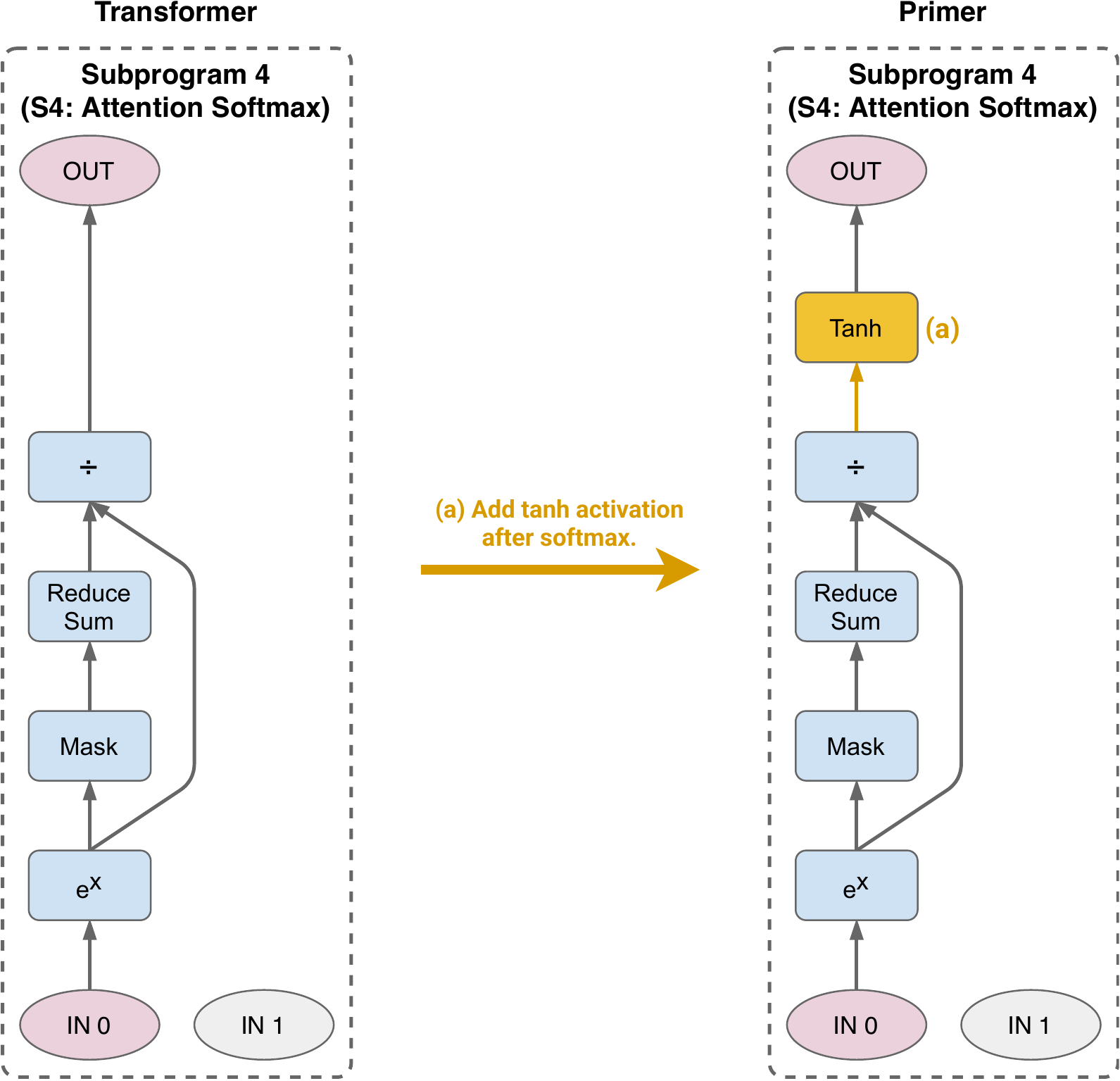}
\caption{Softmax subprograms. Changes are highlighted in orange.}
\label{fig:subprogram4_diff}
\end{figure}

\begin{figure}[h!]
\centering
\includegraphics[height=6cm]{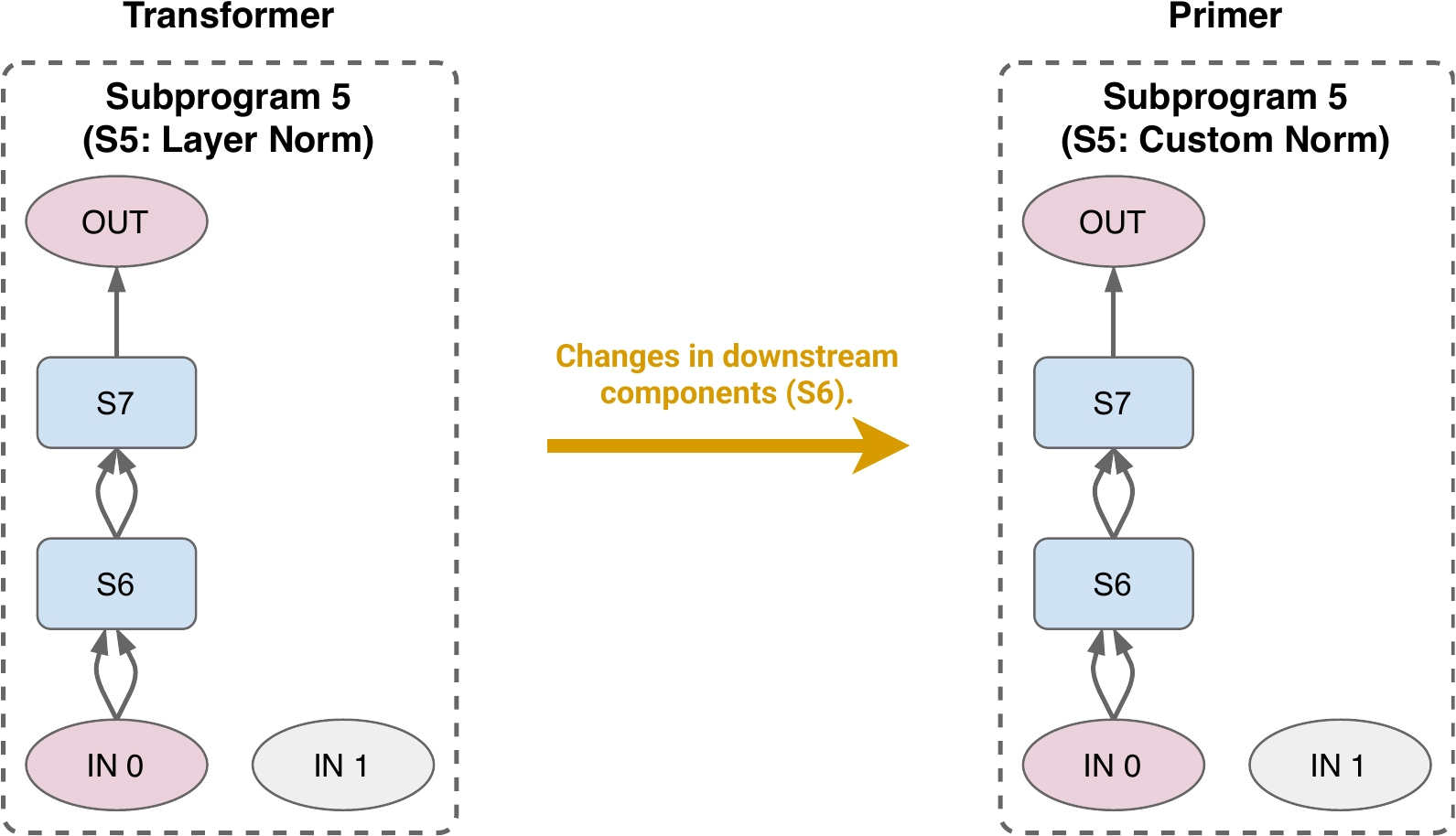}
\caption{Normalization subprograms. Changes to this subprogram are realized in downstream changes to S6.}
\label{fig:subprogram5_diff}
\end{figure}

\begin{figure}[h!]
\centering
\includegraphics[height=11.6cm]{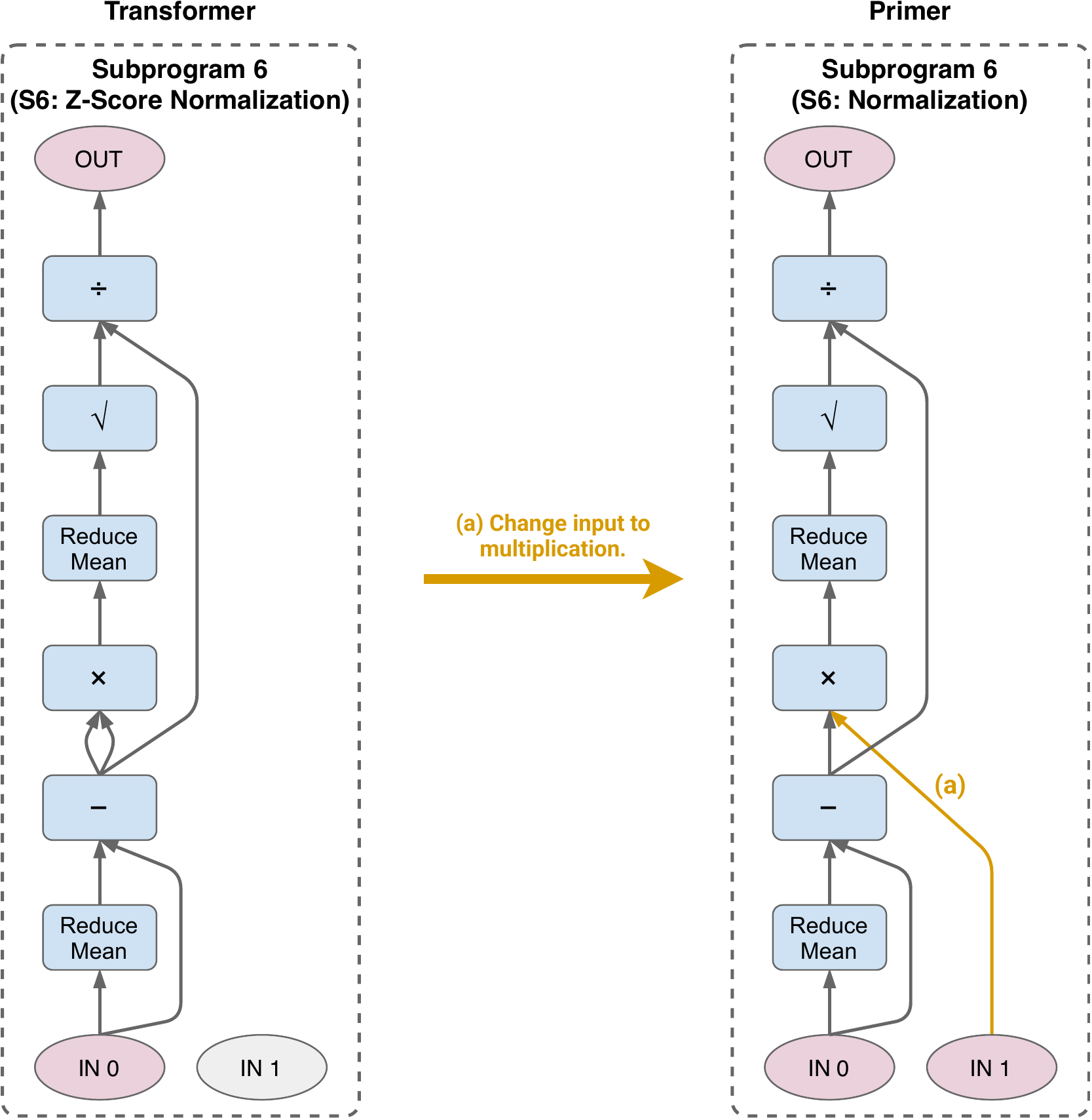}
\caption{Z-score normalization subprograms. Changes are highlighted in orange.}
\label{fig:subprogram6_diff}
\end{figure}

\begin{figure}[h!]
\centering
\includegraphics[height=6cm]{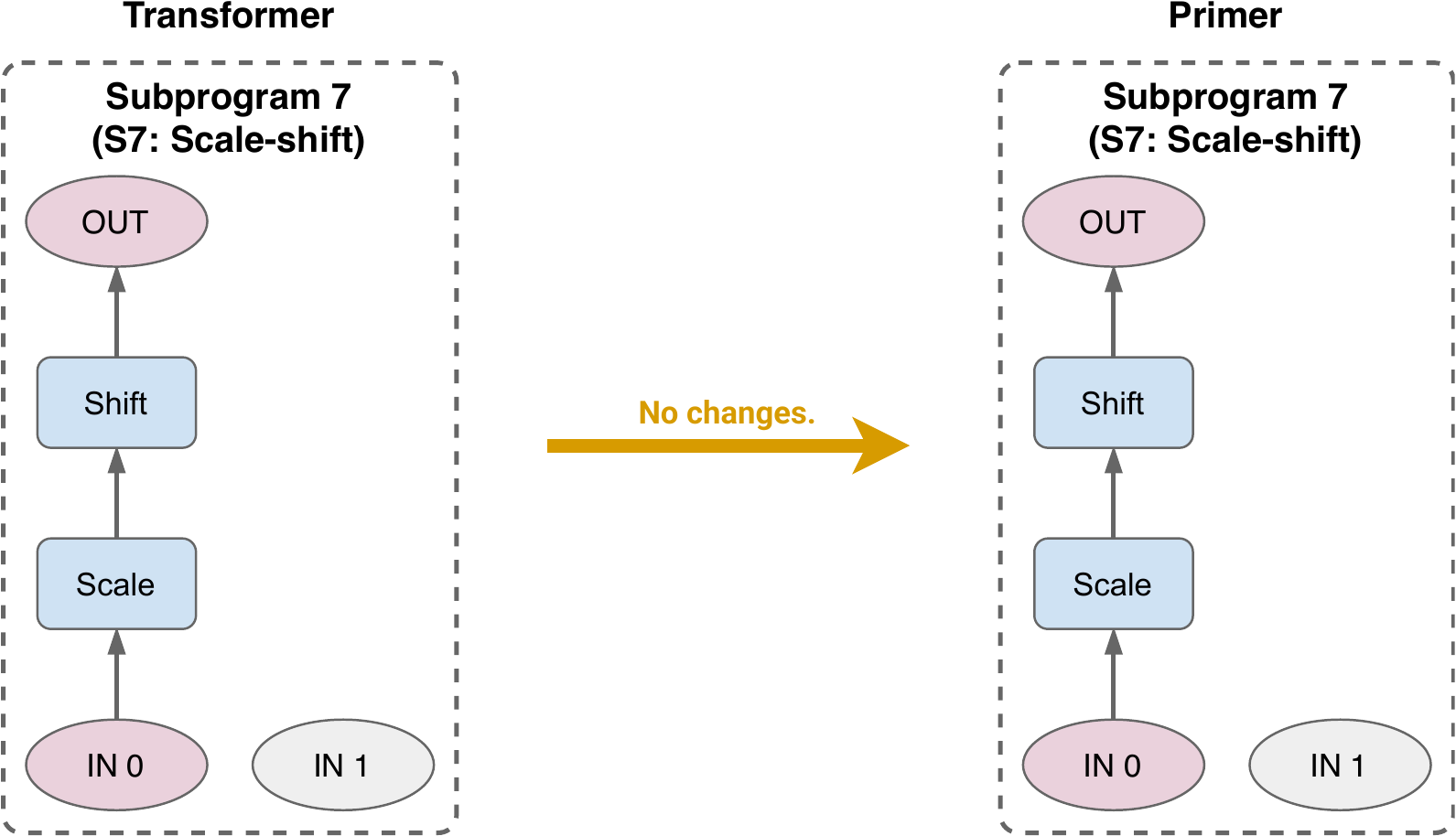}
\caption{Scale-shift subprograms. No changes here.}
\label{fig:subprogram7_diff}
\end{figure}

\begin{figure}[h!]
\centering
\includegraphics[height=6cm]{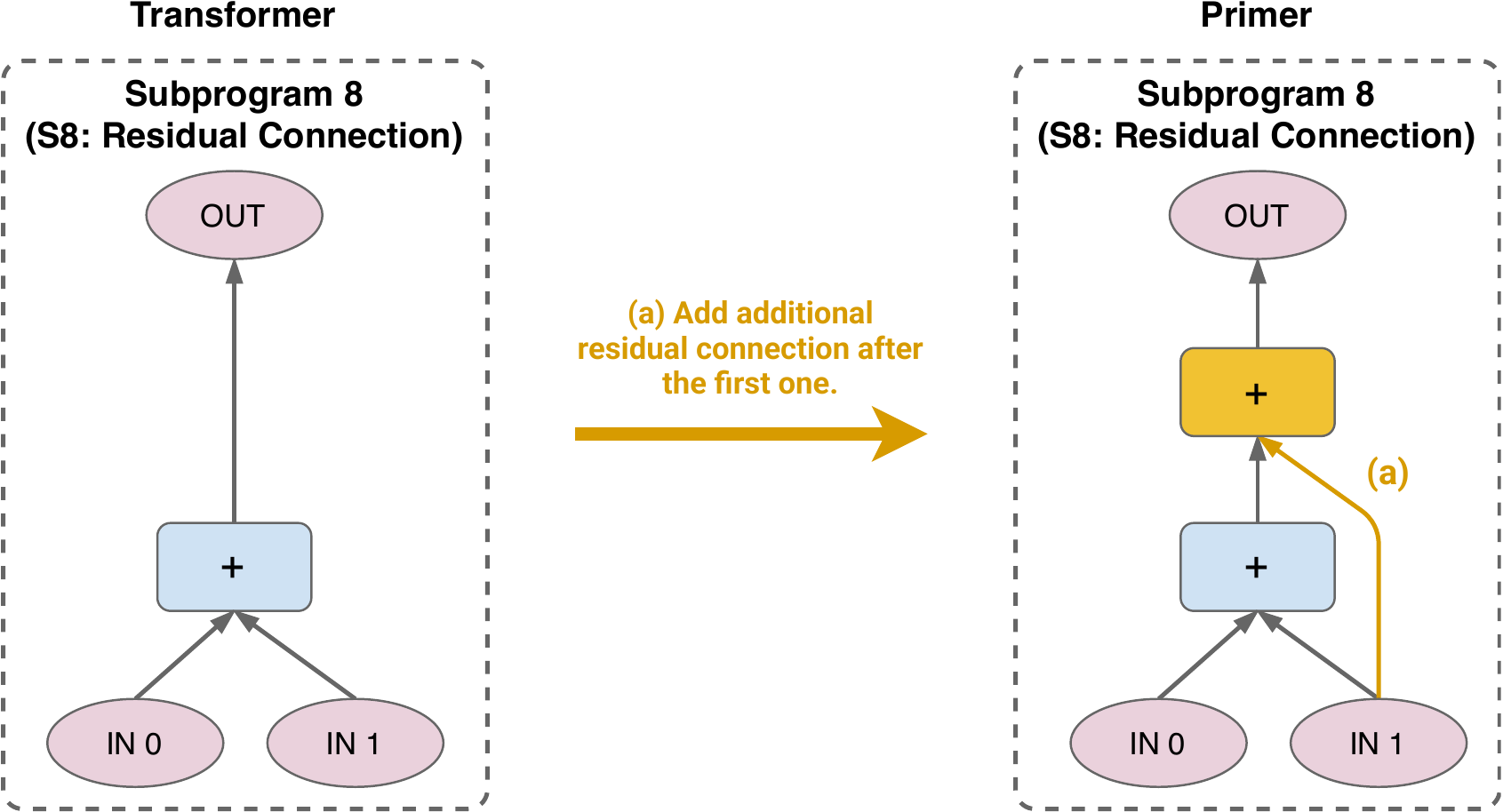}
\caption{Residual connection subprograms. This change is essentially a functional no-op.}
\label{fig:subprogram8_diff}
\end{figure}

\begin{figure}[h!]
\centering
\includegraphics[height=6cm]{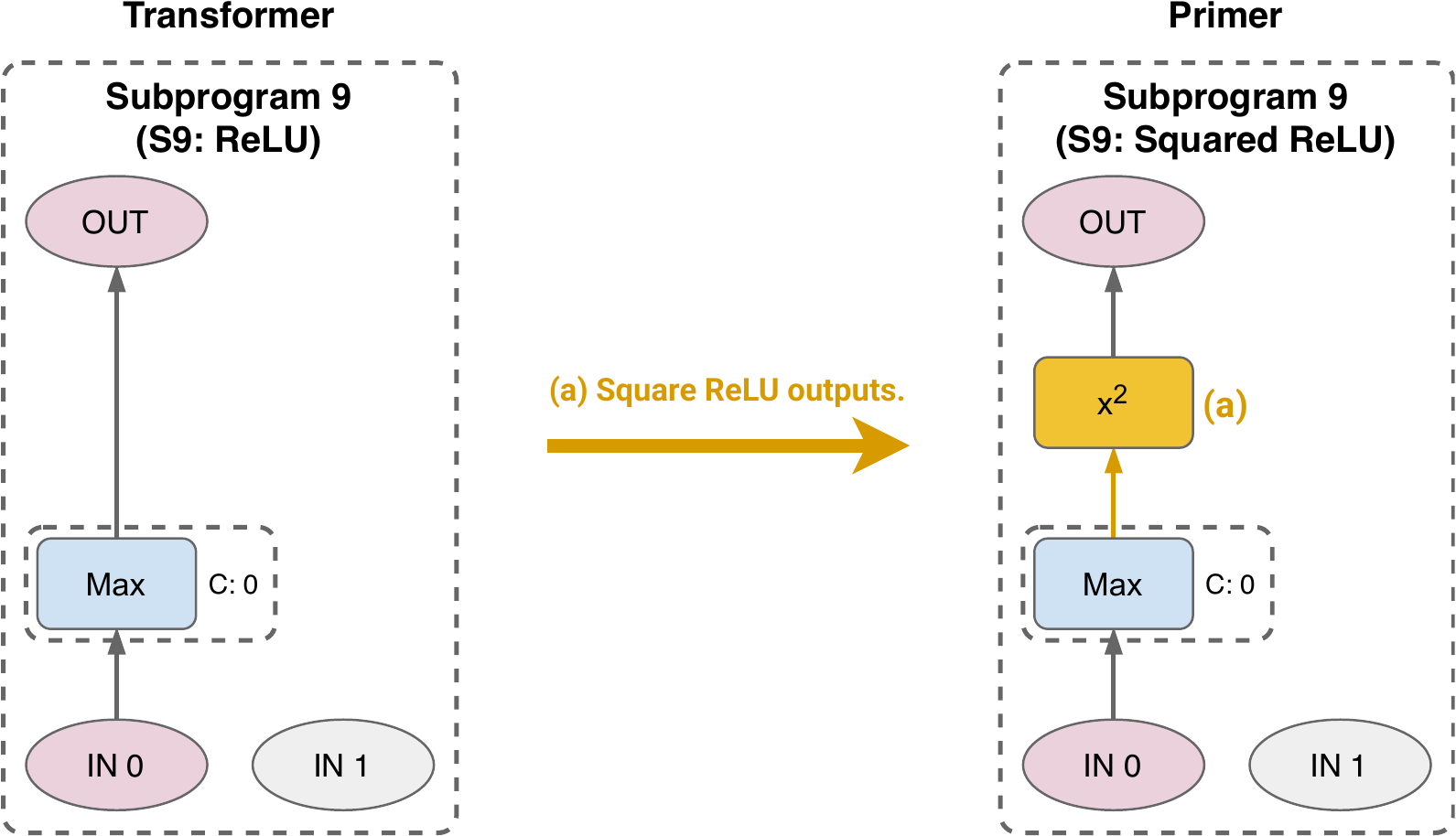}
\caption{Activation function subprograms. Changes are highlighted in orange.}
\label{fig:subprogram9_diff}
\end{figure}

\begin{figure}[h!]
  \centering
  \begin{minipage}{0.45\linewidth}
    \raggedleft
    \includegraphics[height=21cm]{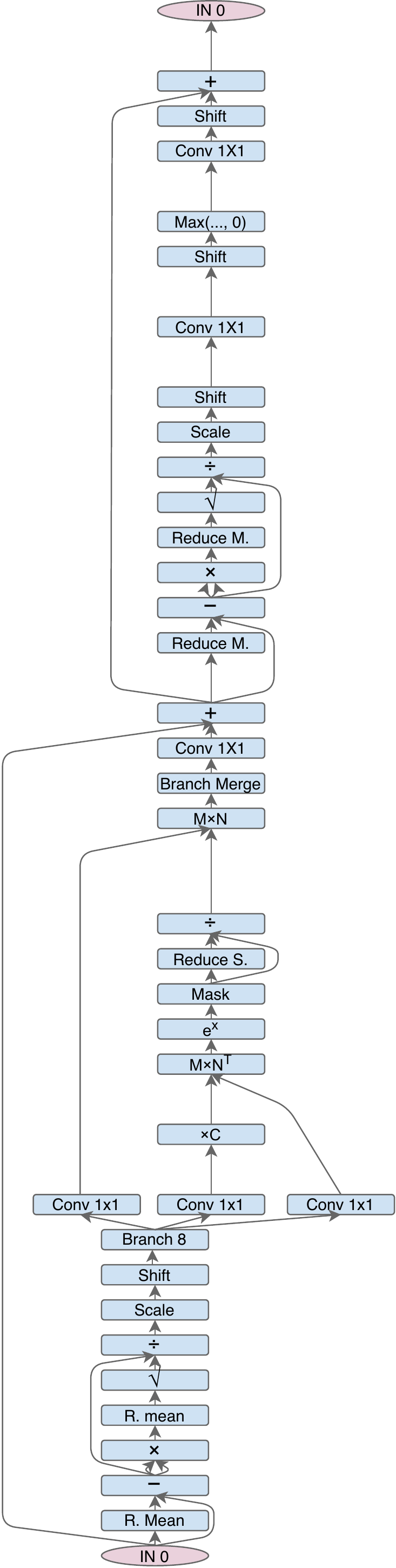}
  \end{minipage}
  \hspace{1.2cm}
    \begin{minipage}{0.45\linewidth}
    \raggedright
    \includegraphics[height=21cm]{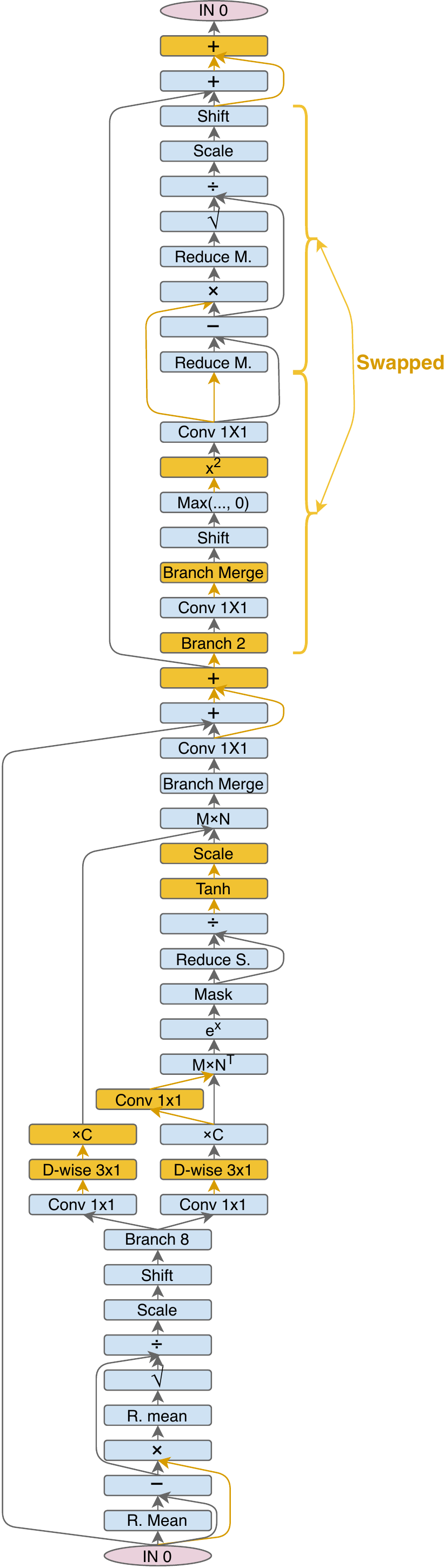}
  \end{minipage}
  \caption{Comparison of Transformer (Left) and \modelname (Right) programs, with all subprograms resolved to their constituent primitives. \modelname differences are highlighted in orange.}
  \label{fig:primer_and_transformer_full}
\end{figure}

Here we present the programs for both the Transformer seed and the discovered \modelname model. Table~\ref{tab:primitives-graphing-symbols} is a key that maps operation names to graph symbols for subsequent graphs. Figures~\ref{fig:subprogram0_diff} to~\ref{fig:subprogram9_diff} depict the subprograms for each model with the \modelname changes highlighted in orange. Figure~\ref{fig:primer_and_transformer_full} depicts the full compute graphs for each model, with all subprograms resolved to their constituent primitives. Figures~\ref{fig:transformer-program} and~\ref{fig:primer-program} depict the DNA programs for Transformer and \modelname with all subprograms resolved and all instruction bank values plugged in.

\begin{table}[h!]
\begin{tabular}{@{}lc@{}}
\toprule
\multirow{2}{*}{Name} & \multirow{2}{*}{Graphing symbol}                      \\
                      &                                                       \\ \midrule
ADD                   & +                                                     \\
DIFFERENCE            & $-$                                                   \\
DIVIDE                & $\div$                                                \\
MULTIPLY              & $\times$                                              \\
ABS ROOT             & $\sqrt{}$                                             \\
SQUARE                & $x^2$                                                 \\
EXP                   & $e^x$                                                 \\
LOG                   & Log                                                   \\
C MUL                & $\times C$                                            \\
ABS                   & $\mid x \mid$                                         \\
RECIP                 & Recip                                                 \\
SIGN                  & Sign                                                  \\
COS                   & Cos                                                   \\
SIN                   & Sin                                                   \\
TANH                  & Tanh                                                  \\
MAX                   & Max                                                   \\
MIN                   & Min                                                   \\
SCALE                 & Scale                                                 \\
SHIFT                 & Shift                                                 \\
SIGMOID               & Sigm                                                  \\
MASK                  & Mask                                                  \\
CUM PROD             & \begin{tabular}[c]{@{}c@{}}Cum  Prod\end{tabular}    \\
CUM SUM              & \begin{tabular}[c]{@{}c@{}}Cum  Sum\end{tabular}     \\
RED MEAN             & \begin{tabular}[c]{@{}c@{}}Reduce  Mean\end{tabular} \\
RED SUM              & \begin{tabular}[c]{@{}c@{}}Reduce  Sum\end{tabular}  \\
RED MIN              & \begin{tabular}[c]{@{}c@{}}Reduce  Min\end{tabular}  \\
RED MAX              & \begin{tabular}[c]{@{}c@{}}Reduce  Max\end{tabular}  \\
RED PROD             & \begin{tabular}[c]{@{}c@{}}Reduce  Prod\end{tabular} \\
MAT MUL              & $M \times N$                                          \\
T-MAT MUL           & $M \times N^T$                                        \\
CONV 1X1                 & \begin{tabular}[c]{@{}c@{}}Conv 1x1\end{tabular}    \\
CONV 3X1             & \begin{tabular}[c]{@{}c@{}}Conv 3x1\end{tabular}    \\
CONV 7X1             & \begin{tabular}[c]{@{}c@{}}Conv  7x1\end{tabular}    \\
CONV 15X1            & \begin{tabular}[c]{@{}c@{}}Conv  15x1\end{tabular}   \\
CONV 31X1            & \begin{tabular}[c]{@{}c@{}}Conv  31x1\end{tabular}   \\
DCONV 3X1            & \begin{tabular}[c]{@{}c@{}}D-wise  3x1\end{tabular}  \\
DCONV 7X1            & \begin{tabular}[c]{@{}c@{}}D-wise  7x1\end{tabular}  \\
DCONV 15X1           & \begin{tabular}[c]{@{}c@{}}D-wise  15x1\end{tabular} \\
DCONV 31X1           & \begin{tabular}[c]{@{}c@{}}D-wise  31x1\end{tabular} \\
\hline
\end{tabular}
 \caption{\label{tab:primitives-graphing-symbols}Key for primitives mapped to corresponding symbols used in the following graphs.}
\end{table}

\begin{figure}[h!]
    \begin{verbatim}
                              TRANSFORMER

(0)  INPUT
(1)  INPUT
(2)  REDUCE_MEAN            In0: 0    In1: 0    Dim: 128    C: 0.00 
(3)  DIFFERENCE             In0: 0    In1: 2    Dim: 128    C: 0.00 
(4)  MULTIPLY               In0: 3    In1: 3    Dim: 128    C: 0.00 
(5)  REDUCE_MEAN            In0: 4    In1: 4    Dim: 128    C: 0.00 
(6)  ABS_SQUARE_ROOT        In0: 5    In1: 5    Dim: 128    C: 0.00 
(7)  DIVIDE                 In0: 3    In1: 6    Dim: 128    C: 0.00 
(8)  SCALE                  In0: 7    In1: 7    Dim: 128    C: 0.00 
(9)  SHIFT                  In0: 8    In1: 8    Dim: 128    C: 0.00 
(10) BRANCH_8_INPUT_1       In0: 9    In1: 9    Dim: 128    C: 0.00 
(11) BRANCH_8_INPUT_2       In0: 9    In1: 9    Dim: 128    C: 0.00 
(12) DENSE                  In0: 10   In1: 10   Dim: 64     C: 0.00 
(13) DENSE                  In0: 10   In1: 10   Dim: 64     C: 0.00 
(14) CONSTANT_MUL           In0: 13   In1: 13   Dim: 128    C: 0.12 
(15) DENSE                  In0: 10   In1: 10   Dim: 64     C: 0.00 
(16) TRANSPOSE_MAT_MUL      In0: 14   In1: 12   Dim: 128    C: 0.00 
(17) EXP                    In0: 16   In1: 16   Dim: 128    C: 0.00 
(18) EMBEDDING_MASK         In0: 17   In1: 17   Dim: 128    C: 0.00 
(19) REDUCE_SUM             In0: 18   In1: 18   Dim: 128    C: 0.00 
(20) DIVIDE                 In0: 18   In1: 19   Dim: 128    C: 0.00 
(21) MAT_MUL                In0: 20   In1: 15   Dim: 128    C: 0.00 
(22) BRANCH_MERGE           In0: 21   In1: 21   Dim: 512    C: 0.00 
(23) DENSE                  In0: 22   In1: 22   Dim: 512    C: 0.00 
(24) ADD                    In0: 0    In1: 23   Dim: 128    C: 0.00 
(25) REDUCE_MEAN            In0: 24   In1: 24   Dim: 128    C: 0.00 
(26) DIFFERENCE             In0: 24   In1: 25   Dim: 128    C: 0.00 
(27) MULTIPLY               In0: 26   In1: 26   Dim: 128    C: 0.00 
(28) REDUCE_MEAN            In0: 27   In1: 27   Dim: 128    C: 0.00 
(29) ABS_SQUARE_ROOT        In0: 28   In1: 28   Dim: 128    C: 0.00 
(30) DIVIDE                 In0: 26   In1: 29   Dim: 128    C: 0.00 
(31) SCALE                  In0: 30   In1: 30   Dim: 128    C: 0.00 
(32) SHIFT                  In0: 31   In1: 31   Dim: 128    C: 0.00 
(33) DENSE                  In0: 32   In1: 32   Dim: 2048   C: 0.00 
(34) SHIFT                  In0: 33   In1: 33   Dim: 128    C: 0.00 
(35) MAX                    In0: 34   In1: 34   Dim: 128    C: 0.00 
(36) DENSE                  In0: 35   In1: 35   Dim: 512    C: 0.00 
(37) SHIFT                  In0: 36   In1: 36   Dim: 128    C: 0.00 
(38) ADD                    In0: 24   In1: 37   Dim: 128    C: 0.00 
\end{verbatim}
  \caption{List of instructions for Transformer program, with all subprograms resolved to their constituent primitives.}
  \label{fig:transformer-program}
\end{figure}
\begin{figure}[h!]
    \begin{verbatim}
                                  PRIMER

(0)  INPUT
(1)  INPUT
(2)  REDUCE_MEAN            In0: 0    In1: 0    Dim: 768    C: -1.12
(3)  DIFFERENCE             In0: 0    In1: 2    Dim: 768    C: -1.12
(4)  MULTIPLY               In0: 3    In1: 0    Dim: 768    C: -1.12
(5)  REDUCE_MEAN            In0: 4    In1: 4    Dim: 768    C: -1.12
(6)  ABS_SQUARE_ROOT        In0: 5    In1: 5    Dim: 768    C: -1.12
(7)  DIVIDE                 In0: 3    In1: 6    Dim: 768    C: -1.12
(8)  SCALE                  In0: 7    In1: 7    Dim: 768    C: -1.12
(9)  SHIFT                  In0: 8    In1: 8    Dim: 384    C: -0.57
(10) BRANCH_8_INPUT_1       In0: 9    In1: 9    Dim: 768    C: -1.12
(11) BRANCH_8_INPUT_2       In0: 9    In1: 9    Dim: 768    C: -1.12
(12) MAX                    In0: 10   In1: 10   Dim: 768    C: -0.57
(13) DENSE                  In0: 10   In1: 10   Dim: 48     C: -1.12
(14) DEPTHWISE_CONV_3X1     In0: 13   In1: 10   Dim: 384    C: -1.12
(15) CONSTANT_MUL           In0: 14   In1: 14   Dim: 384    C: -1.12
(16) MAX                    In0: 11   In1: 11   Dim: 768    C: -0.57
(17) DENSE                  In0: 10   In1: 10   Dim: 48     C: -1.12
(18) DEPTHWISE_CONV_3X1     In0: 17   In1: 10   Dim: 384    C: -1.12
(19) CONSTANT_MUL           In0: 18   In1: 18   Dim: 384    C: -1.12
(20) DENSE                  In0: 19   In1: 11   Dim: 48     C: -1.12
(21) MAX                    In0: 10   In1: 10   Dim: 768    C: -0.57
(22) DENSE                  In0: 10   In1: 10   Dim: 48     C: -1.12
(23) DEPTHWISE_CONV_3X1     In0: 22   In1: 10   Dim: 384    C: -1.12
(24) CONSTANT_MUL           In0: 23   In1: 23   Dim: 384    C: -1.12
(25) TRANSPOSE_MAT_MUL      In0: 20   In1: 19   Dim: 768    C: -1.12
(26) EXP                    In0: 25   In1: 25   Dim: 768    C: -1.12
(27) EMBEDDING_MASK         In0: 26   In1: 26   Dim: 768    C: -1.12
(28) REDUCE_SUM             In0: 27   In1: 27   Dim: 768    C: -1.12
(29) DIVIDE                 In0: 27   In1: 28   Dim: 768    C: -1.12
(30) TANH                   In0: 29   In1: 25   Dim: 384    C: -1.12
(31) SCALE                  In0: 30   In1: 19   Dim: 384    C: -1.12
(32) MAT_MUL                In0: 31   In1: 24   Dim: 768    C: -1.12
(33) BRANCH_MERGE           In0: 32   In1: 32   Dim: 384    C: -1.12
(34) DENSE                  In0: 33   In1: 33   Dim: 384    C: -1.12
(35) ADD                    In0: 0    In1: 34   Dim: 768    C: -1.12
(36) ADD                    In0: 35   In1: 34   Dim: 768    C: -1.12
(37) BRANCH_2_INPUT_1       In0: 36   In1: 36   Dim: 2304   C: -1.12
(38) BRANCH_2_INPUT_2       In0: 36   In1: 36   Dim: 2304   C: -1.12
(39) DENSE                  In0: 37   In1: 38   Dim: 2304   C: -1.12
(40) BRANCH_MERGE           In0: 39   In1: 39   Dim: 4608   C: -1.12
(41) SHIFT                  In0: 40   In1: 40   Dim: 768    C: -1.12
(42) MAX                    In0: 41   In1: 41   Dim: 768    C: -0.57
(43) SQUARE                 In0: 42   In1: 41   Dim: 768    C: -1.12
(44) DENSE                  In0: 43   In1: 43   Dim: 384    C: -1.12
(45) REDUCE_MEAN            In0: 44   In1: 44   Dim: 768    C: -1.12
(46) DIFFERENCE             In0: 44   In1: 45   Dim: 768    C: -1.12
(47) MULTIPLY               In0: 46   In1: 44   Dim: 768    C: -1.12
(48) REDUCE_MEAN            In0: 47   In1: 47   Dim: 768    C: -1.12
(49) ABS_SQUARE_ROOT        In0: 48   In1: 48   Dim: 768    C: -1.12
(50) DIVIDE                 In0: 46   In1: 49   Dim: 768    C: -1.12
(51) SCALE                  In0: 50   In1: 50   Dim: 768    C: -1.12
(52) SHIFT                  In0: 51   In1: 51   Dim: 384    C: -0.57
(53) ADD                    In0: 36   In1: 52   Dim: 768    C: -1.12
(54) ADD                    In0: 53   In1: 52   Dim: 768    C: -1.12
\end{verbatim}
  \caption{List of instructions for \modelname program, with all subprograms resolved to their constituent primitives.}
  \label{fig:primer-program}
\end{figure}

\clearpage

\subsection{Exact LM1B Numbers}
\label{sect:app_lm1b_t2t}

\begin{table}[h!]
\caption{Comparison on the search task, auto-regressive language modeling on LM1B, across two different hardware platforms (TPUv2s and V100 GPUs) and two different libraries (Tensor2Tensor and T5), using those libraries' default hyperparameters. This table contains the precise numbers for Figure~\ref{fig:search_task_bars}. ``Speedup'' describes the fraction of compute used by each model to achieve the same results as the vanilla Transformer baseline trained with the full compute budget. Even though \modelname was developed in Tensor2Tensor using TPUv2s, it shows strong performance on GPU and in T5. Perplexity is reported with respect to each library's default tokenization.}
\label{table:lm1b}
\begin{tabular}{@{}l|ccccc@{}}
\toprule
\multirow{2}{*}{\textbf{Model}} & \textbf{Params} & \textbf{Train Steps} & \textbf{Step/Sec} & \textbf{PPLX}                      & \textbf{Speedup}       \\ \cmidrule(l){2-6} 
                                & \multicolumn{5}{c|}{\textit{Tensor2Tensor, TPUv2}}                                                                    \\ \midrule
Vanilla Transformer             & 35M             & 1.9M                 & 22.4              & 35.44 +/- 0.30                     & -                      \\
Transformer+GELU                & 35M             & 1.9M                 & 22.4              & 35.00 +/- 0.12                     & 1.23 +/- 0.07          \\
Transformer++                   & 35M             & 1.9M                 & 22.0              & 34.87 +/- 0.46                     & 1.37 +/- 0.24          \\
Primer                          & 34M             & 1.9M                 & 21.7              & 33.77 +/- 0.15                     & 2.12 +/- 0.09          \\
Primer-EZ                       & 35M             & 1.8M                 & 21.0     & \textbf{33.53 +/- 0.09}            & \textbf{2.34 +/- 0.04} \\
\\
Transformer+MDHA                       & 35M             & 1.8M                 & 21.0     & 34.26 +/- 0.12            & 1.76 +/- 0.06 \\
Transformer+Sep Conv                       & 35M             & 1.8M                 & 21.0     & 34.34 +/- 0.10            & 1.54 +/- 0.05 \\\midrule
                                & \multicolumn{5}{c|}{\textit{Tensor2Tensor, V100}}                                                                        \\ \midrule
Vanilla Transformer             & 35M             & 1.3M                 & 15.4              & 37.19 +/- 0.07                     & -                      \\
Transformer+GELU                & 35M             & 1.2M                 & 14.1              & 37.11 +/- 0.02                     & 1.05 +/- 0.02          \\
Transformer++                   & 35M             & 1.3M                 & 14.7              & 36.23 +/- 0.11                     & 1.54 +/- 0.05          \\
Primer                          & 34M             & 1.2M                 & 13.8              & \textbf{35.06 +/- 0.15}            & \textbf{2.13 +/- 0.11} \\
Primer-EZ                       & 35M             & 1.1M                 & 13.3              & 35.16 +/- 0.13                     & 2.03 +/- 0.09          \\ \midrule                                 & \multicolumn{5}{c|}{\textit{T5, TPUv2}}                                                                                  \\ \midrule
Vanilla Transformer             & 35M             & 2.1M                 & 23.9              & \multicolumn{1}{l}{23.30 +/- 0.02} & -                      \\
Transformer+GELU                & 35M             & 2.1M                 & 23.8              & 23.39 +/- 0.02                     & 0.97 +/- 0.03          \\
Transformer++                   & 35M             & 2.1M                 & 24.2              & 23.04 +/- 0.02                     & 1.33 +/- 0.05          \\
Evolved Transformer                   & 38M             & 1.6M                 & 18.7              & 23.08 +/- 0.02                     & 1.23 +/- 0.02          \\
Primer                          & 36M             & 2.0M                 & 22.9              & 22.71 +/- 0.03                     & 1.72 +/- 0.01          \\
Primer-EZ                       & 36M             & 2.0M                 & 22.5     & \textbf{22.62 +/- 0.02}            & \textbf{1.75 +/- 0.03}  \\ \bottomrule
\end{tabular}
\end{table}

\subsection{Ablation and Insertion Studies}
\label{sec:ablation_and_insertion}

One of the core motivations of this work is to develop simple and robust Transformer modifications. To that end, we study the individual effectiveness of each \modelname modification, described in Section~\ref{sect:primer_model} of the main text. We measure this effectiveness using insertion and ablation studies. In the insertion studies we add each modification in isolation to a vanilla Transformer. In the ablation studies we remove each modification from \modelname one at a time. We are interested in how these modifications affect performance not just in our search library, Tensor2Tensor, but also in other libraries. Thus, we perform these insertion and ablation studies in a different library, T5, as a well, and use modification transferability as the key guiding metric for our modeling recommendations.\looseness=-1

\begin{figure}[h!]
  \centering
  \begin{minipage}{0.45\linewidth}
    \centering
    \includegraphics[width=0.8\linewidth]{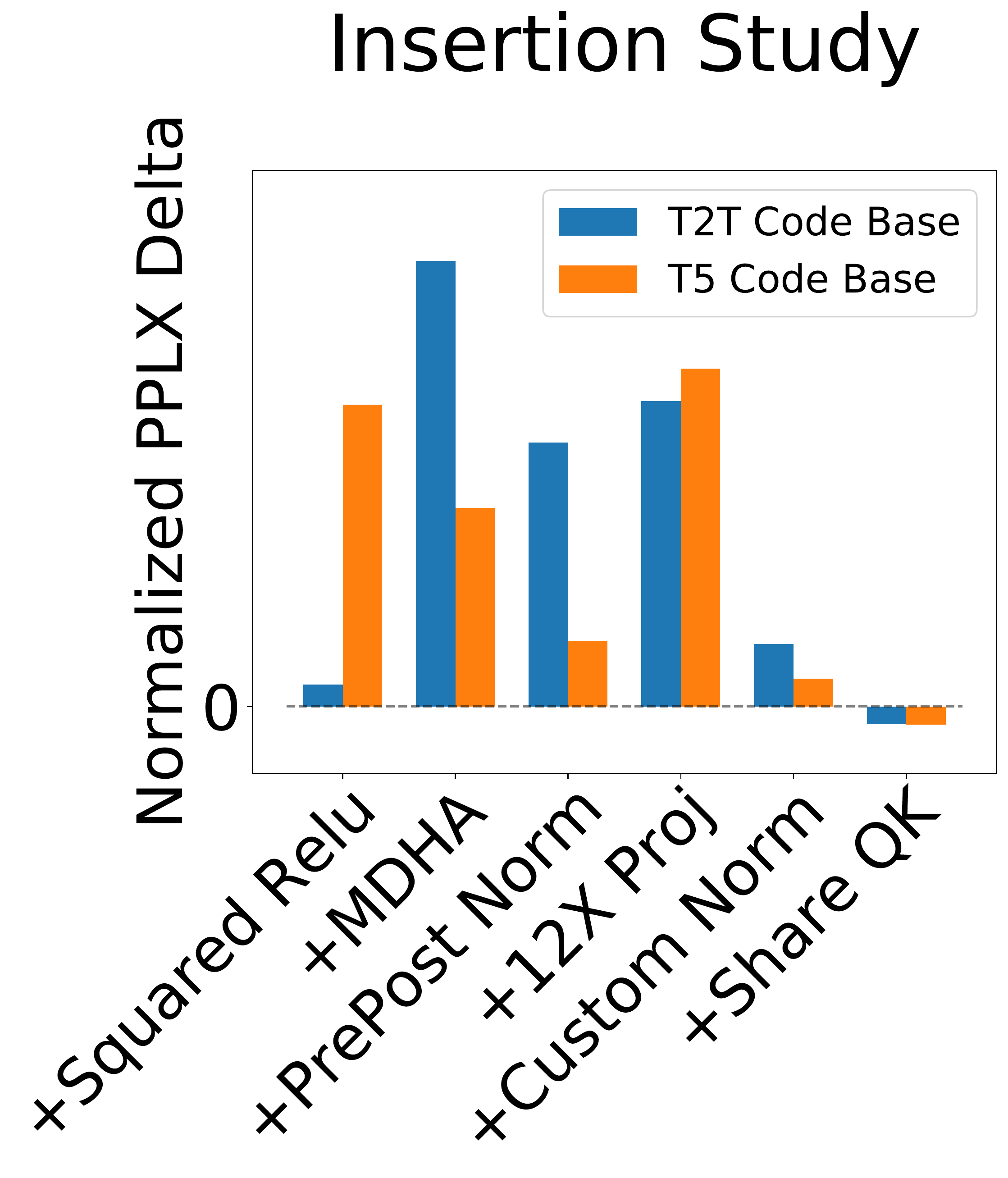}
  \end{minipage}
\begin{minipage}{0.45\linewidth}
    \centering
    \includegraphics[width=0.8\linewidth]{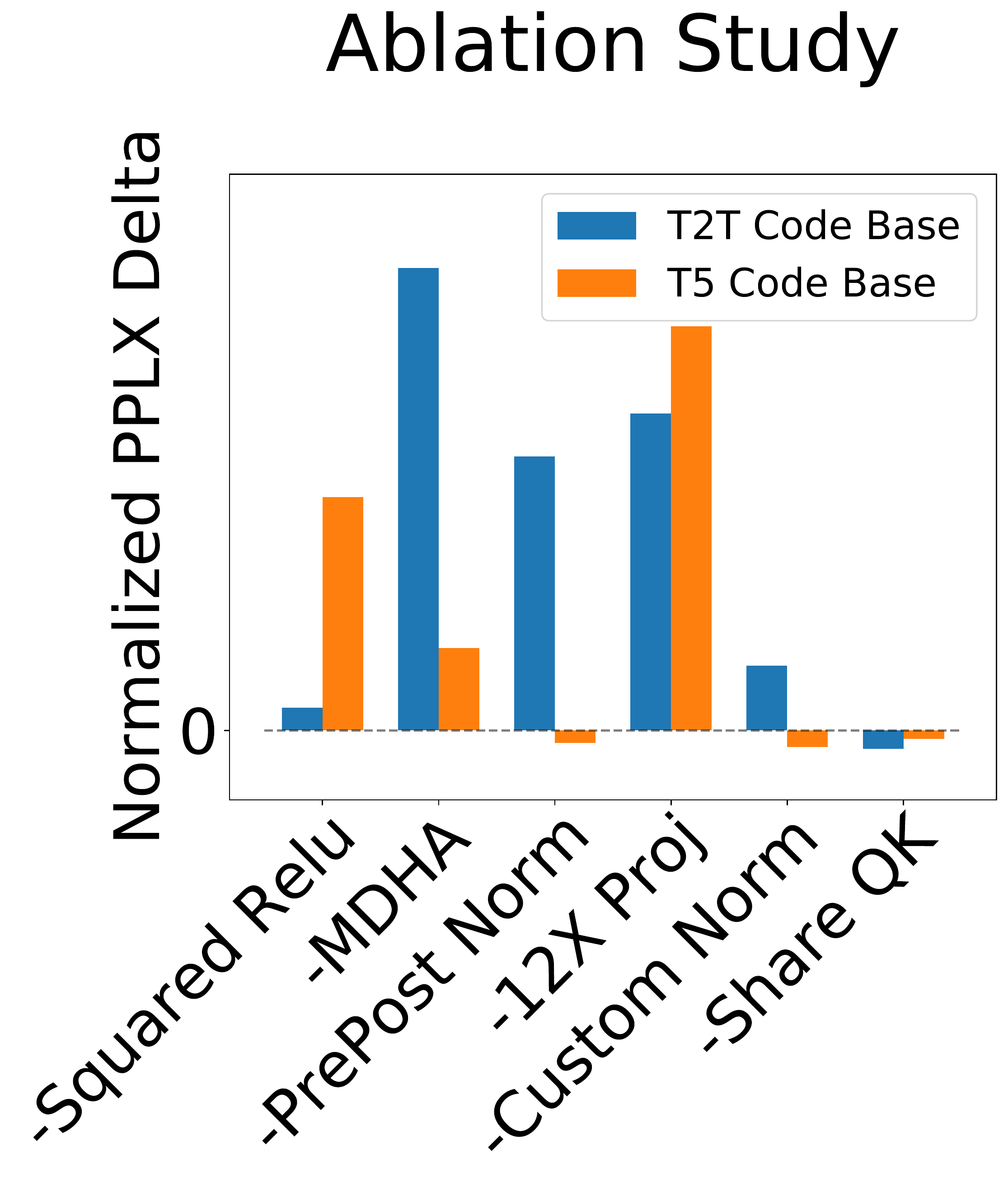}
  \end{minipage}
  \caption{Investigation into transferability of \modelname modifications on LM1B at $\sim$35M parameters. In the ``Insertion Study'' we insert each of the modifications into a vanilla Transformer. In the ``Ablation Study,'' we remove each modification from \modelname. ``Normalized PPLX Delta'' indicates the degree to which the treated models are affected by these modifications; values are normalized to be comparable across code bases and so that positive values indicate beneficial techniques in both studies. Likewise, negative values indicate harmful techniques in both studies.}
  \label{fig:ablation_study}
\end{figure}

The results of these studies are shown in Figure~\ref{fig:ablation_study}. ``Normalized PPLX Delta'' describes the degree to which a modification helps or hurts performance. For baseline perplexity, $P_b$, and modification perplexity, $P_m$, ``Normalized PPLX Delta'' is defined as $\frac{P_b - P_m}{P_b}$ in the insertion study and $\frac{P_m - P_b}{P_b}$ for the ablation study. These definitions differ so that a positive value always indicates that the modification is good and a negative value always indicates that the modification is bad. Three techniques are beneficial in all scenarios. The first is ``12X proj,'' which increases the size of the Transformer feed forward upwards projection while controlling for parameters. We find this works well for smaller models but is not useful at larger sizes. The second two, MDHA and squared ReLUs, are the defining modifications of \modelnamelite, a simpler model that captures much of the gains of the full \modelname.

\subsection{Full Training Details}
\label{sect:app_full_training_details}

In all experiments, we use previously published hyperparameter settings that were tuned for Transformer, with regularization disabled and no additional tuning for \modelname. In Tensor2Tensor (T2T) these are the \textsc{transformer\_tpu} hyperparameters and in T5 and Lingvo these are the open-sourced parameters used in previous T5  studies~\cite{2020t5, narang2021}. They both specify an Adafactor optimizer~\cite{Shazeer2018AdafactorAL}, with 10K warmup steps at a learning rate of 0.01, followed by reciprocal square root learning rate decay. T2T uses positional embeddings and subword tokenization, while T5 and Lingvo use relative attention~\cite{Shaw2018SelfAttentionWR} and SentencePieces~\cite{Kudo2018SentencePieceAS}.

For LM1B, we use the T2T default settings of max sequence length of 64 and batches of 4096 tokens; this is appropriate because LM1B has an average sequence length of roughly 32. For C4 and PG19, we use the T5 default of a max sequence length of 512. For one-shot pretraining, we use a max sequence length of 1024. In Section~\ref{sect:t5_medium_experiments} we use batches of 65K tokens, in Section~\ref{sect:t5_large_experiments} we use batches of 1M tokens, and in Section~\ref{sect:one_shot} we uses batches of 2M tokens.

\subsection{Power Law Compute Savings Derivations}
\label{sect:app_power_law_derivations}

In Section~\ref{sect:search_task_comparison} of the main text, we reproduce the results of Kaplan et al.~\cite{Kaplan2020ScalingLF} and show that, at optimal parameter sizing, the relationship between language model quality and training compute follows a power law: $l = ac^{-k}$, where $l$ is validation loss, $c$ is training compute, and $a$ and $k$ are empirical constants. This is represented as a line in double log space (Figure~\ref{fig:power_law_comparison}): $\log l = -k\log c + \log a$. However, these lines are not the same for each architecture we compare. The lines are roughly parallel but shifted up and down. Thus, defining the shift between two architectures' lines as $\log b^{k}$, we can derive the relationship of their training costs as:

\begin{gather*}
-k\log c_{0} + \log a_{0} = -k\log c_{1} + \log a_{0} + \log b^{k} \\
-k\log c_{0} = -k\log c_{1} + \log b^{k} \\
c_{0}^{-k} = {b}^{k} c_{1} ^ {-k} \\
c_{0} = c_{1}/b \\
\end{gather*}

where $b$ is a consistent reduction factor regardless of $l$. Compute savings, $s$, for using a superior architecture can now be calculated as:

\begin{gather*}
s = c_{1} - c_{0} \\
s = c_{1} - c_{1}/b = c_{1} (1 - 1/b) \\
\text{or} \\
c_{1} = \frac{s}{1 - 1/b} \\
\end{gather*}

Plugging this into the original power law relationship for $c_{1}$ we get:

\begin{gather*}
l = a_1 \left( \frac{s}{1 - 1/b} \right)^{-k} \\
l = a_1(1 - 1/b)^ks^{-k} \\
\end{gather*}

Thus, the relationship between quality and compute savings yielded by an improved architecture also follows a power law with coefficient $a_1(1 - 1/b)^k$. This relationship is intuitive when recognizing that the compute reduction factor ${b}$ is consistent for all values of $l$ and thus a power law investment of training compute with relation to $l$ results in a power law savings with relation to $l$ as well.

\subsection{Exact T5 Numbers for Medium Sized Experiments}
\label{sect:app_t5_medium_numbers}

\begin{table}[h]
\caption{Language modeling comparison on larger datasets, transferring \modelname to the T5 codebase. In this transferred regime, \modelname improves upon all baselines. Furthermore, \modelnamelite not only reaches parity with \modelname, but in some cases, surpasses it. Switch Transformer and Synthesizer also benefit from the \modelnamelite modifications. Compute budget comparison points are chosen according to how long it takes vanilla baselines to reach 525K and 1M training steps. Perplexities are given with respect to SentencePieces. This table has the precise numbers for Figure~\ref{fig:robustness}.}
\label{table:medium_t2t}
\begin{tabular}{@{}lc|ccc|ccc@{}}
                                            & \multicolumn{1}{l|}{} & \multicolumn{3}{c|}{\textit{Baseline Compute @525K}}                       & \multicolumn{3}{c}{\textit{Baseline Compute @1M}}                         \\
                                            & \multicolumn{1}{l|}{} & \multicolumn{1}{l}{} & \multicolumn{1}{l}{} & \multicolumn{1}{l|}{}        & \multicolumn{1}{l}{} & \multicolumn{1}{l}{} & \multicolumn{1}{l}{}        \\
\multicolumn{1}{l|}{\multirow{2}{*}{Model}} & Params                & Steps                & PPLX                 & \multicolumn{1}{l|}{Speedup} & Steps                & PPLX                 & \multicolumn{1}{l}{Speedup} \\ \cmidrule(l){2-8} 
\multicolumn{1}{l|}{}                       & \multicolumn{7}{c|}{\textit{C4}}                                                                                                                                               \\ \midrule
\multicolumn{1}{l|}{Vanilla Transformer}                & 110M                  & 525K                 & 20.61                & -                            & 1M                   & 19.82                & -                           \\
\multicolumn{1}{l|}{Transformer+GELU}                 & 110M                  & 524K                 & 20.34                & 1.20                         & 998K                 & 19.58                & 1.26                        \\
\multicolumn{1}{l|}{Transformer++}          & 110M                  & 524K                 & 20.03                & 1.52                         & 998K                 & 19.28                & 1.64                        \\
\multicolumn{1}{l|}{Evolved Transformer}          & 110M                  & 351K                 & 20.79                & 0.89                         & 668K                 & 19.84                & 0.98                        \\
\multicolumn{1}{l|}{Primer}                 & 110M                  & 483K                 & \textbf{19.82}                & 1.68                         & 920K                 & \textbf{19.07}                & \textbf{1.91}                        \\
\multicolumn{1}{l|}{Primer-EZ}              & 110M                  & 471K                 & 19.83                & \textbf{1.71}                         & 896K                 & \textbf{19.07}                & 1.90                        \\ \midrule
\multicolumn{1}{l|}{Switch Transformer}     & 550M                  & 525K                 & 17.16                & -                            & 1M                   & 16.32                & -                           \\
\multicolumn{1}{l|}{Switch Primer}          & 550M                  & 474K                 & \textbf{16.56}                & \textbf{1.45}                         & 900K                 & \textbf{15.82}                & \textbf{1.56}                        \\ \midrule
\multicolumn{1}{l|}{Synthesizer}            & 145M                  & 525K                 & 20.35                & -                            & 1M                   & 19.57                & -                           \\
\multicolumn{1}{l|}{\hspace{0.4cm}+ Squared ReLU}         & 145M                  & 523K                 & \textbf{19.55}                & \textbf{1.74}                         & 996K                 & \textbf{18.83}                & \textbf{1.96}                        \\ \midrule
\multicolumn{1}{l|}{}                       & \multicolumn{7}{c|}{\textit{PG19}}                                                                                                                                             \\ \midrule
\multicolumn{1}{l|}{Vanilla Transformer}    & 110M                  & 525K                 & 16.39                & -                            & 1M                   & 15.83                & -                           \\
\multicolumn{1}{l|}{Transformer+GELU}       & 110M                  & 524K                 & 16.35                & 1.01                         & 998K                 & 15.84                & 0.95                        \\
\multicolumn{1}{l|}{Transformer++}          & 110M                  & 524K                 & 16.15                & 1.18                         & 998K                 & 15.64                & 1.20                         \\
\multicolumn{1}{l|}{Primer}                 & 110M                  & 483K                 & 15.96                & 1.68                         & 920K                 & \textbf{15.31}                & 1.81                        \\
\multicolumn{1}{l|}{Primer-EZ}              & 110M                  & 471K                 & \textbf{15.84}                & \textbf{1.74}                         & 896K                 & 15.37                & \textbf{1.98}                        \\ \bottomrule
\end{tabular}
\end{table}

\subsection{Performance on Individual One-Shot Tasks}
\label{sect:individual_one_shot}

\begin{table}[H]
\begin{tabular}{@{}ccccccc@{}}
\textbf{Task}        & \textbf{Metric}      & \textbf{\begin{tabular}[c]{@{}c@{}}Transf.\\ 1/3\end{tabular}} & \textbf{\begin{tabular}[c]{@{}c@{}}Transf.\\ Full\end{tabular}} & \textbf{\begin{tabular}[c]{@{}c@{}}Primer\\ 1/3\end{tabular}} & \multicolumn{1}{c|}{\textbf{\begin{tabular}[c]{@{}c@{}}Primer\\ Full\end{tabular}}} & \textbf{\begin{tabular}[c]{@{}c@{}}GPT-3\\ XL\end{tabular}} \\ \midrule
Pretraining & pplx                 & 15.3                                                                                & 14.3                                                                                 & 14.3                                                                           & \multicolumn{1}{c|}{13.5}                                                                            & -                          \\
\multicolumn{1}{l}{} & \multicolumn{1}{l}{} & \multicolumn{1}{l}{}                                                                & \multicolumn{1}{l}{}                                                                 & \multicolumn{1}{l}{}                                                           & \multicolumn{1}{l|}{}                                                                                & \multicolumn{1}{l}{}       \\ \midrule
\multicolumn{7}{c}{\textit{Question Answering Tasks}}\\ \midrule
TriviaQA             & acc                  & \cellcolor[HTML]{C0C0C0}$22.5 \pm 0.4$                                                                      & $26.8 \pm 0.5$                                                                       & $27.5 \pm 0.4$                                                                 & \multicolumn{1}{c|}{$\mathbf{32.2 \pm 0.5}$}                                                                  & 26.5                       \\
WebQs                & acc                  & $9.1 \pm 0.5$                                                                       & $9.6 \pm 0.4$                                                                        & $9.8 \pm 0.8$                                                                  & \multicolumn{1}{c|}{$10.4 \pm 0.3$}                                                                  & 9.2                       \\
NQs                  & acc                  & \cellcolor[HTML]{C0C0C0}$5.8 \pm 0.2$                                                                       & $6.7 \pm 0.2$                                                                        & $\mathbf{7.8 \pm 0.5}$                                                                  & \multicolumn{1}{c|}{$\mathbf{9.1 \pm 0.3}$}                                                                   & 5.4                       \\
\multicolumn{1}{l}{} & \multicolumn{1}{l}{} & \multicolumn{1}{l}{}                                                                & \multicolumn{1}{l}{}                                                                 & \multicolumn{1}{l}{}                                                           & \multicolumn{1}{l|}{}                                                                                & \multicolumn{1}{l}{}       \\
SQuADv2              & f1                   & \cellcolor[HTML]{C0C0C0}$54.2 \pm 2.4$                                                                      & $65.4 \pm 2.9$                                                                       & $64.2 \pm 3.7$                                                                 & \multicolumn{1}{c|}{$67.8 \pm 1.2$}                                                                  & 54                         \\
CoQa                 & f1                   & \cellcolor[HTML]{C0C0C0}$52.5 \pm 1.1$                                                                      & $57.7 \pm 1.2$                                                                       & $59.1 \pm 0.9$                                                                 & \multicolumn{1}{c|}{$\mathbf{61.2 \pm 0.7}$}                                                                  & 66.1                       \\
DROP                 & f1                   & \cellcolor[HTML]{C0C0C0}$21.5 \pm 0.4$                                                                      & $23.4 \pm 0.2$                                                                       & $\mathbf{24.8 \pm 0.5}$                                                                 & \multicolumn{1}{c|}{$\mathbf{26.5 \pm 0.2}$}                                                                  & 23                         \\
Quac                 & f1                   & $30.1 \pm 0.5$                                                                      & $30.9 \pm 0.7$                                                                       & $28.9 \pm 0.9$                                                                 & \multicolumn{1}{c|}{$30.2 \pm 0.7$}                                                                  & 32.3                       \\
\multicolumn{1}{l}{} & \multicolumn{1}{l}{} & \multicolumn{1}{l}{}                                                                & \multicolumn{1}{l}{}                                                                 & \multicolumn{1}{l}{}                                                           & \multicolumn{1}{l|}{}                                                                                & \multicolumn{1}{l}{}       \\
LAMBADA              & acc                  & \cellcolor[HTML]{C0C0C0}$51.5 \pm 0.9$                                                                      & $55.2 \pm 1.3$                                                                       & $54.5 \pm 1.1$                                                                 & \multicolumn{1}{c|}{$56.8 \pm 0.9$}                                                                  & 58.3                       \\
\multicolumn{1}{l}{} & \multicolumn{1}{l}{} & \multicolumn{1}{l}{}                                                                & \multicolumn{1}{l}{}                                                                 & \multicolumn{1}{l}{}                                                           & \multicolumn{1}{l|}{}                                                                                & \multicolumn{1}{l}{}       \\
QA Average           & avg              & 30.9                                                                                & 34.5                                                                                 & 34.6                                                                           & \multicolumn{1}{c|}{36.8}                                                                            & 34.3                       \\
\midrule
\multicolumn{7}{c}{\textit{Multi-Choice Schema Tasks}}\\ \midrule
HellaSwag            & acc                  & \cellcolor[HTML]{C0C0C0}$55.7 \pm 0.3$                                                                      & $59.5 \pm 0.2$                                                                       & $\mathbf{60.2 \pm 0.3}$                                                                 & \multicolumn{1}{c|}{$\mathbf{63.3 \pm 0.2}$}                                                                  & 53.5                       \\
StoryCloze           & acc                  & $75.2 \pm 0.3$                                                                      & $75.9 \pm 0.4$                                                                       & $\mathbf{76.9 \pm 0.2}$                                                                 & \multicolumn{1}{c|}{$\mathbf{77.5 \pm 0.3}$}                                                                  & 74.2                       \\
\multicolumn{1}{l}{} & \multicolumn{1}{l}{} & \multicolumn{1}{l}{}                                                                & \multicolumn{1}{l}{}                                                                 & \multicolumn{1}{l}{}                                                           & \multicolumn{1}{l|}{}                                                                                & \multicolumn{1}{l}{}       \\
Winogrande           & acc                  & \cellcolor[HTML]{C0C0C0}$55.4 \pm 0.3$                                                                      & $58.4 \pm 0.4$                                                                       & $58.8 \pm 0.3$                                                                 & \multicolumn{1}{c|}{$\mathbf{60.4 \pm 0.2}$}                                                                  & 59.1                       \\
\multicolumn{1}{l}{} & \multicolumn{1}{l}{} & \multicolumn{1}{l}{}                                                                & \multicolumn{1}{l}{}                                                                 & \multicolumn{1}{l}{}                                                           & \multicolumn{1}{l|}{}                                                                                & \multicolumn{1}{l}{}       \\
PIQA                 & acc                  & $72.6 \pm 0.5$                                                                      & $72.6 \pm 0.3$                                                                       & $73.7 \pm 0.5$                                                                 & \multicolumn{1}{c|}{$\mathbf{75.0 \pm 0.4}$}                                                                  & 74.4                       \\
ARC (Challenge)      & acc                  & \cellcolor[HTML]{C0C0C0}$32.7 \pm 0.4$                                                                      & $34.4 \pm 0.3$                                                                       & $35.6 \pm 0.9$                                                                 & \multicolumn{1}{c|}{$\mathbf{37.4 \pm 0.4}$}                                                                  & 36.4                       \\
ARC (Easy)           & acc                  & $64.5 \pm 0.5$                                                                      & $64.9 \pm 0.5$                                                                       & $65.6 \pm 0.6$                                                                 & \multicolumn{1}{c|}{$\mathbf{67.5 \pm 0.5}$}                                                                  & 55.9                       \\
OpenBookQA           & acc                  & $45.3 \pm 0.9$                                                                      & $46.8 \pm 0.8$                                                                       & $47.9 \pm 0.4$                                                                 & \multicolumn{1}{c|}{$\mathbf{49.3 \pm 0.5}$}                                                                  & 46.4                       \\
\multicolumn{1}{l}{} & \multicolumn{1}{l}{} & \multicolumn{1}{l}{}                                                                & \multicolumn{1}{l}{}                                                                 & \multicolumn{1}{l}{}                                                           & \multicolumn{1}{l|}{}                                                                                & \multicolumn{1}{l}{}       \\
ANLI R1              & acc                  & $33.9 \pm 1.2$                                                                      & $35.5 \pm 0.2$                                                                       & $35.5 \pm 0.4$                                                                 & \multicolumn{1}{c|}{\cellcolor[HTML]{C0C0C0}$34.8 \pm 0.3$}                                                                  & 34.6                       \\
ANLI R2              & acc                  & $33.5 \pm 0.7$                                                                      & $33.4 \pm 0.5$                                                                       & $34.5 \pm 0.6$                                                                 & \multicolumn{1}{c|}{$33.5 \pm 0.4$}                                                                  & 32.7                       \\
ANLI R3              & acc                  & $34.5 \pm 0.7$                                                                      & $35.2 \pm 0.1$                                                                       & \cellcolor[HTML]{C0C0C0}$33.0 \pm 0.3$                                                                 & \multicolumn{1}{c|}{\cellcolor[HTML]{C0C0C0}$33.8 \pm 0.5$}                                                                  & 33.9                       \\
\multicolumn{1}{l}{} & \multicolumn{1}{l}{} & \multicolumn{1}{l}{}                                                                & \multicolumn{1}{l}{}                                                                 & \multicolumn{1}{l}{}                                                           & \multicolumn{1}{l|}{}                                                                                & \multicolumn{1}{l}{}       \\
ReCoRD               & acc                  & \cellcolor[HTML]{C0C0C0}$84.8 \pm 0.1$                                                                      & $86.3 \pm 0.2$                                                                       & $85.8 \pm 0.3$                                                                 & \multicolumn{1}{c|}{$\mathbf{86.7 \pm 0.0}$}                                                                  & 83                         \\
WSC                  & acc                  & $67.4 \pm 0.8$                                                                      & $66.8 \pm 1.2$                                                                       & $69.3 \pm 1.3$                                                                 & \multicolumn{1}{c|}{$68.9 \pm 1.2$}                                                                  & 62.5                       \\
BoolQ                & acc                  & \cellcolor[HTML]{C0C0C0}$58.9 \pm 1.1$                                                                      & $63.6 \pm 2.1$                                                                       & $60.7 \pm 0.8$                                                                 & \multicolumn{1}{c|}{$64.7 \pm 2.0$}                                                                  & 63.7                       \\
CB                   & acc                  & $56.3 \pm 2.5$                                                                      & $53.0 \pm 2.7$                                                                       & $55.4 \pm 3.3$                                                                 & \multicolumn{1}{c|}{$56.6 \pm 9.6$}                                                                  & 48.2                       \\
RTE                  & acc                  & \cellcolor[HTML]{C0C0C0}$48.4 \pm 1.2$                                                                      & $53.6 \pm 2.5$                                                                       & $54.3 \pm 1.5$                                                                 & \multicolumn{1}{c|}{$52.9 \pm 2.8$}                                                                  & 49.5                       \\
COPA                 & acc                  & \cellcolor[HTML]{C0C0C0}$80.2 \pm 3.2$                                                                      & $87.2 \pm 1.2$                                                                       & $84.8 \pm 1.5$                                                                 & \multicolumn{1}{c|}{$87.5 \pm 1.1$}                                                                  & 74                         \\
WiC                  & acc                  & $51.6 \pm 0.2$                                                                      & $51.0 \pm 0.5$                                                                       & $\mathbf{51.7 \pm 0.1}$                                                                 & \multicolumn{1}{c|}{$\mathbf{51.8 \pm 0.1}$}                                                                  & 49.2                       \\
\multicolumn{1}{l}{} & \multicolumn{1}{l}{} & \multicolumn{1}{l}{}                                                                & \multicolumn{1}{l}{}                                                                 & \multicolumn{1}{l}{}                                                           & \multicolumn{1}{l|}{}                                                                                & \multicolumn{1}{l}{}       \\
RACE-h               & acc                  & \cellcolor[HTML]{C0C0C0}$39.4 \pm 0.4$                                                                      & $40.8 \pm 0.4$                                                                       & $40.4 \pm 0.4$                                                                 & \multicolumn{1}{c|}{$\mathbf{43.7 \pm 0.3}$}                                                                  & 42                         \\
RACE-m               & acc                  & \cellcolor[HTML]{C0C0C0}$50.0 \pm 1.0$                                                                      & $52.6 \pm 0.4$                                                                       & $51.8 \pm 0.8$                                                                 & \multicolumn{1}{c|}{$\mathbf{54.0 \pm 0.4}$}                                                                  & 55.2                       \\
\multicolumn{1}{l}{} & \multicolumn{1}{l}{} & \multicolumn{1}{l}{}                                                                & \multicolumn{1}{l}{}                                                                 & \multicolumn{1}{l}{}                                                           & \multicolumn{1}{l|}{}                                                                                & \multicolumn{1}{l}{}       \\
Multi-Choice Average      & avg             & 53.1                                                                                & 54.7                                                                                 & 55                                                                             & \multicolumn{1}{c|}{56.2}                                                                            & 54.1                       \\
\bottomrule
\end{tabular}
    \caption{Comparison between Transformer+GELU and \modelname at 1.9B parameters on downstream one-shot tasks at 1/3 and full pretraining compute budgets. One-shot sample means and standard deviations are computed using the evaluated performance of 5 weight checkpoints. \textbf{Bold numbers} denote improved one-shot performance and \hl{shaded numbers} denote worse one-shot performance compared to Transformer with full compute that is statistically significant under an independent t-test with p-value threshold 0.05. \modelname achieves the same performance as Transformer when given 1/3 the training compute and stronger performance on a majority of tasks when given the same training compute. GPT-3 XL~\cite{brown2020language} scores are provided as a grounding reference point; they should not be closely compared to our results as the models have different pretraining configurations.}
    \label{table:one_shot_scores}
\end{table}

\begin{figure}[H]
    \centering
    \includegraphics[width=0.9\linewidth]{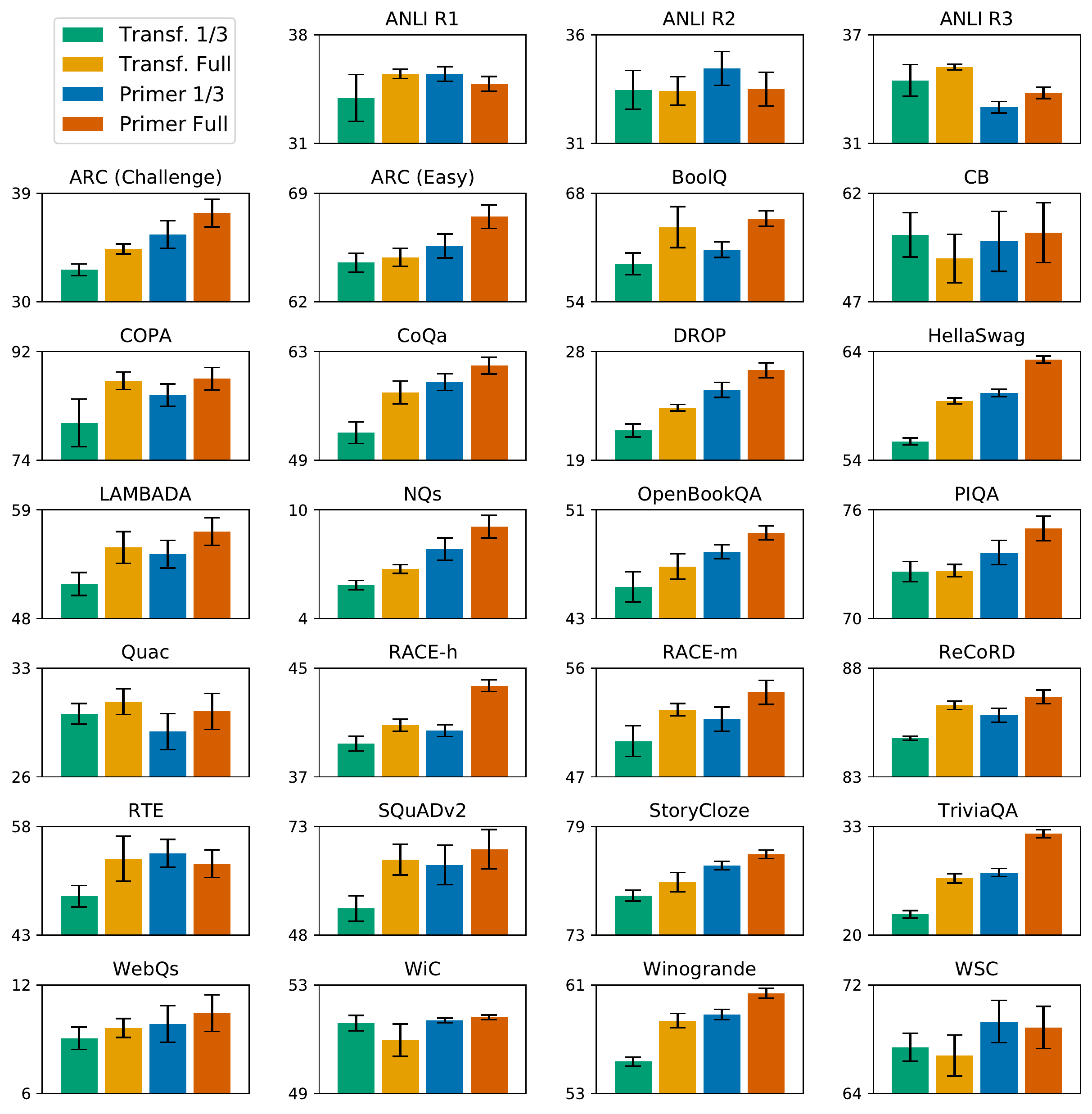}
    \caption{Comparison between Transformer+GELU and \modelname at 1.9B parameters on downstream one-shot tasks at 1/3 and full pretraining compute budgets. 95\% confidence intervals are provided according to an independent t-test, using a sample of 5 pretraining weight checkpoints. \modelname achieves roughly the same performance as Transformer when given 1/3 the pretraining compute and stronger performance on a majority of tasks when given the same pretraining compute. Exact numbers are presented in Table~\ref{table:one_shot_scores}.}
    \label{fig:one_shot_bars}
\end{figure}

\subsection{Masked Language Modeling}
\label{sect:app_masked_language_modeling}

Encoder-decoder style masked language modeling (MLM) is not the focus of this work. However, because it was the focus of the original T5 project, we include MLM comparisons here for completeness (Table~\ref{}). Specifically, we use the exact comparison configuration used by Narang et al.\cite{narang2021}, who benchmarked several Transformer variants; the one difference is that we only run model training one time, since this regime is not the focus of our study. For ``\modelnamelite Decoder'' we use a Transformer++ encoder and a \modelnamelite decoder. Our treatments demonstrate that the \modelnamelite modifications have the capacity to improve encoder-decoder MLM models, but perhaps to a lesser degree, when compared to Transformer++. We believe this indicates that decoder-only LM and encoder-decoder MLM benefit from different modeling decisions -- something that could be studied in future works. We also believe that running our search on encoder-decoder MLM directly could yield modifications that are more beneficial for this task.

\begin{table}[H]
\begin{tabular}{@{}lccccc@{}}
\toprule
\textbf{Model}       & \textbf{Params} & \textbf{Pretraining Log PPLX} & \textbf{SGLUE} & \textbf{XSum}  & \textbf{WebQ}  \\ \midrule
Vanilla Transformer* & 223M            & 1.838                         & 70.97          & 17.78          & 23.02          \\
Transformer+GeLU*    & 223M            & 1.838                         & 73.67          & 17.86          & 25.13          \\
Transformer++        & 224M            & 1.792                         & 75.65          & \textbf{17.90} & \textbf{25.92} \\
Primer-EZ Decoder      & 224M            & \textbf{1.787}                & \textbf{76.69} & 17.87          & 24.87          \\ \bottomrule
\end{tabular}
\caption{Masked language modeling comparison on C4 in T5 with encoder-decoder style models. These results are run in the exact same configuration as Narang et al.~\cite{narang2021}, although we only run our models once, as MLM is not the focus of our work. * indicates rows that are taken from that study.}
\label{}
\end{table}

\subsection{Carbon Emission Estimates}
\label{sect:app_carbon_estimates}

Following the recommendations of Patterson et al.~\cite{Patterson2021CarbonEA}, we release the carbon emission estimates for our largest experiments.

To estimate the carbon emissions\footlabel{ft:co2e_accounting_methodology}{Our CO\textsubscript{2}e accounting methodology for data center net carbon intensity does not currently fit the Greenhouse Gas (GHG) protocol for emissions reporting (Scope 2 and 3 for electricity). This deviation is due to a change in methodology where Google uses hourly life cycle emission factors, while the GHG Protocol generally relies on annual operating emission factor data. Google chooses to share these modified metrics as part of our 24/7 carbon-free energy (CFE) program, focused on our goal of achieving 100\% 24/7 local CFE by 2030. Google’s target for 2030 goes beyond the traditional Scope 2 rules to restrict both the location and the accounting period. This means that, instead of anywhere in a continent, the CFE purchase should be on the same geographically local grid; and instead of the accounting period being one year, the accounting should be within the same hour.}$^{,}$\footlabel{ft:co2e_strategies}{While electricity consumption is relatively straightforward, strategies to reduce greenhouse gas emissions are not. For details on the distinction between conventional carbon offsets, Google’s goal for 2030 of 24/7 CFE for its global data centers and campuses, and what it is doing now to set the groundwork for 2030, please see Appendix B of Patterson et al.~\cite{Patterson2021CarbonEA}.} for our architecture search, we build off of the measurements taken by Patterson et al. Their emissions estimate for architecture search is 3.2 MTCO\textsubscript{2}e for 1360 days of TPUv2 usage~\cite{Patterson2021CarbonEA}. Here, we use 1145.8 days of TPUv2 compute for our search. Additionally, the PUE for our data center\footlabel{ft:soco_grid}{Each data center is located within a Regional Grid, which is the geographic basis for Google’s 24/7 CFE goals. For our data center in Georgia, the Regional Grid is the Southern Company balancing authority.} at the time of our search was 1.08 instead of 1.10, and its net carbon intensity average was 0.336 MTCO\textsubscript{2}e/MWh instead of 0.431 MTCO\textsubscript{2}e/MWh.\footlabel{ft:co2e_intensity}{The net carbon intensity at a particular data center is based on accounting for hourly emission reductions via real time, local carbon-free energy purchases. This is calculated using the 24/7 carbon-free energy methodology, which can be reviewed in greater depth in ``24/7 Carbon-Free Energy: Methodologies and Metrics''~\cite{googlecarbon}.}$^{,}$\footlabel{ft:co2e_2020}{The carbon intensity values utilized in this paper are at the annual 2020 grid level for each data center in which the models were run.} Thus, the proportional emissions estimate for our architecture search experiments is 3.2 MTCO\textsubscript{2}e $* \frac{1145.8}{1360} * \frac{1.08}{1.10} * \frac{336}{431} =$ 2.06 MTCO\textsubscript{2}e. For comparison, a round trip plane ticket from San Francisco to New York for a single passenger is $\sim$1.2 MTCO\textsubscript{2}e~\cite{Patterson2021CarbonEA} and so our search costs roughly 1.72 such plane tickets.

We follow the same process of building off of the Patterson et al. measurements to estimate emissions for our large scale T5 experiments. The Patterson et al. emissions estimate for 11B parameter T5 is 46.7 tCO\textsubscript{2}e for 10,249 days of TPUv3 usage. Our T5 models are smaller, and so only require 687.5 TPUv3 days to train on average. We run 3 trainings (\modelname, original T5 and T5++) to show \modelname's improvements over baselines, yielding a total of 2062.5 TPUv3 days. When we ran our experiments, the data center\footlabel{ft:tpe_grid}{For our data center in Taipei, for purposes of Google's 24/7 CFE accounting, the Regional Grid is Taiwan.} PUE was 1.10 instead of 1.12 and its net carbon intensity average was 0.540 MTCO\textsubscript{2}e/MWh instead of 0.545 MTCO\textsubscript{2}e/MWh. Thus, the proportional total estimate for these T5 model trainings is 46.7 MTCO\textsubscript{2}e $* \frac{2062.5}{10,249} * \frac{1.10}{1.12} * \frac{540}{545} =$ 8.54 MTCO\textsubscript{2}e.

To estimate the emissions of our one-shot pretrainings in Lingvo, we measure system average power in the same manner as Patterson et al.~\cite{Patterson2021CarbonEA}. Including memory, network interface, fans, and host CPU, the average power per TPUv4 chip is 343W. We use the same equation as Patterson et al. to calculate CO\textsubscript{2}e for our 2 large scale pretrainings: 2 $*$ 343W $*$ 71,800h $*$ 1.08(PUE) $*$ 0.055 MTCO\textsubscript{2}e/MWh $=$ 29.26 MTCO\textsubscript{2}e.\footlabel{ft:spp_grid}{For our data center in Oklahoma, for purposes of Google's 24/7 CFE accounting, the Regional Grid is the Southwest Power Pool (SPP) Independent System Operator.}

The emission cost for our large scale T5 and one-shot comparisons are higher than the cost of the architecture search itself. We invest in these large scale comparisons to demonstrate the potential savings of our efficient modifications. For instance, the savings for using \modelname over Transformer described in Section~\ref{sect:one_shot} of the main text equates to 9.75 MTCO\textsubscript{2}e, which alone is $\sim$4.7X the cost of the architecture search. Note, differences in hardware setups affect these savings. For example, the one-shot models were trained in Oklahoma, which has favorable MTCO\textsubscript{2}e/MWh when compared to Georgia, where the \modelname search was conducted. To factor out the effects of these hardware differences, we can instead analyze \modelname's return on investment in terms of FLOPs. The search for Primer cost $\sim$2.14E+21 FLOPs. Training Transformer for the one-shot comparison cost $\sim$2.96E+22 FLOPs, which means the compute saved by \modelname is $\sim$1.98E+22 FLOPs, given that it only requires a third of the compute to achieve the same quality. Thus, \modelname's savings in the one-shot experiments are 9.24X the cost of the architecture search itself, yielding returns on investing in the search. Note that the search cost is a one-time cost and that \modelname can be reused in future trainings to save more compute. For example, our largest models are roughly 100X smaller than the full scale GPT-3~\cite{brown2020language}, and so the return on our search investment can grow if \modelname is scaled up to larger training configurations.

\subsection{Comparison to Evolved Transformer}
\label{sect:app_evolved_transformer}

\begin{table}[h]
\begin{tabular}{@{}l|cc|cc@{}}
& \multicolumn{2}{c|}{\multirow{2}{*}{\textit{LM1B}}} & \multicolumn{2}{c}{\multirow{2}{*}{\textit{C4}}} \\
\multicolumn{1}{c|}{\multirow{2}{*}{Model}} & \multicolumn{2}{c|}{}                               & \multicolumn{2}{c}{}                             \\ \cmidrule(l){2-5} 
\multicolumn{1}{c|}{}                       & Params              & PPLX @ 1.5M Steps             & Params              & PPLX @ 1M Steps             \\ \midrule
Vanilla Transformer                         & 35M                 & 23.45                         & 110M                & 19.82                       \\
Transformer+GELU                            & 35M                 & 23.68                         & 110M                & 19.58                       \\
Transformer++                               & 35M                 & 23.35                         & 110M                & 19.29                       \\
Evolved Transformer                         & 38M                 & 23.11                         & 110M                & 19.37                       \\
Primer                                      & 36M                 & 22.97                         & 110M                & 18.99                       \\
Primer-EZ                                   & 36M                 & \textbf{22.89}                & 110M                & \textbf{18.93}              \\ \bottomrule
\end{tabular}
\caption{Auto-regressive language modeling comparison between \modelname and various baselines, including the Evolved Transformer, controlling for training steps in T5. These are the same experiments featured in Tables~\ref{table:lm1b} and~\ref{table:medium_t2t}, but with the data presented to compare sample efficiency instead of training compute efficiency.}
\label{table:sample_efficiency}
\end{table}

This work builds off of the Evolved Transformer~\cite{so2019evolved}, which also sought to discover improved sequence models using architecture search. Compute efficiency comparisons to the Evolved Transformer architecture are provided in T5 on LM1B in Table~\ref{table:lm1b} and on C4 in Table~\ref{table:medium_t2t}. Sample efficiency comparisons to the Evolved Transformer architecture are offered in Table~\ref{table:sample_efficiency} on those same experiments. In this section we discuss these comparisons and how they highlight the improvements of our \modelname search over the Evolved Transformer search.

Firstly, our \modelname search aims to improve training compute efficiency, which yields more practical results than the sample efficiency objective of So et al.~\cite{so2019evolved}, who controlled for number of train steps when evaluating models. Evolved Transformer is effective in this controlled-train-step regime when comparing to other baselines, as shown in Table~\ref{table:sample_efficiency}. When controlling for number of training steps in this way, Evolved Transformer is roughly on par with Transformer++ on C4 and is better than Transformer++ on LM1B. However, Evolved Transformer is substantially slower than all other models (see Tables~\ref{table:lm1b} and~\ref{table:medium_t2t}) because it is deeper; we follow the same scaling policy as So et al. of adding additional layers to control for parameters, given that an Evolved Transformer layer has significantly less parameters than a standard Transformer layer. Evolved Transformer’s slowness counteracts its sample efficiency and for this reason its speedup factor is diminished on LM1B and less than 1.0 (indicating a slowdown over vanilla Transformer) on C4 (see Tables~\ref{table:lm1b} and~\ref{table:medium_t2t}). This limits Evolved Transformer's practicality. In contrast, \modelname is designed to specifically address this shortcoming and thus delivers the practical result of substantial compute savings. 

The open-ended nature of the \modelname search also allows for effective modifications that were not available to the Evolved Transformer search. In fact, \textit{none} of the \modelname modifications (see Section~\ref{sect:primer_model}) can be represented in the Evolved Transformer search space, aside from resizing hidden dimension sizes. This is because the Evolved Transformer search space followed a rigid ordering of components and used a vocabulary of unalterable high level building blocks. For example, normalization always preceded weighted transformations and, although there were different weighted transformations to choose from such as self-attention and GLU, those transformations could not be modified by the search. In contrast, the Primer search space allows for the modification of all initialized modules -- such as weighted transformations, activation functions and normalization functions -- as well as allows for macro-level reordering, such as moving normalization after weighted transformations. We believe that this difference in openness is what allowed Primer to develop definitively superior modifications, as demonstrated not only by improved compute efficiency, but also by improved sample efficiency (Table~\ref{table:sample_efficiency}), which is what Evolved Transformer was meant to optimize.

\subsection{Practical Discussion}
\label{sect:app_practical_discussion}

The main motivation of this work is to develop simple and practical changes to Transformers that can be easily adopted. To that end, we provide answers to some questions that practitioners may ask:

\begin{itemize}
\item \textit{Are the \modelname training compute savings going to be the same in all setups?} No. Across our own provided experiments, \modelname yields various compute savings. This is because the compute savings depend on hardware specifics, deep learning library operation speeds, model sample efficiencies on specific tasks, and other factors that may vary across setups. We use the exact replica of T5 training as a demonstration of what savings look like in an established configuration (4.2X), but expect results to vary across configurations.

\item \textit{Can \modelname improve BERT~\cite{devlin2018bert}?} This work has focused on the specific task of auto-regressive language modeling, which, with the development of GPT-3, proves to be important for both traditional NLP applications as well as generative applications. We have only briefly investigated \modelname's application to masked language modeling and encoder-decoder models (Appendix~\ref{sect:app_masked_language_modeling}). Our investigations show that, while \modelname improves upon vanilla Transformer, it is not obviously better than Transformer++. Thus, modifications that work well for auto-regressive language modeling may not be as effective for masked language modeling. Future work could investigate if the \modelname modifications can be integrated into encoder-decoder and encoder-only models in a more effective way that can improve models like BERT. Future work could also apply the search method described here to finding better encoder-based masked language models. 

\item \textit{Do hyperparameter configurations need to be retuned to use \modelname?} Our intention is for \modelname modifications to not require any additional hyperparameter tuning. To that end, in our experiments we did not tune any hyperparameters, and instead used the Transformer hyperparameters from established libraries. However, \modelname may work even better with additional tuning.\looseness=-1

\item \textit{Is \modelnamelite better than \modelname?}
In our comparison experiments, we find that \modelnamelite is sometimes better than \modelname in the T5 codebase. However, in application to other codebases, such as Lingvo and T2T, we find that the full \modelname can give improved performance over \modelnamelite. Thus, we recommend that practitioners first try using \modelnamelite for its ease of implementation and then move on to implementing the full \modelname if they are interested in achieving further gains.

\end{itemize}

\end{document}